\documentclass{article}
\PassOptionsToPackage{numbers, compress}{natbib}
\usepackage[final, eandd, nonanonymous]{neurips_2026}
\usepackage[utf8]{inputenc}
\usepackage[T1]{fontenc}
\usepackage[colorlinks=true, linkcolor=blue!70!black, citecolor=blue!60!black, urlcolor=blue!70!black]{hyperref}
\usepackage{url}
\usepackage{underscore}

\usepackage{booktabs}
\usepackage{amsfonts}
\usepackage{nicefrac}
\usepackage{microtype}
\usepackage{xcolor}
\usepackage{listings}
\usepackage{wrapfig}
\usepackage{graphicx}
\usepackage{xspace}
\usepackage[most]{tcolorbox}   % 
\usepackage{caption}
\usepackage[table]{xcolor}
\usepackage{array}
\usepackage{wrapfig}
\usepackage{placeins}
\usepackage{subcaption}
\usepackage{fontawesome5}
\definecolor{deltahl}{RGB}{255,230,230}
\definecolor{codebg}{RGB}{250,252,255}          % 
\definecolor{coderule}{RGB}{180,200,230}         % 
\definecolor{jkey}{RGB}{0,70,180}               % 
\definecolor{jstr}{RGB}{20,110,40}              % 
\definecolor{jnum}{RGB}{160,60,0}               % 
\definecolor{shellbg}{RGB}{245,248,245}          % 
\definecolor{shellrule}{RGB}{160,185,160}        %
\definecolor{tocrulecolor}{RGB}{40,80,180}       % 
\definecolor{toctitlecolor}{RGB}{30,60,160}      % 

\lstdefinelanguage{json}{
  basicstyle      = \small\ttfamily,
  breaklines      = true,
  captionpos      = t,
  frame           = single,
  rulecolor       = \color{coderule},
  framesep        = 5pt,
  backgroundcolor = \color{codebg},
  showstringspaces= false,
  moredelim       = [is][\color{jkey}\bfseries]{|>}{<|},
  morestring      = [b][\color{jstr}]",
  literate        =
    *{0}{{{\color{jnum}0}}}{1}
     {1}{{{\color{jnum}1}}}{1}
     {2}{{{\color{jnum}2}}}{1}
     {3}{{{\color{jnum}3}}}{1}
     {4}{{{\color{jnum}4}}}{1}
     {5}{{{\color{jnum}5}}}{1}
     {6}{{{\color{jnum}6}}}{1}
     {7}{{{\color{jnum}7}}}{1}
     {8}{{{\color{jnum}8}}}{1}
     {9}{{{\color{jnum}9}}}{1}
     {true}{{{\color{jnum}true}}}{4}
     {false}{{{\color{jnum}false}}}{5}
     {null}{{{\color{jnum}null}}}{4},
}

\lstdefinestyle{shell}{
  language        = bash,
  basicstyle      = \small\ttfamily,
  breaklines      = true,
  captionpos      = t,
  frame           = single,
  rulecolor       = \color{shellrule},
  framesep        = 5pt,
  backgroundcolor = \color{shellbg},
  commentstyle    = \color{gray!70}\itshape,
  showstringspaces= false,
}

\lstdefinestyle{plain}{
  basicstyle      = \small\ttfamily,
  breaklines      = true,
  captionpos      = t,
  frame           = single,
  rulecolor       = \color{gray!35},
  framesep        = 5pt,
  backgroundcolor = \color{gray!4},
  showstringspaces= false,
}
\usepackage{mdframed}
\definecolor{termAcol}{HTML}{C9A227}  %gold
\definecolor{termBcol}{HTML}{2E5C8A}  %blue
\definecolor{termCcol}{HTML}{A33B5E}  %pink
\definecolor{termDcol}{HTML}{2E8B57}  %green
\definecolor{termEcol}{HTML}{6A4C9C}  %purple
\newmdenv[
  topline=false, bottomline=false,
  leftline=true, rightline=true,
  linewidth=1pt,
  innerleftmargin=4pt, innerrightmargin=4pt,
  innertopmargin=0pt, innerbottommargin=0pt,
  skipabove=3pt, skipbelow=3pt,
  linecolor=termAcol,
]{sidemarkA}

\newmdenv[
  topline=false, bottomline=false,
  leftline=true, rightline=true,
  linewidth=1pt,
  innerleftmargin=4pt, innerrightmargin=4pt,
  innertopmargin=0pt, innerbottommargin=0pt,
  skipabove=3pt, skipbelow=3pt,
  linecolor=termBcol,
]{sidemarkB}

\newmdenv[
  topline=false, bottomline=false,
  leftline=true, rightline=true,
  linewidth=1pt,
  innerleftmargin=4pt, innerrightmargin=4pt,
  innertopmargin=0pt, innerbottommargin=0pt,
  skipabove=3pt, skipbelow=3pt,
  linecolor=termCcol,
]{sidemarkC}

\newmdenv[
  topline=false, bottomline=false,
  leftline=true, rightline=true,
  linewidth=1pt,
  innerleftmargin=4pt, innerrightmargin=4pt,
  innertopmargin=0pt, innerbottommargin=0pt,
  skipabove=3pt, skipbelow=3pt,
  linecolor=termDcol,
]{sidemarkD}

\newmdenv[
  topline=false, bottomline=false,
  leftline=true, rightline=true,
  linewidth=1pt,
  innerleftmargin=4pt, innerrightmargin=4pt,
  innertopmargin=0pt, innerbottommargin=0pt,
  skipabove=3pt, skipbelow=3pt,
  linecolor=termEcol,
]{sidemarkE}

\usepackage{soul}
\newcommand{\keytermA}[1]{{\setulcolor{termAcol}\setul{2pt}{1.2pt}\ul{#1}}}
\newcommand{\keytermB}[1]{{\setulcolor{termBcol}\setul{2pt}{1.2pt}\ul{#1}}}
\newcommand{\keytermC}[1]{{\setulcolor{termCcol}\setul{2pt}{1.2pt}\ul{#1}}}
\newcommand{\keytermD}[1]{{\setulcolor{termDcol}\setul{2pt}{1.2pt}\ul{#1}}}
\newcommand{\keytermE}[1]{{\setulcolor{termEcol}\setul{2pt}{1.2pt}\ul{#1}}}
\DeclareCaptionFormat{listing}{\textit{#1#2#3}}
\tcbuselibrary{breakable}
\newcommand{\apptocline}[3]{%        #1=label  #2=entry text  #3=page
  \hyperref[#1]{\textcolor{tocrulecolor}{#2}}%
  \dotfill\textcolor{tocrulecolor}{#3}\\[3pt]}
\newcommand{\apptocsubline}[3]{%
  \hspace{1.4em}\hyperref[#1]{\textcolor{tocrulecolor}{#2}}%
  \dotfill\textcolor{tocrulecolor}{#3}\\[2pt]}
\newcommand{\cready}[1]{}
\newcommand{\eee}{\textit{Every Eval Ever}\xspace}
\newcommand{\eeeshort}{\textit{EEE}\xspace}
\newcommand{\todo}[1]{\textcolor{red}{\textbf{[#1]}}}

\usepackage{fontawesome5}   
\newcommand{\githubrepo}[1]{\href{#1}{\raisebox{-0.15ex}{\faGithub}}}
\newcommand{\hfrepo}[1]{\href{#1}{\raisebox{-0.1ex}{\includegraphics[height=1em]{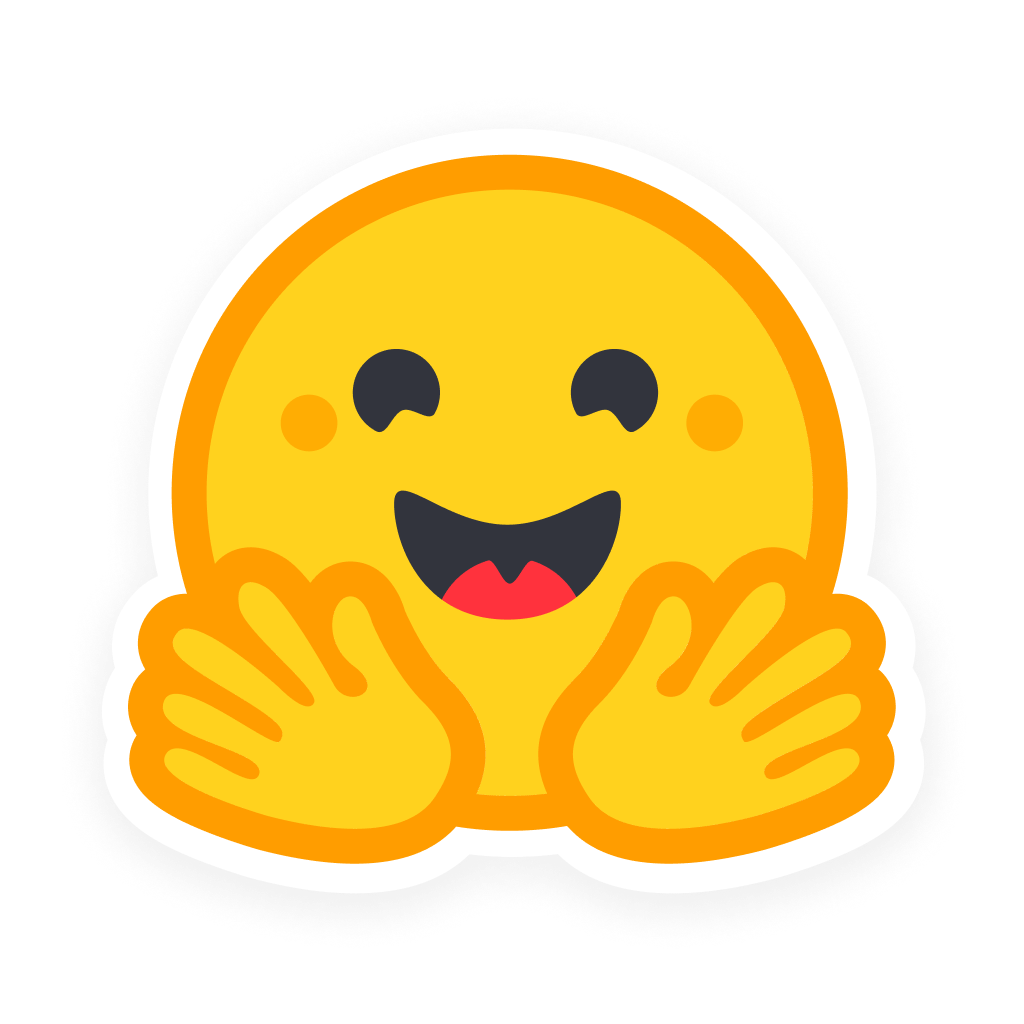}}}}

\title{Every Eval Ever: A Unifying Schema and \\ Community Repository for AI Evaluation Results}

\author{%
\begin{minipage}{\textwidth}
\centering
Jan Batzner\textsuperscript{*,1-3} \quad Sree Harsha Nelaturu\textsuperscript{*,4} \quad 
\\ Damian Stachura\textsuperscript{*,5} \quad Anastassia Kornilova\textsuperscript{*,6}
\\[1em]
Jon Crall\textsuperscript{\ensuremath{\diamond}, 7}  \quad Tommaso Cerruti\textsuperscript{\ensuremath{\diamond}, 8}  \quad Yanan Long\textsuperscript{\ensuremath{\diamond}, 9} \quad Yifan Mai\textsuperscript{\ensuremath{\diamond}, 10} \quad Sanchit Ahuja\textsuperscript{\ensuremath{\diamond}, 11} \quad  Asaf Yehudai\textsuperscript{\ensuremath{\diamond}, 12} \quad Marek Šuppa\textsuperscript{\ensuremath{\diamond}, 13,14} \quad John P. Lalor\textsuperscript{\ensuremath{\diamond}, 15} \quad Oluwagbemike Olowe\textsuperscript{\ensuremath{\diamond}, 16}
\\[1em]
Jatin Ganhotra\textsuperscript{12} \quad Brian H. Hu\textsuperscript{7} \quad Eliya Habba\textsuperscript{17} \quad Andrew M. Bean\textsuperscript{18} \quad Chang Liu\textsuperscript{19} \\ Sander Land\textsuperscript{20} \enspace Steven Dillmann\textsuperscript{10} \enspace
Aniketh Garikaparthi\textsuperscript{21} \enspace Elron Bandel\textsuperscript{12} \enspace Saki Imai\textsuperscript{11} \quad
James Edgell\textsuperscript{22} \quad Wm. Matthew Kennedy\textsuperscript{18} \quad Jenny Chim\textsuperscript{23} \quad  
Patrick Meusling\textsuperscript{24} \\ Asteria Kaeberlein\textsuperscript{11} \quad
Venkata Ramachandra Karthik Chundi\textsuperscript{16}  \quad Manasi Patwardhan\textsuperscript{21} \quad 
Martin Ku\textsuperscript{22} \quad
Austin Meek\textsuperscript{25} \quad Leon Knauer\textsuperscript{26} \quad 
Brian Wingenroth\textsuperscript{27} \quad Srishti Yadav\textsuperscript{28,29} \quad Usman Gohar\textsuperscript{30} \quad
Felix Friedrich\textsuperscript{31} \quad Michelle Lin\textsuperscript{32,33} \quad  Jennifer Mickel\textsuperscript{34} \quad
Arman Cohan\textsuperscript{35} 
\\[1em]
Stella Biderman\textsuperscript{\ensuremath{\dagger}, 34} \quad Irene Solaiman\textsuperscript{\ensuremath{\dagger}, 36} \quad  Zeerak Talat\textsuperscript{\ensuremath{\dagger}, 37} \quad Anka Reuel\textsuperscript{\ensuremath{\dagger}, 10,38} \quad\\ Mubashara Akhtar\textsuperscript{\ensuremath{\dagger}, 39,8} \quad 
Gjergji Kasneci\textsuperscript{\ensuremath{\dagger}, 1,2} \quad
Avijit Ghosh\textsuperscript{\ensuremath{\dagger}, 36} \quad Leshem Choshen\textsuperscript{\ensuremath{\dagger}, 40,41,12}
\\[1em]
{\vspace{-0.2cm}\footnotesize\normalfont\mdseries
    \noindent
    \begin{center}
    \textsuperscript{*}\,Lead Author \quad
    \textsuperscript{\ensuremath{\diamond}}\,Top Contributor \quad
    \textsuperscript{\ensuremath{\dagger}}\,Advisor
    \\
    This project was a part of the Evaluating Evaluations (EvalEval) Coalition: \raisebox{-0.2ex}{\includegraphics[height=1em]{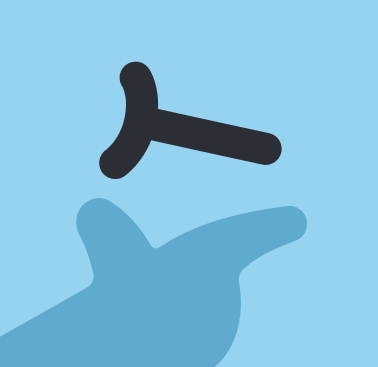}} \url{https://evalevalai.com/}
    \end{center}
}
\smallskip
{\small
\normalfont
\textsuperscript{1}Technical University Munich \quad
\textsuperscript{2}Munich Center for Machine Learning \quad
\textsuperscript{3}Weizenbaum Institute \\
\textsuperscript{4}Zuse Institute Berlin \quad
\textsuperscript{5}Evidence Prime \quad
\textsuperscript{6}Trustible \quad
\textsuperscript{7}Kitware \quad
\textsuperscript{8}ETH Zurich \quad
\textsuperscript{9}StickFlux Labs \quad
\textsuperscript{10}Stanford University \quad
\textsuperscript{11}Northeastern University \quad
\textsuperscript{12}IBM Research \quad
\textsuperscript{13}Comenius University Bratislava\quad
\textsuperscript{14}Cisco  \quad 
\textsuperscript{15}University of Notre Dame \quad
\textsuperscript{16}Independent \quad
\textsuperscript{17}Hebrew University of Jerusalem \\
\textsuperscript{18}University of Oxford \quad
\textsuperscript{19}Ohio University \quad
\textsuperscript{20}Writer \quad
\textsuperscript{21}TCS Research \quad
\textsuperscript{22}Oxford University Press \quad
\textsuperscript{23}Queen Mary University of London \quad
\textsuperscript{24}Technical University Berlin \quad
\textsuperscript{25}University of Delaware \quad
\textsuperscript{26}Cinemo \quad
\textsuperscript{27}Johns Hopkins University \quad
\textsuperscript{28}University of Copenhagen \quad
\textsuperscript{29}ELLIS \quad
\textsuperscript{30}Iowa State University \\
\textsuperscript{31}Meta FAIR \quad
\textsuperscript{32}University of Montreal \quad
\textsuperscript{33}Mila Quebec AI Institute \quad
\textsuperscript{34}EleutherAI \quad
\textsuperscript{35}Yale University \quad
\textsuperscript{36}Hugging Face \quad
\textsuperscript{37}University of Edinburgh \quad
\textsuperscript{38}Harvard University \quad
\textsuperscript{39}ETH AI Center \\
\textsuperscript{40}MIT \quad
\textsuperscript{41}MIT-IBM Watson Lab}
\end{minipage}
}
\begin{document}
\maketitle
\begin{abstract}
AI evaluations are widely used for testing and understanding progress. However, the diverse evaluators bring with them inconsistencies that challenge analysis and comparison.
First, results are saved in incompatible formats, scattered across leaderboards, papers, blog posts, evaluation harness logs, and custom repositories. Second, results are created by different evaluation frameworks, which produce divergent scores for nominally identical evaluations and record metadata inconsistently, hindering comparison, cross-community evaluation science, cost reduction, and reuse.
We introduce \textbf{\eee{}}, the first shared schema and community-crowdsourced repository for AI evaluation results. The schema standardizes how evaluations are represented, %capturing source provenance, model details, benchmark metadata, generation configuration, metric semantics, among others, 
in a unified, single JSON document. It is source-agnostic by design, ingesting results from evaluation harnesses and papers alike, and optionally stores per-instance outputs for fine-grained analysis. We contribute: (i)~a community-governed metadata schema with a companion instance-level schema ~\href{https://github.com/evaleval/every_eval_ever}{\raisebox{-0.15ex}{\faGithub}\,\texttt{evaleval/every\_eval\_ever}}, the first standardization effort of its kind; (ii)~automatic converters from popular formats, evaluation harnesses, and leaderboards to the unified schema~\githubrepo{https://github.com/evaleval/every_eval_ever} ; and (iii)~a crowdsourced community database hosted on Hugging Face, currently spanning to date 22,235 models, 2,273 unique benchmarks, and 31 evaluation formats ~\href{https://huggingface.co/datasets/evaleval/EEE_datastore}{\raisebox{-0.6ex}[0pt][0pt]{\includegraphics[height=1.3em]{figs/hf-logo.png}}\,\texttt{evaleval/EEE\_datastore}}.
\end{abstract}

\section{Introduction}
\label{sec:intro}
Evaluations are critical for measuring AI progress, yet how they are reported is inconsistent, incomplete, and difficult to interpret. Evaluation results are often reduced to aggregated scores in a table, with important evaluation metadata, such as generation parameters, evaluation settings, and data provenance, omitted or scattered across papers, ad hoc log files, and code repositories. This fragmentation undermines reproducibility, complicates cross-benchmark comparisons, and limits the potential for systematic meta-analysis.

\begin{figure}[h!]
    \centering
    \includegraphics[width=\columnwidth]{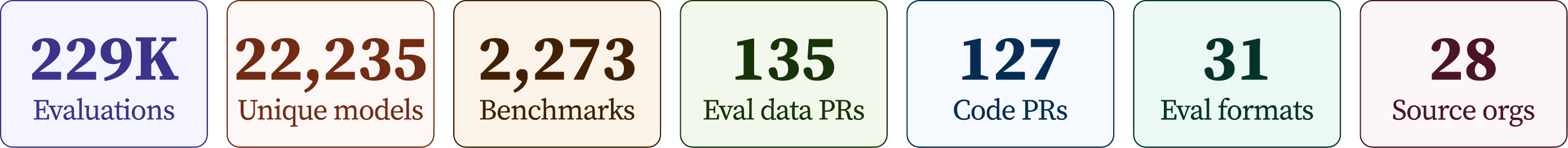}
    \caption{Overview of the scale and diversity for \eee{} data.}
    \label{fig:EEE_stats}
\end{figure}

In practice, this creates fundamental challenges for both researchers and practitioners. Comparative evaluation studies are typically constrained by the subset of results that can be reliably reproduced (e.g., architecture scaling \citep{choshen2025hitchhiker,ruan2024observational} or quantization comparisons \citep{kurtic2025give}), often requiring substantial computational and financial resources \citep{ghosh2026evalbottleneck, perlitz-etal-2024-efficient}. A lack of comparability is especially misleading when different parties evaluate the same model or benchmark, yet produce different scores \citep[see \S\ref{sec:case3} and][]{yuanunderstanding,fireworks2025kimik2p5}. For example, the LLaMA~65B model has been reported to achieve both 63.7 and 48.8 on MMLU~\citep{mmlu}. On a closer look, the difference in scores was found to arise from the use of different evaluation harnesses. Without this context, the scores are not directly comparable \citep{fourrier2023openllmleaderboard}. Similarly, our analysis of evaluations across over 22,235 models and 2,273 benchmarks reveals 31 distinct reporting formats, highlighting the lack of standardization and motivating the need for more structured reporting practices (See statistics in Fig.~\ref{fig:EEE_stats}). 

Other parts of the AI pipeline have benefited from standardization: shared metadata schemas such as DCAT, Schema.org/Dataset, and Croissant \citep{w3c_dcat3_2024,schemaorg_dataset,croissant2024}; documentation practices such as Datasheets for Datasets and Model Cards \citep{gebru2021datasheets,mitchell2019modelcards}; and common evaluation and benchmarking protocols such as GLUE, SuperGLUE, HELM, BIG-bench, and MLPerf \citep{wang2019glue,wang2019superglue,helm,srivastava2023imitationgamequantifyingextrapolating,reddi2020mlperf} have improved reproducibility, comparability, and transparency. In contrast, evaluation reporting remains fragmented~\citep{liao2021are,bowman-dahl-2021-will,raji2021ai,ethayarajh-jurafsky-2020-utility,burnell2023rethink}, with reported implications for downstream analysis such as benchmark saturation studies \citep{benchmark_saturation,akhtar2026aibenchmarksplateausystematic}. 
Similarly, psychometric analyses in the field depend on standardized example-level data, which is rare in current evaluation reporting \citep{li2026adaptivetestingllmevaluation,polo2024tinybenchmarksevaluatingllmsfewer}. Finally, governance frameworks such as the EU AI Act~\citep{eu_ai_act} mandate reproducible risk assessments, yet current evaluation tooling and reporting lack even the basic standardization that reproducibility requires.

\begin{figure}[h]
    \centering
    \includegraphics[width=\columnwidth]{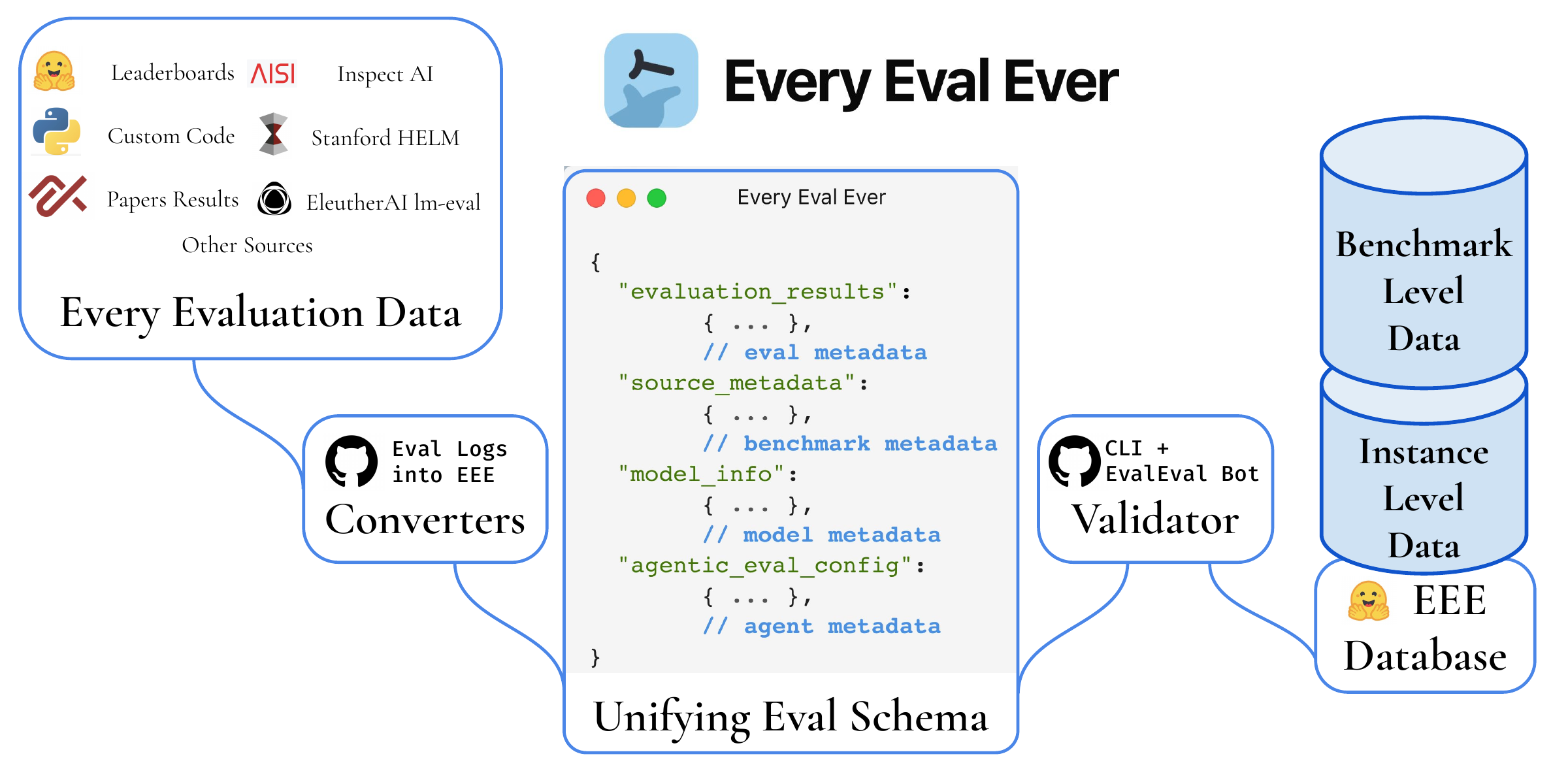}
    \caption{\eee{} has four components: (1) heterogeneous evaluation data (leaderboards, papers, harness logs, custom scripts); (2) converters for known log formats (HELM, Inspect AI, lm-eval) and metadata parsers for community formats (Hugging Face, leaderboards); (3) a unified metadata schema supporting aggregate and instance-level results; and (4) a crowdsourced community database making public evaluation results accessible and processable.}
    \label{fig:overview_EEE_components}
\end{figure}

\eee{} (\eeeshort{}) addresses these gaps through a shared reporting schema and a crowdsourced repository for AI evaluation results. Just as data~\citep{croissant2024} and models~\citep{mitchell2019modelcards} have documentation standards, \eeeshort{} standardizes the core aspects of evaluation:  who ran it, under what settings, and what the resulting scores mean. It ingests results from any source, like harness logs, leaderboard scrapes, and paper results, and represents them in a single, interoperable format.

In summary, \eeeshort{} makes the following contributions:
\begin{enumerate}

    \item A \textbf{shared, versioned JSON schema} for AI evaluation results that captures source provenance, model access mode, generation configuration, and metric semantics in a single record, with an optional instance-level companion schema supporting single-and multi-turn interaction types.

    \item \textbf{Automatic converters} from major harnesses (HELM, lm-eval-harness, Inspect AI) and common formats producing schema-compliant records, including per-instance outputs where source logs provide them, paired with a validation pipeline that ensures schema compliance at contribution time.

    \item A \textbf{crowdsourced, community repository} hosted on Hugging Face, already spanning 22,235 models, 2,273 unique benchmarks, and 31 evaluation formats, that for the first time enables cross-framework comparison of evaluation results at scale.

    \item Exemplary \textbf{empirical analyses enabled by unified repository}, where \eeeshort{} can identify cost-accuracy tradeoffs in agentic evaluations (\ref{case1}), reveal implementation-dependent perplexity scores (\ref{case2}), captures evaluation harness reproducibility gaps (\ref{sec:case3}), and enable meta-analysis using Item Response Theory (\ref{case4}), none of which were previously feasible without a unified result format.
\end{enumerate}

\section{Related Work}
\label{sec:related}
\paragraph{Evaluation harnesses:} Evaluation harnesses describe software to standardize model evaluation, from input prompts to output metrics.
While evaluation harnesses like lm-eval-harness~\citep{eval_harness}, HELM~\citep{helm}, and InspectAI~\citep{inspectAI} have proliferated, their format for results remain mutually incompatible~\citep{bandel2024unitxt,biderman2024lessonstrenchesreproducibleevaluation}. \eee{} is not a new evaluation harness, but a translation layer that sits above those and enables better aggregation of evaluation results.

\paragraph{Evaluation sharing:} There are a few large sources that share evaluations. The main sources for those are leaderboards  \citep{helm,kapoor2025holisticagentleaderboardmissing}, or websites \citep{artificialanalysis2026,metr_blog,epoch_ai_about} efforts that release what they run, and two concurrent works to ours that collect instance-level \citep{jiang2026positionscienceaievaluation} or Inspect framework outputs specifically \citep{abbas2025developing} and share them publicly. We collaborate with them to aggregate their results to \eee. Public Leaderboards like Open LLM Leaderboard~\citep{openllm_leaderboard}, Chatbot Arena~\citep{chatbot_arena},  AlpacaEval~\citep{alpacaeval}, MT-Bench~\citep{zheng2023mtbench}, aggregated results at scale but export limited structured metadata \citep{wang2024benchmark}. We created \eee{} to combine all of those scores in a unified format and database, alongside local harness runs within the same format.

\paragraph{Reproducibility:} Comparison is unreliable when different evaluation settings are underspecified and carry the same benchmark name. Lacking standards prevents the community from reliably comparing, replicating, and reusing cost-intensive evaluations \citep{biderman2024lessonstrenchesreproducibleevaluation}. The same model, accessed through different providers or run with different engine configurations, can produce different outputs \citep{nelaturu2024hardware}. Moreover, prompt ordering~\citep{prompt_sensitivity} and data contamination~\citep{benchmark_contamination} can introduce score variance. Large-scale analysis of evaluation results can require weeks of data wrangling before any research can begin \citep[e.g.][]{ruan2024observational,choshen2025hitchhiker,perlitz-etal-2024-efficient,akhtar2026aibenchmarksplateausystematic}, if such analysis is possible at all without rerunning full leaderboards at extreme cost \citep[e.g.][]{habba2025dove,perlitz-etal-2024-efficient,ghosh2026evalbottleneck}, we estimate the inference cost to reproduce our  data in \S\ref{sec:stats}.  

\paragraph{Dataset and model documentation:} Although larger efforts in the ML community centered around datasets and model documentation, evaluation, and result documentation itself remain a gap in the community \citep{liao2021are,bowman-dahl-2021-will}; where multiple suggestions for metadata to report exist \citep{staufer2025audit,bordes2025evalfactsheetsstructuredframework,sokol2025benchmarkcardsstandardizeddocumentationlarge}, but not the low-level evaluation ones. For datasets, Datasheets for Datasets~\citep{gebru2021datasheets} and Croissant~\citep{croissant2024} standardize metadata for ML datasets. For benchmarks, those efforts have been tailored to benchmark needs \citep{reuel2024betterbenchassessingaibenchmarks, hofmann2025autobenchmarkcardautomatedsynthesisbenchmark,sokol2025benchmarkcardsstandardizeddocumentationlarge}. For models, Model Cards~\citep{mitchell2019modelcards, liu-etal-2024-automatic} document artefacts and their intended uses. For evaluation results, \eee{} addresses the most pressing remaining gap: a shared schema for the run-time context that determines whether two scores can be aggregated and compared.

\paragraph{Agentic evaluation standardization:} Recent work begun to understand the importance of standardizing agentic evaluation~\citep{bandel2026readyforgeneral, kapoor2025holisticagentleaderboardmissing}, and made first steps towards achieving it~\citep{bandel2026generalagentevaluation, lacoste2026cubestandardunifyingagent, merrill2026terminalbenchbenchmarkingagentshard, Harbor_Framework, yehudai2026agenticclearautomatingmultilevel}. These efforts focus on the runtime and execution layers, standardizing agent evaluation across task representation, environment type, interface protocol, and tool specification format to enable easy and scalable agent and benchmark integration. \eee{} provides a complementary focus by standardizing how agentic evaluation results are represented and stored. Hence, allowing for easy results analysis across different sources (Section \ref{case1}).
% General Agent Evaluation introduces a unified protocol and the Exgentic framework for evaluating general-purpose agents across heterogeneous environments \citep{bandel2026generalagentevaluation}. CocoaBench complements this line of work with a benchmark of 153 long-horizon tasks that require the composition of vision, search, and coding capabilities, together with system-level runtime and cost analysis over industry-leading agent implementations \citep{cocoabenchteam2026cocoabenchevaluatingunifieddigital}.  These efforts focus heavily on the runtime and execution layers;

\begin{table}[t]
\caption{Design decisions behind \eee{} and the capabilities they enable.}
\label{tab:schema-design-decisions}
\centering
\small
\setlength{\tabcolsep}{4pt}
\renewcommand{\arraystretch}{1.12}
\begin{tabular}{c p{0.23\columnwidth} p{0.36\columnwidth} p{0.30\columnwidth}}
\toprule
& \textbf{Design Decision} & \textbf{Implementation in \eee{}} &
\textbf{Enabled capability} \\
\midrule
\rowcolor{gray!10} \faPuzzlePiece & Accept partial records &
Required fields are minimal; unavailable metadata can be omitted or recorded in
\texttt{additional\_details}. &
Includes existing results that would be excluded by a fully specified reporting
standard. \\
\addlinespace[4pt]
\faDatabase & Use wide, typed metadata coverage &
Common fields capture model identity, provenance, generation settings,
benchmark metadata, and metric semantics. &
Makes heterogeneous results comparable and auditable across papers,
leaderboards, and logs. \\
\addlinespace[4pt]
\rowcolor{gray!10} \faFingerprint & Preserve run identity without a canonical fingerprint &
Each run receives a UUID and stores available context, rather than assuming a
fixed parameter tuple uniquely identifies an evaluation. &
Keeps repeated, conflicting, or underspecified runs visible for later
deduplication and longitudinal analysis. \\
\addlinespace[4pt]
\faSeedling & Version and extend the schema &
New versions add fields for new modalities, reasoning models, and agentic
traces without rewriting old records. &
Keeps the schema forward-compatible.\\
\addlinespace[4pt]
\rowcolor{gray!10} \faMicroscope & Separate aggregate and instance-level records &
Aggregate JSON files link to optional sidecar files with prompts, outputs,
scores, and metadata. &
Enables instance-level reanalysis, auditing, and meta-evaluation beyond
headline scores. \\
\addlinespace[4pt]
\faCubes & Organize reusable information blocks &
Source, model, configuration, result, and instance-level metadata are grouped
into recurring schema components. &
Improves interoperability across converters and downstream analysis tools. \\
\bottomrule
% \vspace{0.1mm}
\end{tabular}
\end{table}

\section{The \textit{Every Eval Ever} Schema} 
\label{sec:schema}
\eee{} is a standardized representation of AI evaluation results across benchmarks, models, and reporting data sources (e.g., public model leaderboards, research papers, evaluation harness logs, among others; Figure \ref{fig:overview_EEE_components}). Instead of storing only final performance scores, each \eee{} record captures the metadata required to interpret, compare, and reuse results: Who ran the evaluation, which model was evaluated, under what generation settings, how metrics were computed, and (if available) instance-level outputs. The schema is modular and organized into reusable information blocks. In this section, we describe how the schema was developed and the design principles that guided its construction (Section~\ref{sec:design-process}), %outline the design principles that guided
%its construction (Section~\ref{sec:schema-design}), 
and present the schema
structure and core components (Section~\ref{sec:schema-source}); full field
references appear in App.~\ref{sec:app-schema} and
Table~\ref{tab:aggregate_evaluation_record}.

\subsection{Schema design principles and development methodology}
\label{sec:design-process} 
\eee{} balances broad adoption with enough structure to support downstream
comparison, auditing, and reanalysis. Table~\ref{tab:schema-design-decisions}
summarizes the main design decisions and the capabilities they enable.
% BH: Moved the previous design principles subsection here to prevent a very small subsection
The schema was developed through an open, iterative community design process inspired by the Croissant metadata format~\citep{croissant2024}. We gathered structured feedback from about 40 researchers and unstructured feedback from about 110 researchers, including benchmark creators, evaluation framework developers, governance experts, leaderboard operators, and industry practitioners. The schema is open and had since discussions on improvements through GitHub and Slack. The schema versions were openly proposed, discussed, and revised by the community, with disagreements resolved by consensus meeting among core contributors (see Governance in App.~\S\ref{app:governance}). The fields were included if they were (i)~reported in at least one existing evaluation framework or published result and (ii)~considered necessary for the interpretability or reuse of the score by the majority of contributors, (iii) anticipated to be available for others to report in the future; fields that did not meet all criteria were moved to \texttt{additional\_details} or excluded. 
\subsection{Schema overview}
Each \eeeshort{} record stores data on a single evaluation run. The \eeeshort{} schema is organized into five metadata blocks: \\ First, \keytermA{Source Metadata:} Who produced the results and from where did it originate? \\ Second, \keytermB{Model Information:} Which model was evaluated and how was it accessed? \\ Third, \keytermC{General Configuration:} Which configuration settings were used during evaluation? \\ Fourth, \keytermD{Evaluation Results:} How were the results reported (e.g., metrics, uncertainty estimates)? \\ Finally, \keytermE{Instance-level Data:} Optionally, what instance-level information is available?

\begin{sidemarkA}
\paragraph{\keytermA{1. Source metadata:}}
\label{sec:schema-source}
The \texttt{source\_metadata} field records who produced the result and how it was collected. \texttt{source\_type} distinguishes results scraped from a leaderboard or paper (\texttt{documentation}) from those produced by a local evaluation run
(\texttt{evaluation\_run}). \texttt{evaluator\_relationship} records whether the evaluation was run by the model developer (\texttt{first\_party}), an independent party (\texttt{third\_party}), or the metadata contributors themselves (\texttt{self}). Capturing the reporting source is important since the incentives, reproducibility, and trustworthiness of the reported results can differ between them.
\end{sidemarkA}

\begin{sidemarkB}
\paragraph{\keytermB{2. Model information and access mode:}}
\label{sec:schema-model}
\texttt{model\_info} records the model identifier in \texttt{developer/name} using a standardized developer/model naming convention and the \emph{access mode}.
We store whether results were obtained through hosted APIs such as \texttt{openai} and \texttt{anthropic} (\texttt{inference\_platform}) or local inference engines like vLLM (\texttt{inference\_engine}).
% \texttt{inference\_platform} covers remote APIs (e.g.\ \texttt{openai}, \texttt{anthropic}); \texttt{inference\_engine} covers local runtimes (e.g.\ vLLM with a pinned version). 
The same weights served through different providers or engine versions can produce different results (see \S\ref{sec:case3}); recording access mode makes these hidden confounds visible.
\end{sidemarkB}

\begin{sidemarkC}
\paragraph{\keytermC{3. Generation configuration:}}
\label{sec:schema-gen}
Parameters such as temperature, number of samples, and stop sequences can effect benchmark outcomes substantially, yet they are frequently missing from leaderboards. \texttt{generation\_config} makes them first-class fields. When a parameter is unknown, the field is omitted and its absence is recorded explicitly rather than silently defaulted.
\end{sidemarkC}

\begin{sidemarkD}
\paragraph{\keytermD{4. Evaluation results and metric semantics:}}
\label{sec:schema-results}
\texttt{evaluation\_results} stores one entry per scored metric. Each entry includes a \texttt{metric\_config} object capturing score direction (\texttt{lower\_is\_better}), type (continuous, binary, ordinal), and range. This prevents silent ambiguity. For example, a score of 0.31 is favorable on toxicity metrics where lower is better, but poor on pass@1 coding metrics where higher scores are desirable.
Ordinal metrics (e.g.\ Low/Medium/High mapped to integers via \texttt{level\_names}), uncertainty fields (standard errors, confidence intervals), and per-result timestamps are also supported (Case Studies \S\ref{case2}).
\end{sidemarkD}

\begin{sidemarkE}
\paragraph{\keytermE{5. Instance-level data:}}
\label{sec:schema-instance}
While aggregate scores support comparisons, understanding \emph{why} scores differ often requires per-sample data \citep{burnell2023rethink}. \eeeshort therefore stores instance outputs in a optional companion file \texttt{\_samples.jsonl} (one JSON object per line). This file can store prompts, model outputs, references, scores, and metadata needed for detailed analysis. Three interaction types are supported:
First, \texttt{single\_turn} — QA, MCQ, classification; uses an  \texttt{output} object. Second, \texttt{multi\_turn} — multi-exchange conversations; uses a   \texttt{messages} array. Third, \texttt{agentic} — tool-using agents with full tool-call traces and sandbox logs; uses \texttt{messages} with nested \texttt{tool\_calls}. For agentic evaluations (e.g.\ SWE-Bench, GAIA), the aggregate record captures tool and sandbox configuration (Case Study \S\ref{case1}).
\end{sidemarkE}
\section{Converters and validation}
\label{sec:infra}
\eee{} schema is supported by two components: (1) converters, which automatically parse existing evaluation outputs into the \eeeshort{} schema, and (2) automated validation, which assesses whether the submitted datasets conform to the \eeeshort{} schema specification:

\paragraph{(1) Converters:}
\label{sec:converters}
Manual re-formatting of existing logs is a large barrier to adoption.
Hence, we provide converters for three widely-used LLM evaluation frameworks:
  HELM~\citep{helm}, lm-eval-harness~\citep{eval_harness}, and Inspect
  AI~\citep{inspectAI}.
App.~\ref{sec:app-converters} describes these converters input formats and
  field mappings.
More converters from leaderboards or other sources were contributed by the
  community (App.~\ref{sec:app-comm}).
Each converter produces a schema-compliant \texttt{.json} file and, where source
  logs include per-sample data, a \texttt{_samples.jsonl}.

\paragraph{(2) Validation:}
\label{sec:validation}
Every time a record is submitted to \eeeshort{}, it is validated against the schema before being entered into the repository. The validation step checks for the required fields, data types, enum constraints, and object consistency. Schema compliance is enforced by Pydantic~v2 models generated from \href{https://github.com/evaleval/every_eval_ever/blob/dec1ae43e0741a37003425eafe6699d3296145ec/every_eval_ever/schemas/eval.schema.json}{\texttt{eval.schema.json}}. Validation runs in two settings: 

 \textbf{(2.1) Locally via the command-line:} Users can validate records locally with the \eee{} CLI before submission. However, there are additional checks available which are provided by the validator. The CLI supports rich terminal, JSON, and GitHub annotation output formats. 
 
 \textbf{(2.2) EvalEvalBot:} When contributors submit data via pull request to the datastore, either the author or a maintainer can request validation using the validator. The validator checks schema syntax compliance, presence of all files mentioned and provides warning for situations presence of duplicate records. 

\section{Community approach}
\textbf{Governance:} 
The community-driven nature of \eeeshort{} necessitates that governance is integrated into \eeeshort{}. The project recognizes three roles: \emph{core maintainers}, \emph{contributors}, and \emph{community reviewers}, where maintainers hold final authority on contested decisions (see App.~\ref{app:governance-decision} for more). Decisions range from routine record additions to substantive schema changes, the latter following a structured community proposal and review process (App.~\ref{app:governance-proposal}).                                 
Records are immutable once accepted. Errors are handled via explicit correction and retraction mechanisms that preserve immutability and reproducibility (App.~\ref{app:governance-corrections}). For example, discrepancy between evaluation results for LLaMA \citep{fourrier2023openllmleaderboard} exists and \eeeshort{} stores both evaluation results as valid records, with the discrepancy visible in metadata rather than discovered through a blog post (App.~\ref{app:governance-conflicts}, ~\ref{app:governance-example1}). Full details, including worked examples, are in App.~\ref{app:governance}.

\textbf{Contribution model:} Evaluation infrastructure is a public good: every stakeholder needs it, yet no single entity has sufficient incentive to build it alone. \eeeshort{} addresses this collectively: contributions range from evaluation records (aggregate JSON files from converters, instance-level companion files, or leaderboard scrapes; (Section~\ref{sec:converters}) to converter extensions, schema proposals, and tooling (Section~\ref{sec:schema}). Benchmark creators can register artifacts in \eeeshort{} format, gaining visibility in downstream use. Evaluation framework developers (e.g., HELM~\citep{helm}, Inspect AI~\citep{inspectAI}) get a standardized export path for their users. Leaderboard operators can offload comparison and cross-source aggregation to shared infrastructure. Evaluation researchers (see Section~\ref{sec:casestudies} for examples) gain access to otherwise unattainable meta-evaluation data. Significant contributions are recognized through co-authorship and formal industry partnerships (See App.~\ref{app:governance-code}).

\section{Data analytics of the \eee datastore}\label{sec:stats}

As of May 4th 2026, current contributions total more than 200K aggregated results across over a hundred data contributions (see summary statistics in App.~\ref{app:stats}), providing a foundation for evaluation research and a lens on community-wide reporting trends. In this section, we highlight how the cost of running evaluations underscores the value of a shared resource like \eeeshort{}, and what the collected data reveals about community-wide evaluation trends.

\paragraph{Cost of AI evals:}   Several works have discussed the cost of evaluating models \citep{perlitz-etal-2024-efficient,ghosh2026evalbottleneck}; while future work should make cost more explicitly derivable from the schema, it remains underreported and difficult to infer. Towards thism, we provide a conservative estimate of the savings such a shared resource may offer. We conservatively estimate that reproducing just the evaluation runs currently collected in \eeeshort{}, would cost hundreds of thousands of dollars (App.~\ref{app:cost}). This figure considers only running costs and excludes factors that would raise it by further orders of magnitude e.g., agentic evaluations (Case Study \S\ref{app:casestudies1}), thinking models, repeated runs, failed attempts, long benchmarks, code execution, and human labeling \citep{bandel2026agentic,kapoor2025holisticagentleaderboardmissing}.

\paragraph{Community trends:} Beyond its value as a shared resource, the corpus offers a data-rich view of community-wide evaluation practices. While acknowledging that coverage is biased by data availability, we analyze what the collected results reveal about how the field approaches AI evaluations (App.~\ref{app:stats}). We find that evaluations follow a long tail: popular benchmarks and models are reported at a scale far above the rest, yet thousands of less common ones appear, and the top 25 in each category account for barely 25\% of all results. Geographically, we observe a strong concentration on U.S.-based models, with GPT models dominating. Excluding human baselines, five companies account for 23 of the 24 most frequently evaluated systems, revealing a focus not only on specific models but on specific sources of models. This concentration carries implications beyond socio-political findings and suggests that much of the field is evaluating commercial products rather than underlying technologies, confirming recent claims in the literature \citep{michaelov2026open}.

\definecolor{evalblue}{HTML}{5BACD1}
\renewcommand{\arraystretch}{1.2}
\setlength{\tabcolsep}{5pt}
\small
\begin{minipage}{400pt}
\captionof{table}{Macro-average fill rate of \eee metadata fields across the 31 evaluation harnesses and formats in the \eeeshort{} datastore. We provide three metadata field examples each for model metadata, benchmark metadata, and evaluation metadata.}
\vspace{2pt}
\resizebox{\textwidth}{!}{%
\begin{tabular}{l r r @{\hspace{8pt}} l r r @{\hspace{8pt}} l r r}
\toprule
\multicolumn{3}{c}{\textbf{Model}} & \multicolumn{3}{c}{\textbf{Benchmark}} & \multicolumn{3}{c}{\textbf{Evaluation}} \\
\cmidrule(lr){1-3} \cmidrule(lr){4-6} \cmidrule(lr){7-9}
\rowcolor{gray!10} model name & \textcolor{evalblue}{\rule{28.0pt}{6pt}} & 100\% & source URL / HF repo & \textcolor{evalblue}{\rule{26.8pt}{6pt}} & 96\% & inference platform & \textcolor{evalblue}{\rule{7.7pt}{6pt}} & 27\% \\
model parameters & \textcolor{evalblue}{\rule{0.9pt}{6pt}} & 3\% & metric ID & \textcolor{evalblue}{\rule{21.7pt}{6pt}} & 77\% & temperature & \textcolor{evalblue}{\rule{6.4pt}{6pt}} & 23\% \\
\rowcolor{gray!10} model license & \textcolor{evalblue}{\rule{0.9pt}{6pt}} & 3\% & uncertainty n\_samples & \textcolor{evalblue}{\rule{8.3pt}{6pt}} & 30\% & max tokens & \textcolor{evalblue}{\rule{6.4pt}{6pt}} & 23\% \\
\bottomrule
\end{tabular}%
\label{tab:fill-rate-metadata}
}
\end{minipage}
\normalsize
\paragraph{Format inconsistencies:} Analyzing the data also corroborates our claims about format inconsistencies. For example, the common source of evaluations is academic papers, which are not machine-readable, and each uses a different reporting format. Moreover, many fields crucial for comparisons are unreported; for instance, the \emph{inference platform} is either explicitly marked as unknown or omitted entirely in 98\% of all evaluation rows (micro-average); even when each of the 31 formats is weighted equally, the field is reported in only 27\% of rows on average (macro-average; Table~\ref{tab:fill-rate-metadata}).

\normalsize
\section{Case Studies}
\label{sec:casestudies}
\eeeshort{} data allows AI evaluations to be compared, reproduced, and reused, enabling broader impact through additional research and meta-analyses. For example, \eeeshort{} data can help identify where the evaluation ecosystem is thin, which capabilities are over-measured, and which risks are neglected. With instance-level data, researchers can move beyond leaderboard averages to study item difficulty, robustness, and temporal drift. \eee{} also enables meta-evaluation: testing evaluation methods themselves to distinguish real progress from artifacts of setup and reporting. While different works already showcased uses for \eeeshort{}-like data \citep[e.g., for efficient benchmarking;][]{perlitz-etal-2024-efficient}, or even already used \eeeshort{} data \citep[e.g., to characterize benchmark saturation;][]{akhtar2026aibenchmarksplateausystematic}, we perform several initial studies to showcase research that \eeeshort{} enables.

\subsection{Case Study 1: \eee identifies cost–accuracy tradeoffs in agentic evaluation}
\label{case1}
We show \eeeshort{} is useful for analyzing agentic evaluations beyond accuracy scores. \eeeshort{} already contains results from several agentic benchmarks, including SWE-bench~\citep{jimenez2024swebenchlanguagemodelsresolve}, HAL~\citep{kapoor2025holisticagentleaderboardmissing}, Exgentic~\citep{bandel2026generalagentevaluation}, and CocoaBench~\citep{cocoabenchteam2026cocoabenchevaluatingunifieddigital}. \eeeshort{} also reports diverse metadata following previous work arguing that agent evaluations should track time, cost~\citep{kapoor2024aiagentsmatter, yehudai2025surveyevaluationllmbasedagents}, and other agent metadata~\citep{bandel2026agentic}, rather than reporting accuracy alone.
 We rely on the additional metadata to reveal cost-performance tradeoffs in scaffold and backbone choice.% shifts the question from maximizing score to identifying Pareto-optimal scaffold–backbone combinations.

\begin{figure}[h]
  \centering
  \begin{subfigure}[t]{0.49\textwidth}
    \centering
    \includegraphics[width=\linewidth]{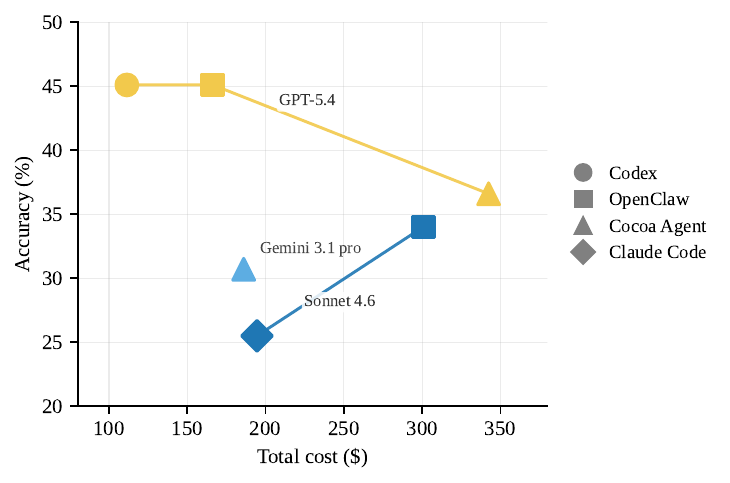}
    \caption{CocoaBench.}
    \label{fig:agent_case1_cocoabench}
  \end{subfigure}
  \hfill
  \begin{subfigure}[t]{0.49\textwidth}
    \centering
    \includegraphics[width=\linewidth]{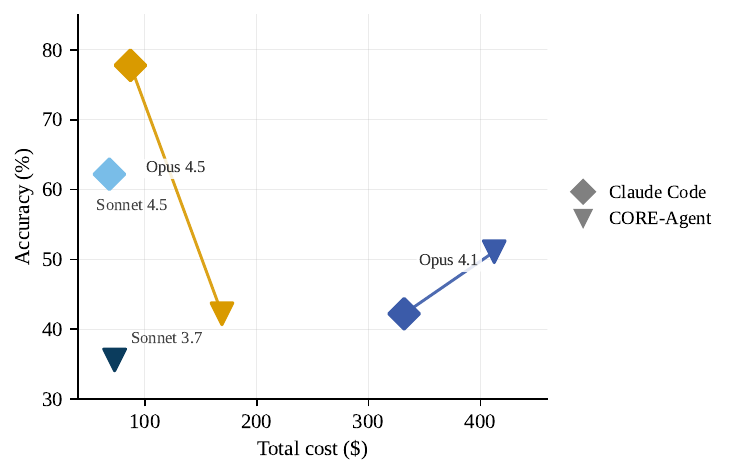}
    \caption{CORE-Bench Hard from HAL.}
    \label{fig:agent_case1_corebench}
  \end{subfigure}

  \caption{\eee{} enables cost--accuracy analysis across agent scaffolds and model backbones. Marker shape denotes the scaffold, and color denotes the backbone, with each point corresponding to one scaffold--backbone pair. Segments connect results sharing a backbone, isolating scaffold effects.}
  \label{fig:agent_case1}
\end{figure}

%\begin{figure}[h]
  %\centering
  %\includegraphics[width=0.8\textwidth]{figs/case1-agent_pareto.pdf}
    %\caption{\eee{} enables cost--accuracy analysis across agent scaffolds and model backbones. Segments connect results sharing a backbone, isolating scaffold effects.}
    %\caption{\eee{} enables cost--accuracy analysis across agent scaffolds and model backbones. Marker shape denotes the scaffold, and color denotes the backbone, with each point corresponding to one scaffold--backbone pair. Segments connect results sharing a backbone, isolating scaffold effects. Left: CocoaBench. Right: CORE-Bench Hard from HAL.}
  %\label{fig:agent_case1}
%\end{figure}

Fig.~\ref{fig:agent_case1} illustrates two concrete findings about cost--accuracy tradeoffs. 
First, CocoaBench \citep{cocoabenchteam2026cocoabenchevaluatingunifieddigital} shows that scaffold choice has substantial implication to costs, without necessarily showing performance gains in return. For example, Codex and OpenClaw with GPT-5.4 reach the same reported accuracy, but Codex costs less and is also faster on average (Appendix~\ref{app:casestudies1}).
Second, CORE-Bench Hard \citep{siegelcore} from HAL \citep{kapoor2025holisticagentleaderboardmissing} shows that scaffold effects can depend on the model backbone. Claude code is cheaper than CORE-Agent for both Opus 4.5 and 4.1, but it substantially increases accuracy for one and decreases for the other. 

% We utilize the CocoaBench results in the \eee{} schema to show why such metadata matters. Table~\ref{tab:cocoabench} shows that Codex and OpenClaw with GPT-5.4 tie for the best reported accuracy at 45.1\%, yet Codex averages 377.8 seconds and \$0.7 per task, compared with 502.1 seconds and \$1.0 for OpenClaw (see Appendix~\ref{app:casestudies1}). This demonstrates that Codex is 25\% faster and 30\% cheaper per task than OpenClaw, while both outperform Cocoa Agent. Within CocoaBench, these results show that GPT-5.4 is the strongest and most consistent model backbone across the various agent scaffolds.

% Relying on more metadata fields, we inspect the agent scaffold and model backbone effects within the same evaluation family. As all benchmarks follow the same schema, we broaden our analysis also to HAL's larger collection of agentic benchmarks. Fig~\ref{fig:agent_case1} (see more in App.~\ref{app:casestudies1}): within Claude Code, performance rises from 42.2\% with Claude Opus 4.1 to 62.2\% with Claude Sonnet 4.5 and 77.8\% with Claude Opus 4.5. Comparing Claude Code with CORE-Agent on shared model backbones such as Claude Opus 4.1 and Claude Opus 4.5 further shows that the agent scaffold choice also significantly affects performance.

Taken together, we find agentic evaluations cannot be interpreted from scalar accuracy alone: agent scaffold and model backbone choice, runtime, and cost all significantly affect the conclusions one draws from a result. This kind of cross-source decomposition is precisely what \eeeshort{} enables: without a common schema, these factors are scattered across incompatible logs and leaderboards, making systematic reanalysis challenging, particularly when metadata is reported inconsistently across sources. 

\subsection{Case Study 2: \eee reveals version-dependent perplexity}
\label{case2}
Model compression techniques~\citep{frantar2023optq,frantarsparseGPT,sun2023a} aim to reduce the size and computational cost of models, while minimizing degradation in performance.
WikiText perplexity \citep{merity2016pointersentinelmixturemodels} is a widely used metric to assess the impact of model compression, where lower perplexity indicates better predictive performance. However, reported values across papers for the same model and dataset can differ substantially based on implementation choices that often go unreported. %Works such as 
GPTQ \citep{frantar2023optq} and SpinQuant \citep{liu2025spinquantllmquantizationlearned} shipped model-specific evaluation scripts that %became standard in the compression literature for comparing against the original results. The 
report perplexity %perplexity reported by them is 
normalized by the number of tokens. %, which is model-dependent. 
In contrast, the LM Evaluation Harness~\citep{eval_harness} %, widely used for evaluating language models, also supports perplexity computation, but
reports \texttt{byte\_perplexity}, \texttt{word\_perplexity}, and \texttt{bits\_per\_byte} rather than token-level perplexity. % used by the GPTQ and SpinQuant scripts. 
``Perplexity'' alone is ambiguous as the same loss normalized by tokens, words, or bytes yields different numbers that are not directly comparable, as shown in Table~\ref{tab:ppl_comparison}. 

\begin{wraptable}{r}{0.61\textwidth}
\vspace{-0.5cm}
\centering
\small
\begin{tabular}{lccc}
% \toprule
 & GPTQ script & vLLM + \texttt{lm-eval} &   \\
 & Token PPL & Word PPL & Mismatch \\
Model & $e^{(\mathcal{L}_{\text{sum}} / N_{\text{tokens}})}$ & $e^{(\mathcal{L}_{\text{sum}} / N_{\text{words}})}$ & \\
 
% \midrule
OPT-6.7B   & 10.8605 & 12.2907 & \cellcolor{deltahl}1.4301  \\
Llama-2-7B &  5.4687 &  8.7939 &  \cellcolor{deltahl}3.3252  \\
% \bottomrule
% \vspace{0.01mm}
\end{tabular}
\caption{Perplexity on WikiText under two evaluation implementations. The summed cross-entropy $\mathcal{L}_{\text{sum}}$ is identical across columns; only the normalization denominator differs. Yet, the resulting values are not directly comparable.}
\label{tab:ppl_comparison}
\end{wraptable}

\eeeshort{} makes these distinctions explicit. Recording the evaluation backend, dataset version, and normalization convention prevents results from being compared %as if identical 
merely because they share the ``perplexity'' label, and helps avoid drawing incorrect conclusions when, for example, a vLLM-based evaluator reports a different variant from the GPTQ-style script needed for direct comparison (see App.~\ref{app:casestudies2} for implementation details).
% \vspace{2pt}

\subsection{Case Study 3: \eee captures reproducibility gaps}
\label{sec:case3}

We use \eeeshort{} to audit instance-level reproducibility. Although evaluation frameworks do not usually promise exact reproducibility, researchers often rerun public evaluations locally and compare them to shared results. We reproduced three models on fourteen single-turn HELM benchmarks and compared aligned per-instance scores between official HELM-released records~\citep{helm}
and local reproductions after converting both sides to \eeeshort{}. Model and benchmark references appear in Fig.~\ref{fig:cs3-helm-heatmap}; implementation details are in App.~\ref{app:case3}. % Case Study 3 heatmap figure fragment.
% Assumes the main document preamble already loads:
%   booktabs, array, graphicx, xcolor with table support, natbib/hyperref.
% Input with: \input{case3_heatmap_figure_input}

\begingroup

% Manually hacked blue/orange heatmap palette.
% Higher agreement = blue; lower agreement = orange; N/A = gray.
\definecolor{cs3blue100}{HTML}{08306B}
\definecolor{cs3blue999}{HTML}{08326F}
\definecolor{cs3blue998}{HTML}{083572}
\definecolor{cs3blue997}{HTML}{083876}
\definecolor{cs3blue996}{HTML}{083B7A}
\definecolor{cs3blue991}{HTML}{0B458A}
\definecolor{cs3blue990}{HTML}{0C498E}
\definecolor{cs3blue987}{HTML}{11559A}
\definecolor{cs3blue985}{HTML}{155CA3}
\definecolor{cs3blue984}{HTML}{1761A8}
\definecolor{cs3blue983}{HTML}{1A66AD}
\definecolor{cs3blue982}{HTML}{1D6CB2}
\definecolor{cs3blue977}{HTML}{2B7DBB}
\definecolor{cs3blue960}{HTML}{5A9DCC}
\definecolor{cs3blue942}{HTML}{A6CFE3}
\definecolor{cs3blue940}{HTML}{ADD3E6}
\definecolor{cs3blue935}{HTML}{BBDCEB}
\definecolor{cs3blue934}{HTML}{BEDDEC}
\definecolor{cs3blue928}{HTML}{D0E7F1}
\definecolor{cs3blue922}{HTML}{E0EFF5}
\definecolor{cs3blue920}{HTML}{E4F1F6}
\definecolor{cs3blue910}{HTML}{F2F7F8}
\definecolor{cs3orange788}{HTML}{E98B3A}
\definecolor{cs3na}{HTML}{E6E6E6}

% Local helpers for the figure.
\def\heatdark#1#2{%
  \cellcolor{#1}\textcolor{white}{\strut #2}%
}
\def\heatlight#1#2{%
  \cellcolor{#1}\textcolor{black}{\strut #2}%
}
\def\nacell{%
  \cellcolor{cs3na}\textcolor{black!60}{\strut N/A}%
}
\def\legendpatch#1{%
  \begingroup
  \setlength{\fboxsep}{0pt}%
  \colorbox{#1}{\makebox[0.14in][c]{\rule{0pt}{0.105in}}}%
  \endgroup
}
\def\cslegend{%
  \scriptsize
  \begin{tabular}{@{}c@{}}
    100\\[-0.2em]
    \vbox{%
      \hbox{\legendpatch{cs3blue100}}%
      \nointerlineskip
      \hbox{\legendpatch{cs3blue960}}%
      \nointerlineskip
      \hbox{\legendpatch{cs3blue920}}%
      \nointerlineskip
      \hbox{\legendpatch{cs3blue910}}%
      \nointerlineskip
      \hbox{\legendpatch{cs3orange788}}%
    }\\[-0.05em]
    75
  \end{tabular}%
}
\def\rotlab#1{%
  \raisebox{-0.08in}[0.44in][0pt]{%
    \rotatebox[origin=lb]{48}{%
      \parbox{0.74in}{\centering\tiny #1}%
    }%
  }%
}
\begin{figure}[h!]
  \centering
  \scriptsize
  \setlength{\tabcolsep}{1.8pt}
  \renewcommand{\arraystretch}{1.18}

  \begin{tabular}{@{}c@{\hspace{1.2em}}c@{}}
  \resizebox{0.885\linewidth}{!}{%
  \begin{tabular}{@{}l*{14}{>{\centering\arraybackslash}p{0.042\linewidth}}@{}}
    % \toprule
    
    & 
    \rotlab{\raggedright \hypertarget{entity}{Entity}\\Imputation~\citep{mei2021capturing}} &
    \rotlab{\raggedright IMDB~\citep{maas-etal-2011-learning}} &
    \rotlab{\raggedright Synth.\\Reasoning~\citep{wu2021lime}} &
    \rotlab{\raggedright TruthfulQA~\citep{lin-etal-2022-truthfulqa}} &
    \rotlab{\raggedright GSM~\citep{cobbe2021training}} &
    \rotlab{\raggedright LSAT QA~\citep{zhong2021arlsat}} &
    \rotlab{\raggedright Civil\\Comments~\citep{borkan2019nuanced}} &
    \rotlab{\raggedright MMLU~\citep{mmlu}} &
    \rotlab{\raggedright BoolQ~\citep{clark-etal-2019-boolq}} &
    \rotlab{\raggedright QuAC~\citep{choi-etal-2018-quac}} &
    \rotlab{\raggedright NarrativeQA~\citep{kocisky-etal-2018-narrativeqa}} &
    \rotlab{\raggedright Natural Syn.\\Reason.~\citep{clark2020transformers}} &
    \rotlab{\raggedright WikiFact~\citep{petroni-etal-2019-language}} &
    \rotlab{\raggedright Entity\\Matching~\citep{kopcke2010evaluation}} \\[8pt]
    % \midrule

    Pythia-6.9B~\citep{pmlr-v202-biderman23a}
      & \heatdark{cs3blue100}{100}
      & \heatdark{cs3blue100}{100}
      & \heatdark{cs3blue999}{99.9}
      & \heatdark{cs3blue998}{99.8}
      & \heatdark{cs3blue996}{99.6}
      & \heatdark{cs3blue998}{99.8}
      & \heatdark{cs3blue999}{99.9}
      & \heatdark{cs3blue100}{100}
      & \heatdark{cs3blue996}{99.6}
      & \heatdark{cs3blue997}{99.7}
      & \heatdark{cs3blue984}{98.4}
      & \heatdark{cs3blue997}{99.7}
      & \heatlight{cs3blue928}{92.8}
      & \nacell \\

    Vicuna-7B v1.3~\citep{vicuna2023}
      & \heatdark{cs3blue100}{100}
      & \heatdark{cs3blue100}{100}
      & \heatdark{cs3blue997}{99.7}
      & \heatdark{cs3blue998}{99.8}
      & \heatdark{cs3blue990}{99.0}
      & \heatdark{cs3blue998}{99.8}
      & \heatdark{cs3blue997}{99.7}
      & \heatdark{cs3blue100}{100}
      & \heatdark{cs3blue100}{100}
      & \heatdark{cs3blue987}{98.7}
      & \heatdark{cs3blue985}{98.5}
      & \heatdark{cs3blue998}{99.8}
      & \heatlight{cs3blue920}{92.0}
      & \nacell \\

    Falcon-7B~\citep{almazrouei2023falcon}
      & \heatdark{cs3blue100}{100}
      & \heatdark{cs3blue991}{99.1}
      & \heatdark{cs3blue982}{98.2}
      & \heatdark{cs3blue977}{97.7}
      & \heatdark{cs3blue983}{98.3}
      & \heatdark{cs3blue960}{96.0}
      & \heatlight{cs3blue940}{94.0}
      & \heatlight{cs3blue935}{93.5}
      & \heatlight{cs3blue934}{93.4}
      & \heatlight{cs3blue942}{94.2}
      & \heatlight{cs3blue910}{91.0}
      & \heatlight{cs3orange788}{78.8}
      & \heatlight{cs3blue922}{92.2}
      & \nacell \\

    % \bottomrule
  \end{tabular}%
  } & \raisebox{-0.3in}{\cslegend}
  \end{tabular}

  \caption{Instance-level score agreement between model--benchmark pairs for official HELM records and local reproductions after conversion to \eee{}. Values report the percentage of aligned \mbox{(instance, core metric)} score pairs with identical official and local scores up to numerical tolerance. N/A denotes no content-hash overlap, making Entity-Matching~\citep{kopcke2010evaluation} incomparable.}
  \label{fig:cs3-helm-heatmap}
\end{figure}

\endgroup Figure~\ref{fig:cs3-helm-heatmap} shows that \eeeshort{} exposes score and example mismatches. Entity-Matching (\hyperlink{entity}{right column}) is incomparable because official and reproduced records select different Abt--Buy~\citep{kopcke2010evaluation} examples despite using the same HELM recipe; App.~\ref{app:case3} traces this to row-order changes in the data-processing stack. SyntheticReasoning-Natural reveals a serving artifact: official Pythia completions are empty and receive zero scores, whereas local completions are non-empty and receive low but non-zero scores. WikiFact~\citep{petroni-etal-2019-language} shows roughly 92\% agreement across models, consistent with stochastic sampling. Smaller residual disagreements remain after these cases are explained. Overall, \eeeshort{} enables reproducibility forensics by surfacing mismatched example sets, empty or truncated completions, stochastic disagreement, and residual score differences. Notably, \eeeshort{} does not replace framework-level provenance: when serving details are missing, the schema can surface the discrepancies but cannot always determine their exact cause.

%%% END CASE STUDY 3 %%%%

\subsection{Case Study 4: \eee enables meta-analysis using Item Response Theory}

\label{case4}
\begin{wrapfigure}{r}{0.7\textwidth}
    \centering
    \includegraphics[width=1\linewidth]{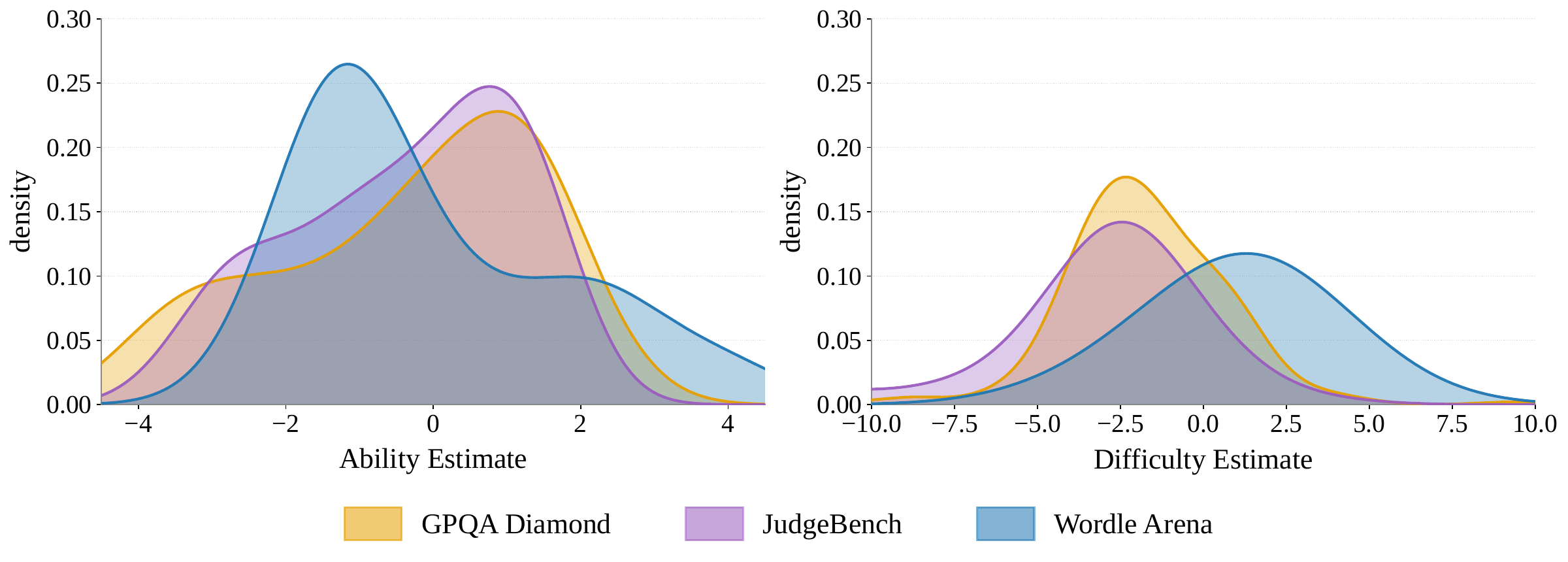}
    \caption{ Estimated model abilities (left) and item difficulties (right) for three datasets included in \eee{}.}
    \label{fig:irt}
\end{wrapfigure}

We showcase how the instance-level data in \eeeshort{} can be used for analysis across datasets with otherwise incomparable data: GPQA Diamond \cite{rein2024gpqa}, Wordle Arena \cite{murphy2025ai}, and JudgeBench \cite{Tan2024JudgeBenchAB}. We fit a unidimensional Item Response Theory (IRT) model to analyze model ability and example difficulty distributions (App.~\ref{app:casestudies4}). At the dataset level, we see that the distribution of item difficulties and model abilities varies (Fig.~\ref{fig:irt}). In particular, the Wordle Arena examples are generally more difficult and their difficulty varies more. This suggests GPQA may quickly shift from hard to saturated, while Wordle Arena will likely continue to surface challenging cases. Overall, this showcases how the cross benchmark consistency simplifies comparing datasets and gaining new insights on existing evaluations.
%IRT-type analyses such as these can augment other aggregate measures by giving more instance-level information, such as difficulty or discriminability. 

\section{Limitations}
\label{sec:limitations}
% Schema coverage: The schema covers text-based, single-model evaluations well. Multi-modal evaluations, human preference judgements (e.g.\ Chatbot Arena Elo), and multi-agent settings are partially supported and will be extended by the community, following the extensibility design of Croissant~\citep{croissant2024}.

% Adoption: The schema's value grows with adoption. Converters lower the contribution cost, but labs and leaderboard operators may still omit generation parameters for proprietary systems. Where metadata is absent, the schema records that absence explicitly.

% Deduplication: The UUID-per-run design is lossless but places deduplication at the analysis layer. Reference implementations of common equivalence criteria are planned as utilities in the package.

% Consistency: While we put measures for automatic verification and community governance of data additions, as the resource is all based on participation, errors or inconsistencies or underreported areas are likely to occur.

We note several limitations to the current \eeeshort{} schema. First, coverage is strongest for text-based, single-model evaluations, while multi-modal evaluations, human preference judgments (e.g.\ Chatbot Arena Elo), and multi-agent settings are only partially supported; these areas are intended to be extended by the community. % following the extensibility design of Croissant~\citep{croissant2024}. 
Second, the value of the schema depends on broad community adoption: although converters reduce the cost of contribution, labs and leaderboard operators may still omit important metadata, such as generation parameters for proprietary systems, and this missing metadata is explicitly recorded. %in which case the schema records the absence explicitly rather than inferring missing details. 
Third, the UUID-per-run design preserves information losslessly but shifts deduplication to the analysis layer, with reference implementations of common equivalence criteria planned as package utilities. Fourth, while the schema aids in finding reproducibility issues, as it does not run evaluations, information unique to this running settings is likely not reported in it. Finally, despite automatic verification mechanisms and community governance for data additions, the resource remains participation-based, so errors, inconsistencies, and uneven reporting across evaluation areas are likely to occur. 

\section{Conclusion}
\label{sec:conclusion}
We present \eee{}: a schema, validation pipeline, and converter suite that establishes a common language for AI evaluation reporting, supported by a growing community dataset of evaluation results. By recording the context needed to interpret a score, not just the score itself, \eeeshort{} makes existing evaluation results reusable and enables analyses that per-paper reporting cannot support. We invite the community to contribute results, extend the schema, and build upon the dataset in future research.

%
%\begin{ack}
%\cready{Fill in after de-anonymisation.}
%\end{ack}

% \bibliographystyle{cha}
%\bibliographystyle{unsrtnat}
\bibliographystyle{acl_natbib}
\bibliography{every_eval_ever}

@misc{fireworks2025kimik2p5,
  author = {Fireworks AI Engineering Team},
  title = {Quality-First With Kimi k2.5: The Importance of Post-Training and Serving Infrastructure},
  year = {2025},
  url = {https://fireworks.ai/blog/quality-first-with-kimi-k2p5},
  urldate = {2026-04-07},
  organization = {Fireworks AI}
}

@misc{bordes2025evalfactsheetsstructuredframework,
      title={Eval Factsheets: A Structured Framework for Documenting AI Evaluations}, 
      author={Florian Bordes and Candace Ross and Justine T Kao and Evangelia Spiliopoulou and Adina Williams},
      year={2025},
      eprint={2512.04062},
      archivePrefix={arXiv},
      primaryClass={cs.LG},
      url={https://arxiv.org/abs/2512.04062}, 
}

@article{staufer2025audit,
  title={Audit cards: Contextualizing ai evaluations},
  author={Staufer, Leon and Yang, Mick and Reuel, Anka and Casper, Stephen},
  journal={arXiv preprint arXiv:2504.13839},
  year={2025}
}

@inproceedings{kurtic2025give,
  title={“Give Me BF16 or Give Me Death”? Accuracy-Performance Trade-Offs in LLM Quantization},
  author={Kurtic, Eldar and Marques, Alexandre Noll and Pandit, Shubhra and Kurtz, Mark and Alistarh, Dan},
  booktitle={Proceedings of the 63rd Annual Meeting of the Association for Computational Linguistics (Volume 1: Long Papers)},
  pages={26872--26886},
  year={2025}
}

@misc{li2026adaptivetestingllmevaluation,
      title={Adaptive Testing for LLM Evaluation: A Psychometric Alternative to Static Benchmarks}, 
      author={Peiyu Li and Xiuxiu Tang and Si Chen and Ying Cheng and Ronald Metoyer and Ting Hua and Nitesh V. Chawla},
      year={2026},
      eprint={2511.04689},
      archivePrefix={arXiv},
      primaryClass={cs.CL},
      url={https://arxiv.org/abs/2511.04689}, 
}

@misc{polo2024tinybenchmarksevaluatingllmsfewer,
      title={tinyBenchmarks: evaluating LLMs with fewer examples}, 
      author={Felipe Maia Polo and Lucas Weber and Leshem Choshen and Yuekai Sun and Gongjun Xu and Mikhail Yurochkin},
      year={2024},
      eprint={2402.14992},
      archivePrefix={arXiv},
      primaryClass={cs.CL},
      url={https://arxiv.org/abs/2402.14992}, 
}

@inproceedings{yuanunderstanding,
  title={Understanding and Mitigating Numerical Sources of Nondeterminism in LLM Inference},
  author={Yuan, Jiayi and Li, Hao and Ding, Xinheng and Xie, Wenya and Li, Yu-Jhe and Zhao, Wentian and Wan, Kun and Shi, Jing and Hu, Xia and Liu, Zirui},
  year={2025},
  booktitle={The Thirty-ninth Annual Conference on Neural Information Processing Systems}
}

@misc{eval_harness,
  title     = {A Framework for Few-Shot Language Model Evaluation},
  author    = {Gao, Leo and Tow, Jonathan and Abbasi, Baber and Biderman,
               Stella and Black, Sid and DiPofi, Anthony and Foster, Charles
               and Golding, Laurence and Hsu, Jeffrey and {Le Noac'h}, Alain
               and Li, Haonan and McDonell, Kyle and Muennighoff, Niklas and
               Ociepa, Chris and Phang, Jason and Reynolds, Laria and
               Schoelkopf, Hailey and Skowron, Aviya and Sutawika, Lintang
               and Tang, Eric and Thite, Anish and Wang, Ben and Wang, Kevin
               and Zou, Andy},
  publisher = {Zenodo},
  month     = dec,
  year      = {2023},
  version   = {v0.4.0},
  doi       = {10.5281/zenodo.5371628},
  url       = {https://doi.org/10.5281/zenodo.5371628},
}

@article{helm,
  title   = {Holistic Evaluation of Language Models},
  author  = {Liang, Percy and Bommasani, Rishi and Lee, Tony and Tsipras,
             Dimitris and Soylu, Dilara and Yasunaga, Michihiro and Zhang,
             Yian and Narayanan, Deepak and Wu, Yuhuai and Kumar, Ananya
             and Newman, Benjamin and Yuan, Binhang and Yan, Bobby and
             Zhang, Ce and Cosgrove, Christian and Manning, Christopher D.
             and R{\'e}, Christopher and Acosta-Navas, Diana and Hudson,
             Drew and Zelikman, Eric and Durmus, Esin and Ladhak, Faisal
             and Rong, Frieda and Ren, Hongyu and Yao, Huaxiu and Wang, Jue
             and Santhanam, Keshav and Orr, Laurel and Zheng, Lucia and
             Y{\"u}ksekg{\"o}n{\"u}l, Mert and Suzgun, Mirac and Kim, Nathan
             and Guha, Neel and Chatterji, Niladri and Khattab, Omar and
             Henderson, Peter and Huang, Qian and Chi, Ryan and Xie, Sang
             Michael and Santurkar, Shibani and Ganguli, Surya and
             Hashimoto, Tatsunori and Icard, Thomas and Zhang, Tianyi and
             Chaudhary, Vishrav and Wang, William and Li, Xuechen and
             Mai, Yifan and Zhang, Yuhui and Koreeda, Yuta},
  journal = {Transactions on Machine Learning Research},
  year    = {2023},
}

@misc{inspectAI,
  author = {AI Security Institute, UK},
  title = {Inspect {AI:} {Framework} for {Large} {Language} {Model}
    {Evaluations}},
  date = {2024-05},
  url = {https://github.com/UKGovernmentBEIS/inspect_ai},
  langid = {en},
    year    = {2024},
}

@inproceedings{mmlu,
  title     = {Measuring Massive Multitask Language Understanding},
  author    = {Hendrycks, Dan and Burns, Collin and Basart, Steven and
               Zou, Andy and Mazeika, Mantas and Song, Dawn and
               Steinhardt, Jacob},
  booktitle = {International Conference on Learning Representations},
  year      = {2021},
}

@inproceedings{croissant2024,
  title     = {Croissant: A Metadata Format for {ML}-Ready Datasets},
  author    = {Akhtar, Mubashara and Benjelloun, Omar and Conforti,
               Costanza and Foschini, Luca and Gijsbers, Pieter and
               {Giner-Miguelez}, Joan and Goswami, Sujata and Jain,
               Nitisha and Karamousadakis, Michalis and Krishna,
               Satyapriya and Kuchnik, Michael and Lesage, Sylvain and
               Lhoest, Quentin and Marcenac, Pierre and Maskey, Manil and
               Mattson, Peter and Oala, Luis and Oderinwale, Hamidah and
               Ruyssen, Pierre and Santos, Tim and Shinde, Rajat and
               Simperl, Elena and Suresh, Arjun and Thomas, Goeffry and
               Tykhonov, Slava and Vanschoren, Joaquin and Varma, Susheel
               and {van der Velde}, Jos and Vogler, Steffen and Wu,
               Carole-Jean and Zhang, Luyao},
  booktitle = {Advances in Neural Information Processing Systems},
  volume    = {37},
  pages     = {82133--82148},
  doi       = {10.52202/079017-2610},
  year      = {2024},
}

@misc{jiang2026positionscienceaievaluation,
title={Position: Science of AI Evaluation Requires Item-level Benchmark Data},
author={Han Jiang and Susu Zhang and Xiaoyuan Yi and Xing Xie and Ziang Xiao},
year={2026},
eprint={2604.03244},
archivePrefix={arXiv},
primaryClass={cs.AI},
url={https://arxiv.org/abs/2604.03244},
}

@misc{epoch_ai_about,
  author       = {{Epoch AI}},
  title        = {About Us: Making sense of AI},
  year         = {2026},
  howpublished = {\url{https://epoch.ai/about}},
  note         = {Accessed: 2026-05-01}
}

@misc{metr_blog,
    title = {Time Horizon 1.1},
    author = {METR},
    howpublished = {\url{https://metr.org/blog/2026-1-29-time-horizon-1-1/}},
    year = {2026},
    month = {01},
}

@misc{artificialanalysis2026,
  author       = {{Artificial Analysis}},
  title        = {Independent analysis of AI models and hosting providers},
  year         = {2026},
  howpublished = {\url{https://artificialanalysis.ai/}},
  note         = {Accessed: 2026-05-01}
}

@inproceedings{
abbas2025developing,
title={Developing and Maintaining an Open-Source Repository of {AI} Evaluations: Challenges and Insights},
author={Alexandra Abbas and Celia Waggoner and Justin Olive},
booktitle={Championing Open-source DEvelopment in ML Workshop @ ICML25},
year={2025},
url={https://openreview.net/forum?id=yw33GWAEOK}
}

@misc{openllm_leaderboard,
  title        = {Open {LLM} Leaderboard},
  author       = {Beeching, Edward and Fourrier, Cl{\'e}mentine and Habib,
                  Nathan and Han, Sheon and Lambert, Nathan and Rajani,
                  Nazneen and Sanseviero, Omar and Belkada, Younes and
                  Wolf, Thomas},
  howpublished = {\url{https://huggingface.co/spaces/HuggingFaceH4/open_llm_leaderboard}},
  year         = {2023},
}

@inproceedings{chatbot_arena,
  title     = {Chatbot Arena: An Open Platform for Evaluating {LLMs} by
               Human Preference},
  author    = {Zheng, Lianmin and Chiang, Wei-Lin and Sheng, Ying and
               Zhuang, Siyuan and Wu, Zhanghao and Zhuang, Yonghao and
               Lin, Zi and Li, Zhuohan and Li, Dacheng and Xing, Eric P.
               and Zhang, Hao and Gonzalez, Joseph E. and Stoica, Ion},
  booktitle = {Proceedings of the 41st International Conference on Machine
               Learning},
  year      = {2024},
}

@inproceedings{rodriguez2021evaluation,
  title={Evaluation examples are not equally informative: How should that change NLP leaderboards?},
  author={Rodriguez, Pedro and Barrow, Joe and Hoyle, Alexander Miserlis and Lalor, John P and Jia, Robin and Boyd-Graber, Jordan Lee},
  booktitle={Proceedings of the 59th Annual Meeting of the Association for Computational Linguistics and the 11th International Joint Conference on Natural Language Processing (Volume 1: Long Papers)},
  pages={4486--4503},
  year={2021}
}

@article{michaelov2026open,
  title={How Open Must Language Models be to Enable Reliable Scientific Inference?},
  author={Michaelov, James A and Arnett, Catherine and Chang, Tyler A and Rivi{\`e}re, Pamela D and Taylor, Samuel M and Jones, Cameron R and Trott, Sean and Levy, Roger P and Bergen, Benjamin K and Altman, Micah},
  journal={arXiv preprint arXiv:2603.26539},
  year={2026}
}

@misc{batzner2026shared,
  title={Shared Task of Every Eval Ever: Building a Unifying, Standardized Database of LLM Evaluations},
  author={Batzner, Jan and Choshen, Leshem and Nelaturu, Sree Harsha and Stachura, Damian and Kornilova, Anastassia and Long, Yanan and Gohar, Usman and Tran, Andrew and Ghosh, Avijit},
  year={2026}, 
  howpublished={Preprint}
}

@inproceedings{bandel2024unitxt,
  title={Unitxt: Flexible, shareable and reusable data preparation and evaluation for generative ai},
  author={Bandel, Elron and Perlitz, Yotam and Venezian, Elad and Friedman, Roni and Arviv, Ofir and Orbach, Matan and Don-Yehiya, Shachar and Sheinwald, Dafna and Gera, Ariel and Choshen, Leshem and others},
  booktitle={Proceedings of the 2024 Conference of the North American Chapter of the Association for Computational Linguistics: Human Language Technologies (Volume 3: System Demonstrations)},
  pages={207--215},
  year={2024}
}

@article{siegelcore,
  title={CORE-Bench: Fostering the Credibility of Published Research Through a Computational Reproducibility Agent Benchmark},
  author={Siegel, Zachary S and Kapoor, Sayash and Nadgir, Nitya and Stroebl, Benedikt and Narayanan, Arvind},
  journal={Transactions on Machine Learning Research},
  year={2024}
}

@article{polo2024efficient,
  title={Efficient multi-prompt evaluation of llms},
  author={Polo, Felipe M and Xu, Ronald and Weber, Lucas and Silva, M{\'\i}rian and Bhardwaj, Onkar and Choshen, Leshem and de Oliveira, Allysson F and Sun, Yuekai and Yurochkin, Mikhail},
  journal={Advances in Neural Information Processing Systems},
  volume={37},
  pages={22483--22512},
  year={2024}
}

@inproceedings{shabtaylivexiv,
  title={LiveXiv-A Multi-Modal live benchmark based on Arxiv papers content},
  author={Shabtay, Nimrod and Polo, Felipe Maia and Doveh, Sivan and Lin, Wei and Mirza, Muhammad Jehanzeb and Choshen, Leshem and Yurochkin, Mikhail and Sun, Yuekai and Arbelle, Assaf and Karlinsky, Leonid and others},
  booktitle={The Thirteenth International Conference on Learning Representations},
  year={2024},
}

@article{kipnis2024metabench,
  title={metabench--A Sparse Benchmark of Reasoning and Knowledge in Large Language Models},
  author={Kipnis, Alex and Voudouris, Konstantinos and Buschoff, Luca M Schulze and Schulz, Eric},
  journal={arXiv preprint arXiv:2407.12844},
  year={2024}
}

@article{habba2026growing,
  title={Growing Pains: Extensible and Efficient LLM Benchmarking Via Fixed Parameter Calibration},
  author={Habba, Eliya and Itzhak, Itay and Yehudai, Asaf and Perlitz, Yotam and Bandel, Elron and Shmueli-Scheuer, Michal and Choshen, Leshem and Stanovsky, Gabriel},
  journal={arXiv preprint arXiv:2604.12843},
  year={2026}
}

@inproceedings{polo2025statistical,
  title={A Statistical Framework for Game-Based AI Evaluation},
  author={Polo, Felipe Maia and Choshen, Leshem and Sun, Yuekai and Greenewald, Kristjan},
  booktitle={NeurIPS 2025 Workshop on Evaluating the Evolving LLM Lifecycle: Benchmarks, Emergent Abilities, and Scaling},
  year={2025},
}

@inproceedings{prompt_sensitivity,
  title     = {Fantastically Ordered Prompts and Where to Find Them:
               Overcoming Few-Shot Prompt Order Sensitivity},
  author    = {Lu, Yao and Bartolo, Max and Moore, Alastair and Riedel,
               Sebastian and Stenetorp, Pontus},
  booktitle = {Proceedings of the 60th Annual Meeting of the Association
               for Computational Linguistics (Volume 1: Long Papers)},
  pages     = {8086--8098},
  year      = {2022},
  publisher = {Association for Computational Linguistics},
}

@inproceedings{benchmark_contamination,
  title     = {Data Contamination: From Memorization to Exploitation},
  author    = {Magar, Inbal and Schwartz, Roy},
  booktitle = {Proceedings of the 60th Annual Meeting of the Association
               for Computational Linguistics (Volume 2: Short Papers)},
  pages     = {157--165},
  year      = {2022},
  publisher = {Association for Computational Linguistics},
}

@misc{benchmark_saturation,
  title         = {Are {NLP} Benchmarks Saturating?},
  author        = {Blagec, Kathrin and Dorffner, Georg and Moradi, Milad
                   and Alam, Mehrdad and Samwald, Matthias},
  year          = {2021},
  eprint        = {2105.13977},
  archivePrefix = {arXiv},
  primaryClass  = {cs.CL},
}

@misc{eu_ai_act,
  title        = {Regulation ({EU}) 2024/1689 of the {European Parliament}
                  and of the {Council}: Artificial Intelligence Act},
  author       = {{European Parliament} and {Council of the European Union}},
  howpublished = {\url{https://eur-lex.europa.eu/legal-content/EN/TXT/?uri=CELEX:32024R1689}},
  year         = {2024},
}

@misc{akhtar2026aibenchmarksplateausystematic,
      title={When AI Benchmarks Plateau: A Systematic Study of Benchmark Saturation}, 
      author={Mubashara Akhtar and Anka Reuel and Prajna Soni and Sanchit Ahuja and Pawan Sasanka Ammanamanchi and Ruchit Rawal and Vilém Zouhar and Srishti Yadav and Chenxi Whitehouse and Dayeon Ki and Jennifer Mickel and Leshem Choshen and Marek Šuppa and Jan Batzner and Jenny Chim and Jeba Sania and Yanan Long and Hossein A. Rahmani and Christina Knight and Yiyang Nan and Jyoutir Raj and Yu Fan and Shubham Singh and Subramanyam Sahoo and Eliya Habba and Usman Gohar and Siddhesh Pawar and Robert Scholz and Arjun Subramonian and Jingwei Ni and Mykel Kochenderfer and Sanmi Koyejo and Mrinmaya Sachan and Stella Biderman and Zeerak Talat and Avijit Ghosh and Irene Solaiman},
      year={2026},
      eprint={2602.16763},
      archivePrefix={arXiv},
      primaryClass={cs.AI},
      url={https://arxiv.org/abs/2602.16763}, 
}

@misc{fourrier2023openllmleaderboard,
  title        = {What's Going On with the Open {LLM} Leaderboard?},
  author       = {Fourrier, Cl\'{e}mentine and Habib, Nathan and Launay, Julien and Wolf, Thomas},
  year         = {2023},
  month        = jun,
  day          = {23},
  howpublished = {Hugging Face Blog},
  url          = {https://huggingface.co/blog/open-llm-leaderboard-mmlu},
}

@inproceedings{liao2021are,
  title     = {Are We Learning Yet? {A} Meta Review of Evaluation Failures Across Machine Learning},
  author    = {Liao, Thomas and Taori, Rohan and Raji, Inioluwa Deborah and Schmidt, Ludwig},
  booktitle = {Proceedings of the Neural Information Processing Systems Track on Datasets and Benchmarks},
  year      = {2021},
  url       = {https://datasets-benchmarks-proceedings.neurips.cc/paper/2021/hash/757b505cfd34c64c85ca5b5690ee5293-Abstract-round2.html}
}

@inproceedings{bowman-dahl-2021-will,
  title     = {What Will it Take to Fix Benchmarking in Natural Language Understanding?},
  author    = {Bowman, Samuel R. and Dahl, George E.},
  booktitle = {Proceedings of the 2021 Conference of the North American Chapter of the Association for Computational Linguistics: Human Language Technologies},
  pages     = {4843--4855},
  year      = {2021},
  address   = {Online},
  publisher = {Association for Computational Linguistics},
  doi       = {10.18653/v1/2021.naacl-main.385},
  url       = {https://aclanthology.org/2021.naacl-main.385/}
}

@inproceedings{ethayarajh-jurafsky-2020-utility,
  title     = {Utility is in the Eye of the User: {A} Critique of {NLP} Leaderboards},
  author    = {Ethayarajh, Kawin and Jurafsky, Dan},
  booktitle = {Proceedings of the 2020 Conference on Empirical Methods in Natural Language Processing (EMNLP)},
  pages     = {4846--4853},
  year      = {2020},
  address   = {Online},
  publisher = {Association for Computational Linguistics},
  doi       = {10.18653/v1/2020.emnlp-main.393},
  url       = {https://aclanthology.org/2020.emnlp-main.393/}
}

@inproceedings{raji2021ai,
  title     = {{AI} and the Everything in the Whole Wide World Benchmark},
  author    = {Raji, Inioluwa Deborah and Denton, Emily and Bender, Emily M. and Hanna, Alex and Paullada, Amandalynne},
  booktitle = {Proceedings of the Neural Information Processing Systems Track on Datasets and Benchmarks},
  year      = {2021},
  url       = {https://datasets-benchmarks-proceedings.neurips.cc/paper/2021/hash/084b6fbb10729ed4da8c3d3f5a3ae7c9-Abstract-round2.html}
}

@article{burnell2023rethink,
  title   = {Rethink reporting of evaluation results in {AI}},
  author  = {Burnell, Ryan and Schellaert, Wout and Burden, John and Ullman, Tomer D. and Martinez-Plumed, Fernando and Tenenbaum, Joshua B. and Rutar, Danaja and Cheke, Lucy G. and Sohl-Dickstein, Jascha and Mitchell, Melanie and Kiela, Douwe and Shanahan, Murray and Voorhees, Ellen M. and Cohn, Anthony G. and Leibo, Joel Z. and Hernandez-Orallo, Jose},
  journal = {Science},
  volume  = {380},
  number  = {6641},
  pages   = {136--138},
  year    = {2023},
  doi     = {10.1126/science.adf6369}
}

@misc{biderman2024lessonstrenchesreproducibleevaluation,
      title={Lessons from the Trenches on Reproducible Evaluation of Language Models}, 
      author={Stella Biderman and Hailey Schoelkopf and Lintang Sutawika and Leo Gao and Jonathan Tow and Baber Abbasi and Alham Fikri Aji and Pawan Sasanka Ammanamanchi and Sidney Black and Jordan Clive and Anthony DiPofi and Julen Etxaniz and Benjamin Fattori and Jessica Zosa Forde and Charles Foster and Jeffrey Hsu and Mimansa Jaiswal and Wilson Y. Lee and Haonan Li and Charles Lovering and Niklas Muennighoff and Ellie Pavlick and Jason Phang and Aviya Skowron and Samson Tan and Xiangru Tang and Kevin A. Wang and Genta Indra Winata and François Yvon and Andy Zou},
      year={2024},
      eprint={2405.14782},
      archivePrefix={arXiv},
      primaryClass={cs.CL},
      url={https://arxiv.org/abs/2405.14782}, 
}

@misc{reuel2024betterbenchassessingaibenchmarks,
      title={BetterBench: Assessing AI Benchmarks, Uncovering Issues, and Establishing Best Practices}, 
      author={Anka Reuel and Amelia Hardy and Chandler Smith and Max Lamparth and Malcolm Hardy and Mykel J. Kochenderfer},
      year={2024},
      eprint={2411.12990},
      archivePrefix={arXiv},
      primaryClass={cs.AI},
      url={https://arxiv.org/abs/2411.12990}, 
}

@misc{hofmann2025autobenchmarkcardautomatedsynthesisbenchmark,
      title={Auto-BenchmarkCard: Automated Synthesis of Benchmark Documentation}, 
      author={Aris Hofmann and Inge Vejsbjerg and Dhaval Salwala and Elizabeth M. Daly},
      year={2025},
      eprint={2512.09577},
      archivePrefix={arXiv},
      primaryClass={cs.HC},
      url={https://arxiv.org/abs/2512.09577}, 
}

@misc{sokol2025benchmarkcardsstandardizeddocumentationlarge,
      title={BenchmarkCards: Standardized Documentation for Large Language Model Benchmarks}, 
      author={Anna Sokol and Elizabeth Daly and Michael Hind and David Piorkowski and Xiangliang Zhang and Nuno Moniz and Nitesh Chawla},
      year={2025},
      eprint={2410.12974},
      archivePrefix={arXiv},
      primaryClass={cs.CL},
      url={https://arxiv.org/abs/2410.12974}, 
}

@article{wang2024benchmark,
  title   = {Benchmark suites instead of leaderboards for evaluating {AI} fairness},
  author  = {Wang, Angelina and Hertzmann, Aaron and Russakovsky, Olga},
  journal = {Patterns},
  volume  = {5},
  number  = {11},
  pages   = {101080},
  year    = {2024},
  doi     = {10.1016/j.patter.2024.101080}
}

@article{nelaturu2024hardware,
  title={On The Fairness Impacts of Hardware Selection in Machine Learning},
  author={Nelaturu, Sree Harsha and Ravichandran, Nishaanth Kanna and
          Tran, Cuong and Hooker, Sara and Fioretto, Ferdinando},
  journal={Proceedings of the 41st International Conference on Machine Learning (ICML)},
  year={2024},
  url={https://proceedings.mlr.press/v235/nelaturu24a.html}
}

@inproceedings{habba2025dove,
  title={DOVE: A Large-Scale Multi-Dimensional Predictions Dataset Towards Meaningful LLM Evaluation},
  author={Habba, Eliya and Arviv, Ofir and Itzhak, Itay and Perlitz, Yotam and Bandel, Elron and Choshen, Leshem and Shmueli-Scheuer, Michal and Stanovsky, Gabriel},
  booktitle={Findings of the Association for Computational Linguistics: ACL 2025},
  pages={11744--11763},
  year={2025}
}

@inproceedings{perlitz-etal-2024-efficient,
    title = "Efficient Benchmarking (of Language Models)",
    author = "Perlitz, Yotam  and
      Bandel, Elron  and
      Gera, Ariel  and
      Arviv, Ofir  and
      Ein-Dor, Liat  and
      Shnarch, Eyal  and
      Slonim, Noam  and
      Shmueli-Scheuer, Michal  and
      Choshen, Leshem",
    editor = "Duh, Kevin  and
      Gomez, Helena  and
      Bethard, Steven",
    booktitle = "Proceedings of the 2024 Conference of the North American Chapter of the Association for Computational Linguistics: Human Language Technologies (Volume 1: Long Papers)",
    month = jun,
    year = "2024",
    address = "Mexico City, Mexico",
    publisher = "Association for Computational Linguistics",
    url = "https://aclanthology.org/2024.naacl-long.139/",
    doi = "10.18653/v1/2024.naacl-long.139",
    pages = "2519--2536"
}

@inproceedings{choshen2025hitchhiker,
  title={A Hitchhiker’s Guide to Scaling Law Estimation},
  author={Choshen, Leshem and Zhang, Yang and Andreas, Jacob},
  booktitle={International Conference on Machine Learning},
  pages={10683--10699},
  year={2025},
  organization={PMLR}
}

@article{ruan2024observational,
  title={Observational scaling laws and the predictability of language model performance},
  author={Ruan, Yangjun and Maddison, Chris J and Hashimoto, Tatsunori B},
  journal={Advances in Neural Information Processing Systems},
  volume={37},
  pages={15841--15892},
  year={2024}
}

@misc{alpacaeval,
  title={{AlpacaEval}: An Automatic Evaluator of Instruction-following Models},
  author={Li, Xuechen and Zhang, Tianyi and Dubois, Yann and Taori, Rohan and
          Gulrajani, Ishaan and Guestrin, Carlos and Liang, Percy and
          Hashimoto, Tatsunori B.},
  year={2023},
  url={https://github.com/tatsu-lab/alpaca_eval}
}

@inproceedings{zheng2023mtbench,
  title={Judging {LLM}-as-a-Judge with {MT-Bench} and Chatbot Arena},
  author={Zheng, Lianmin and Chiang, Wei-Lin and Sheng, Ying and Zhuang, Siyuan
          and Wu, Zhanghao and Zhuang, Yonghao and Lin, Zi and Li, Zhuohan and
          Li, Dacheng and Xing, Eric P. and Zhang, Hao and
          Gonzalez, Joseph E. and Stoica, Ion},
  booktitle={Advances in Neural Information Processing Systems},
  volume={36},
  year={2023}
}

@misc{bandel2026generalagentevaluation,
      title={General Agent Evaluation}, 
      author={Elron Bandel and Asaf Yehudai and Lilach Eden and Yehoshua Sagron and Yotam Perlitz and Elad Venezian and Natalia Razinkov and Natan Ergas and Shlomit Shachor Ifergan and Segev Shlomov and Michal Jacovi and Leshem Choshen and Liat Ein-Dor and Yoav Katz and Michal Shmueli-Scheuer},
      year={2026},
      eprint={2602.22953},
      archivePrefix={arXiv},
      primaryClass={cs.AI},
      url={https://arxiv.org/abs/2602.22953}, 
}

@misc{cocoabenchteam2026cocoabenchevaluatingunifieddigital,
      title={CocoaBench: Evaluating Unified Digital Agents in the Wild}, 
      author={Shibo Hao and Zhining Zhang and Zhiqi Liang and Tianyang Liu and Yuheng Zha and Qiyue Gao and Jixuan Chen and Zilong Wang and Zhoujun Cheng and Haoxiang Zhang and Junli Wang and Hexi Jin and Boyuan Zheng and Kun Zhou and Yu Wang and Feng Yao and Licheng Liu and Yijiang Li and Zhifei Li and Zhengtao Han and Pracha Promthaw and Tommaso Cerruti and Xiaohan Fu and Ziqiao Ma and Jingbo Shang and Lianhui Qin and Julian McAuley and Eric P. Xing and Zhengzhong Liu and Rupesh Kumar Srivastava and Zhiting Hu},
      year={2026},
      eprint={2604.11201},
      archivePrefix={arXiv},
      primaryClass={cs.CL},
      url={https://arxiv.org/abs/2604.11201}, 
}

@misc{yehudai2025surveyevaluationllmbasedagents,
      title={Survey on Evaluation of LLM-based Agents}, 
      author={Asaf Yehudai and Lilach Eden and Alan Li and Guy Uziel and Yilun Zhao and Roy Bar-Haim and Arman Cohan and Michal Shmueli-Scheuer},
      year={2025},
      eprint={2503.16416},
      archivePrefix={arXiv},
      primaryClass={cs.AI},
      url={https://arxiv.org/abs/2503.16416}, 
}

@misc{kapoor2024aiagentsmatter,
      title={AI Agents That Matter}, 
      author={Sayash Kapoor and Benedikt Stroebl and Zachary S. Siegel and Nitya Nadgir and Arvind Narayanan},
      year={2024},
      eprint={2407.01502},
      archivePrefix={arXiv},
      primaryClass={cs.LG},
      url={https://arxiv.org/abs/2407.01502}, 
}

@misc{jimenez2024swebenchlanguagemodelsresolve,
      title={SWE-bench: Can Language Models Resolve Real-World GitHub Issues?}, 
      author={Carlos E. Jimenez and John Yang and Alexander Wettig and Shunyu Yao and Kexin Pei and Ofir Press and Karthik Narasimhan},
      year={2024},
      eprint={2310.06770},
      archivePrefix={arXiv},
      primaryClass={cs.CL},
      url={https://arxiv.org/abs/2310.06770}, 
}

@misc{kapoor2025holisticagentleaderboardmissing,
      title={Holistic Agent Leaderboard: The Missing Infrastructure for AI Agent Evaluation}, 
      author={Sayash Kapoor and Benedikt Stroebl and Peter Kirgis and Nitya Nadgir and Zachary S Siegel and Boyi Wei and Tianci Xue and Ziru Chen and Felix Chen and Saiteja Utpala and Franck Ndzomga and Dheeraj Oruganty and Sophie Luskin and Kangheng Liu and Botao Yu and Amit Arora and Dongyoon Hahm and Harsh Trivedi and Huan Sun and Juyong Lee and Tengjun Jin and Yifan Mai and Yifei Zhou and Yuxuan Zhu and Rishi Bommasani and Daniel Kang and Dawn Song and Peter Henderson and Yu Su and Percy Liang and Arvind Narayanan},
      year={2025},
      eprint={2510.11977},
      archivePrefix={arXiv},
      primaryClass={cs.AI},
      url={https://arxiv.org/abs/2510.11977}, 
}

@article{bandel2026agentic, title={Agentic Systems Should be General}, author={Bandel, Elron and Yehudai, Asaf and Lacoste, Alexandre and Ghosh, Avijit and Neubig, Graham and Mitchell, Margaret and Shmueli-Scheuer, Michal and Choshen, Leshem}, journal={SSRN Electronic Journal}, url= {https://ssrn.com/abstract=6176178}, year={2026}
}

@misc{lacoste2026cubestandardunifyingagent,
      title={CUBE: A Standard for Unifying Agent Benchmarks}, 
      author={Alexandre Lacoste and Nicolas Gontier and Oleh Shliazhko and Aman Jaiswal and Kusha Sareen and Shailesh Nanisetty and Joan Cabezas and Manuel Del Verme and Omar G. Younis and Simone Baratta and Matteo Avalle and Imene Kerboua and Xing Han Lù and Elron Bandel and Michal Shmueli-Scheuer and Asaf Yehudai and Leshem Choshen and Jonathan Lebensold and Sean Hughes and Massimo Caccia and Alexandre Drouin and Siva Reddy and Tao Yu and Yu Su and Graham Neubig and Dawn Song},
      year={2026},
      eprint={2603.15798},
      archivePrefix={arXiv},
      primaryClass={cs.AI},
      url={https://arxiv.org/abs/2603.15798}, 
}

@misc{Harbor_Framework,
author = {{Harbor Framework Team}},
month = jan,
title = {{Harbor: A framework for evaluating and optimizing agents and models in container environments}},
url = {https://github.com/harbor-framework/harbor},
year = {2026}
}

@inproceedings{bandel2026readyforgeneral,
  author = {Bandel, Elron and Yehudai, Asaf and Shmueli-Scheuer, Michal},
  title = {Ready For General Agents? Let's Test It.},
  booktitle = {ICLR Blogposts 2026},
  year = {2026},
  date = {April 27, 2026},
  note = {https://iclr-blogposts.github.io/2026/blog/2026/general-agent-evaluation/},
  url  = {https://iclr-blogposts.github.io/2026/blog/2026/general-agent-evaluation/}
}

@misc{yehudai2026agenticclearautomatingmultilevel,
      title={Agentic CLEAR: Automating Multi-Level Evaluation of LLM Agents}, 
      author={Asaf Yehudai and Lilach Eden and Michal Shmueli-Scheuer},
      year={2026},
      eprint={2605.22608},
      archivePrefix={arXiv},
      primaryClass={cs.CL},
      url={https://arxiv.org/abs/2605.22608}, 
}

@misc{merity2016pointersentinelmixturemodels,
      title={Pointer Sentinel Mixture Models}, 
      author={Stephen Merity and Caiming Xiong and James Bradbury and Richard Socher},
      year={2016},
      eprint={1609.07843},
      archivePrefix={arXiv},
      primaryClass={cs.CL},
      url={https://arxiv.org/abs/1609.07843}, 
}

@misc{ghosh2026evalbottleneck,
  author       = {Ghosh, Avijit and Mai, Yifan and Channing, Georgia and Choshen, Leshem},
  title        = {{AI} evals are becoming the new compute bottleneck},
  year         = {2026},
  month        = apr,
  howpublished = {EvalEval Coalition Blog},
  url          = {https://evalevalai.com/research/2026/04/29/eval-costs-bottleneck/}
}

@inproceedings{
frantar2023optq,
title={{OPTQ}: Accurate Quantization for Generative Pre-trained Transformers},
author={Elias Frantar and Saleh Ashkboos and Torsten Hoefler and Dan Alistarh},
booktitle={The Eleventh International Conference on Learning Representations },
year={2023},
url={https://openreview.net/forum?id=tcbBPnfwxS}
}

@inproceedings{frantarsparseGPT, author = {Frantar, Elias and Alistarh, Dan}, title = {SparseGPT: massive language models can be accurately pruned in one-shot}, year = {2023}, publisher = {JMLR.org}, booktitle = {Proceedings of the 40th International Conference on Machine Learning}, articleno = {414}, numpages = {15}, location = {Honolulu, Hawaii, USA}, series = {ICML'23} }

@inproceedings{sun2023a,
title={A Simple and Effective Pruning Approach for Large Language Models},
author={Mingjie Sun and Zhuang Liu and Anna Bair and J Zico Kolter},
booktitle={Workshop on Efficient Systems for Foundation Models @ ICML2023},
year={2023},
url={https://openreview.net/forum?id=tz9JV2PRSv}
}

@misc{liu2025spinquantllmquantizationlearned,
      title={SpinQuant: LLM quantization with learned rotations}, 
      author={Zechun Liu and Changsheng Zhao and Igor Fedorov and Bilge Soran and Dhruv Choudhary and Raghuraman Krishnamoorthi and Vikas Chandra and Yuandong Tian and Tijmen Blankevoort},
      year={2025},
      eprint={2405.16406},
      archivePrefix={arXiv},
      primaryClass={cs.LG},
      url={https://arxiv.org/abs/2405.16406}, 
}

@inproceedings{liu-etal-2024-automatic,
    title = "Automatic Generation of Model and Data Cards: A Step Towards Responsible {AI}",
    author = "Liu, Jiarui  and
      Li, Wenkai  and
      Jin, Zhijing  and
      Diab, Mona",
    editor = "Duh, Kevin  and
      Gomez, Helena  and
      Bethard, Steven",
    booktitle = "Proceedings of the 2024 Conference of the North American Chapter of the Association for Computational Linguistics: Human Language Technologies (Volume 1: Long Papers)",
    month = jun,
    year = "2024",
    address = "Mexico City, Mexico",
    publisher = "Association for Computational Linguistics",
    url = "https://aclanthology.org/2024.naacl-long.110/",
    doi = "10.18653/v1/2024.naacl-long.110",
    pages = "1975--1997"
}

@article{lalor2023py,
  title={py-irt: A scalable item response theory library for python},
  author={Lalor, John Patrick and Rodriguez, Pedro},
  journal={INFORMS Journal on Computing},
  volume={35},
  number={1},
  pages={5--13},
  year={2023},
  publisher={INFORMS}
}

@misc{srivastava2023imitationgamequantifyingextrapolating,
      title={Beyond the Imitation Game: Quantifying and extrapolating the capabilities of language models}, 
      author={Aarohi Srivastava and Abhinav Rastogi and Abhishek Rao and Abu Awal Md Shoeb and Abubakar Abid and Adam Fisch and Adam R. Brown and Adam Santoro and Aditya Gupta and Adrià Garriga-Alonso and Agnieszka Kluska and Aitor Lewkowycz and Akshat Agarwal and Alethea Power and Alex Ray and Alex Warstadt and Alexander W. Kocurek and Ali Safaya and Ali Tazarv and Alice Xiang and Alicia Parrish and Allen Nie and Aman Hussain and Amanda Askell and Amanda Dsouza and Ambrose Slone and Ameet Rahane and Anantharaman S. Iyer and Anders Andreassen and Andrea Madotto and Andrea Santilli and Andreas Stuhlmüller and Andrew Dai and Andrew La and Andrew Lampinen and Andy Zou and Angela Jiang and Angelica Chen and Anh Vuong and Animesh Gupta and Anna Gottardi and Antonio Norelli and Anu Venkatesh and Arash Gholamidavoodi and Arfa Tabassum and Arul Menezes and Arun Kirubarajan and Asher Mullokandov and Ashish Sabharwal and Austin Herrick and Avia Efrat and Aykut Erdem and Ayla Karakaş and B. Ryan Roberts and Bao Sheng Loe and Barret Zoph and Bartłomiej Bojanowski and Batuhan Özyurt and Behnam Hedayatnia and Behnam Neyshabur and Benjamin Inden and Benno Stein and Berk Ekmekci and Bill Yuchen Lin and Blake Howald and Bryan Orinion and Cameron Diao and Cameron Dour and Catherine Stinson and Cedrick Argueta and César Ferri Ramírez and Chandan Singh and Charles Rathkopf and Chenlin Meng and Chitta Baral and Chiyu Wu and Chris Callison-Burch and Chris Waites and Christian Voigt and Christopher D. Manning and Christopher Potts and Cindy Ramirez and Clara E. Rivera and Clemencia Siro and Colin Raffel and Courtney Ashcraft and Cristina Garbacea and Damien Sileo and Dan Garrette and Dan Hendrycks and Dan Kilman and Dan Roth and Daniel Freeman and Daniel Khashabi and Daniel Levy and Daniel Moseguí González and Danielle Perszyk and Danny Hernandez and Danqi Chen and Daphne Ippolito and Dar Gilboa and David Dohan and David Drakard and David Jurgens and Debajyoti Datta and Deep Ganguli and Denis Emelin and Denis Kleyko and Deniz Yuret and Derek Chen and Derek Tam and Dieuwke Hupkes and Diganta Misra and Dilyar Buzan and Dimitri Coelho Mollo and Diyi Yang and Dong-Ho Lee and Dylan Schrader and Ekaterina Shutova and Ekin Dogus Cubuk and Elad Segal and Eleanor Hagerman and Elizabeth Barnes and Elizabeth Donoway and Ellie Pavlick and Emanuele Rodola and Emma Lam and Eric Chu and Eric Tang and Erkut Erdem and Ernie Chang and Ethan A. Chi and Ethan Dyer and Ethan Jerzak and Ethan Kim and Eunice Engefu Manyasi and Evgenii Zheltonozhskii and Fanyue Xia and Fatemeh Siar and Fernando Martínez-Plumed and Francesca Happé and Francois Chollet and Frieda Rong and Gaurav Mishra and Genta Indra Winata and Gerard de Melo and Germán Kruszewski and Giambattista Parascandolo and Giorgio Mariani and Gloria Wang and Gonzalo Jaimovitch-López and Gregor Betz and Guy Gur-Ari and Hana Galijasevic and Hannah Kim and Hannah Rashkin and Hannaneh Hajishirzi and Harsh Mehta and Hayden Bogar and Henry Shevlin and Hinrich Schütze and Hiromu Yakura and Hongming Zhang and Hugh Mee Wong and Ian Ng and Isaac Noble and Jaap Jumelet and Jack Geissinger and Jackson Kernion and Jacob Hilton and Jaehoon Lee and Jaime Fernández Fisac and James B. Simon and James Koppel and James Zheng and James Zou and Jan Kocoń and Jana Thompson and Janelle Wingfield and Jared Kaplan and Jarema Radom and Jascha Sohl-Dickstein and Jason Phang and Jason Wei and Jason Yosinski and Jekaterina Novikova and Jelle Bosscher and Jennifer Marsh and Jeremy Kim and Jeroen Taal and Jesse Engel and Jesujoba Alabi and Jiacheng Xu and Jiaming Song and Jillian Tang and Joan Waweru and John Burden and John Miller and John U. Balis and Jonathan Batchelder and Jonathan Berant and Jörg Frohberg and Jos Rozen and Jose Hernandez-Orallo and Joseph Boudeman and Joseph Guerr and Joseph Jones and Joshua B. Tenenbaum and Joshua S. Rule and Joyce Chua and Kamil Kanclerz and Karen Livescu and Karl Krauth and Karthik Gopalakrishnan and Katerina Ignatyeva and Katja Markert and Kaustubh D. Dhole and Kevin Gimpel and Kevin Omondi and Kory Mathewson and Kristen Chiafullo and Ksenia Shkaruta and Kumar Shridhar and Kyle McDonell and Kyle Richardson and Laria Reynolds and Leo Gao and Li Zhang and Liam Dugan and Lianhui Qin and Lidia Contreras-Ochando and Louis-Philippe Morency and Luca Moschella and Lucas Lam and Lucy Noble and Ludwig Schmidt and Luheng He and Luis Oliveros Colón and Luke Metz and Lütfi Kerem Şenel and Maarten Bosma and Maarten Sap and Maartje ter Hoeve and Maheen Farooqi and Manaal Faruqui and Mantas Mazeika and Marco Baturan and Marco Marelli and Marco Maru and Maria Jose Ramírez Quintana and Marie Tolkiehn and Mario Giulianelli and Martha Lewis and Martin Potthast and Matthew L. Leavitt and Matthias Hagen and Mátyás Schubert and Medina Orduna Baitemirova and Melody Arnaud and Melvin McElrath and Michael A. Yee and Michael Cohen and Michael Gu and Michael Ivanitskiy and Michael Starritt and Michael Strube and Michał Swędrowski and Michele Bevilacqua and Michihiro Yasunaga and Mihir Kale and Mike Cain and Mimee Xu and Mirac Suzgun and Mitch Walker and Mo Tiwari and Mohit Bansal and Moin Aminnaseri and Mor Geva and Mozhdeh Gheini and Mukund Varma T and Nanyun Peng and Nathan A. Chi and Nayeon Lee and Neta Gur-Ari Krakover and Nicholas Cameron and Nicholas Roberts and Nick Doiron and Nicole Martinez and Nikita Nangia and Niklas Deckers and Niklas Muennighoff and Nitish Shirish Keskar and Niveditha S. Iyer and Noah Constant and Noah Fiedel and Nuan Wen and Oliver Zhang and Omar Agha and Omar Elbaghdadi and Omer Levy and Owain Evans and Pablo Antonio Moreno Casares and Parth Doshi and Pascale Fung and Paul Pu Liang and Paul Vicol and Pegah Alipoormolabashi and Peiyuan Liao and Percy Liang and Peter Chang and Peter Eckersley and Phu Mon Htut and Pinyu Hwang and Piotr Miłkowski and Piyush Patil and Pouya Pezeshkpour and Priti Oli and Qiaozhu Mei and Qing Lyu and Qinlang Chen and Rabin Banjade and Rachel Etta Rudolph and Raefer Gabriel and Rahel Habacker and Ramon Risco and Raphaël Millière and Rhythm Garg and Richard Barnes and Rif A. Saurous and Riku Arakawa and Robbe Raymaekers and Robert Frank and Rohan Sikand and Roman Novak and Roman Sitelew and Ronan LeBras and Rosanne Liu and Rowan Jacobs and Rui Zhang and Ruslan Salakhutdinov and Ryan Chi and Ryan Lee and Ryan Stovall and Ryan Teehan and Rylan Yang and Sahib Singh and Saif M. Mohammad and Sajant Anand and Sam Dillavou and Sam Shleifer and Sam Wiseman and Samuel Gruetter and Samuel R. Bowman and Samuel S. Schoenholz and Sanghyun Han and Sanjeev Kwatra and Sarah A. Rous and Sarik Ghazarian and Sayan Ghosh and Sean Casey and Sebastian Bischoff and Sebastian Gehrmann and Sebastian Schuster and Sepideh Sadeghi and Shadi Hamdan and Sharon Zhou and Shashank Srivastava and Sherry Shi and Shikhar Singh and Shima Asaadi and Shixiang Shane Gu and Shubh Pachchigar and Shubham Toshniwal and Shyam Upadhyay and Shyamolima and Debnath and Siamak Shakeri and Simon Thormeyer and Simone Melzi and Siva Reddy and Sneha Priscilla Makini and Soo-Hwan Lee and Spencer Torene and Sriharsha Hatwar and Stanislas Dehaene and Stefan Divic and Stefano Ermon and Stella Biderman and Stephanie Lin and Stephen Prasad and Steven T. Piantadosi and Stuart M. Shieber and Summer Misherghi and Svetlana Kiritchenko and Swaroop Mishra and Tal Linzen and Tal Schuster and Tao Li and Tao Yu and Tariq Ali and Tatsu Hashimoto and Te-Lin Wu and Théo Desbordes and Theodore Rothschild and Thomas Phan and Tianle Wang and Tiberius Nkinyili and Timo Schick and Timofei Kornev and Titus Tunduny and Tobias Gerstenberg and Trenton Chang and Trishala Neeraj and Tushar Khot and Tyler Shultz and Uri Shaham and Vedant Misra and Vera Demberg and Victoria Nyamai and Vikas Raunak and Vinay Ramasesh and Vinay Uday Prabhu and Vishakh Padmakumar and Vivek Srikumar and William Fedus and William Saunders and William Zhang and Wout Vossen and Xiang Ren and Xiaoyu Tong and Xinran Zhao and Xinyi Wu and Xudong Shen and Yadollah Yaghoobzadeh and Yair Lakretz and Yangqiu Song and Yasaman Bahri and Yejin Choi and Yichi Yang and Yiding Hao and Yifu Chen and Yonatan Belinkov and Yu Hou and Yufang Hou and Yuntao Bai and Zachary Seid and Zhuoye Zhao and Zijian Wang and Zijie J. Wang and Zirui Wang and Ziyi Wu},
      year={2023},
      eprint={2206.04615},
      archivePrefix={arXiv},
      primaryClass={cs.CL},
      url={https://arxiv.org/abs/2206.04615}, 
}

@misc{w3c_dcat3_2024,
  title        = {{Data Catalog Vocabulary (DCAT) -- Version 3}},
  author       = {{W3C Dataset Exchange Working Group}},
  year         = {2024},
  howpublished = {W3C Recommendation},
  url          = {https://www.w3.org/TR/vocab-dcat-3/},
  note         = {Accessed 2026-05-01}
}

@misc{schemaorg_dataset,
  title        = {{Dataset - Schema.org Type}},
  author       = {{Schema.org}},
  howpublished = {Schema.org vocabulary documentation},
  url          = {https://schema.org/Dataset},
  note         = {Accessed 2026-05-01},
    year    = {2026},
}

@article{gebru2021datasheets,
  title   = {Datasheets for Datasets},
  author  = {Timnit Gebru and Jamie Morgenstern and Briana Vecchione and Jennifer Wortman Vaughan and Hanna Wallach and Hal Daum{\'e} III and Kate Crawford},
  journal = {Communications of the ACM},
  volume  = {64},
  number  = {12},
  pages   = {86--92},
  year    = {2021},
  doi     = {10.1145/3458723}
}

@inproceedings{mitchell2019modelcards,
  title     = {Model Cards for Model Reporting},
  author    = {Margaret Mitchell and Simone Wu and Andrew Zaldivar and Parker Barnes and Lucy Vasserman and Ben Hutchinson and Elena Spitzer and Inioluwa Deborah Raji and Timnit Gebru},
  booktitle = {Proceedings of the Conference on Fairness, Accountability, and Transparency},
  pages     = {220--229},
  year      = {2019},
  publisher = {Association for Computing Machinery},
  doi       = {10.1145/3287560.3287596}
}

@inproceedings{wang2019glue,
  title     = {{GLUE}: A Multi-Task Benchmark and Analysis Platform for Natural Language Understanding},
  author    = {Alex Wang and Amanpreet Singh and Julian Michael and Felix Hill and Omer Levy and Samuel R. Bowman},
  booktitle = {International Conference on Learning Representations},
  year      = {2019},
  url       = {https://openreview.net/forum?id=rJ4km2R5t7}
}

@inproceedings{wang2019superglue,
  title     = {{SuperGLUE}: A Stickier Benchmark for General-Purpose Language Understanding Systems},
  author    = {Alex Wang and Yada Pruksachatkun and Nikita Nangia and Amanpreet Singh and Julian Michael and Felix Hill and Omer Levy and Samuel R. Bowman},
  booktitle = {Advances in Neural Information Processing Systems},
  year      = {2019},
  url       = {https://arxiv.org/abs/1905.00537}
}

@inproceedings{reddi2020mlperf,
  title     = {{MLPerf} Inference Benchmark},
  author    = {Vijay Janapa Reddi and Christine Cheng and David Kanter and Peter Mattson and Guenther Schmuelling and Carole-Jean Wu and Brian Anderson and Maximilien Breughe and Mark Charlebois and William Chou and Ramesh Chukka and Cody Coleman and Sam Davis and Pan Deng and Greg Diamos and Jared Duke and Dave Fick and J. Scott Gardner and Itay Hubara and Sachin Idgunji and Thomas B. Jablin and Jeff Jiao and Tom St. John and Pankaj Kanwar and David Lee and Jeffery Liao and Anton Lokhmotov and Francisco Massa and Peng Meng and Paulius Micikevicius and Colin Osborne and Gennady Pekhimenko and Arun Tejusve Raghunath Rajan and Dilip Sequeira and Ashish Sirasao and Fei Sun and Hanlin Tang and Michael Thomson and Frank Wei and Ephrem Wu and Lingjie Xu and Koichi Yamada and Bing Yu and George Yuan and Aaron Zhong and Peizhao Zhang and Yuchen Zhou},
  booktitle = {Proceedings of the ACM/IEEE 47th Annual International Symposium on Computer Architecture},
  year      = {2020},
  doi       = {10.1109/ISCA45697.2020.00045},
  url       = {https://arxiv.org/abs/1911.02549}
}

@inproceedings{pmlr-v202-biderman23a,
  title     = {Pythia: A Suite for Analyzing Large Language Models Across Training and Scaling},
  author    = {Biderman, Stella and Schoelkopf, Hailey and Anthony, Quentin Gregory and Bradley, Herbie and O'Brien, Kyle and Hallahan, Eric and Khan, Mohammad Aflah and Purohit, Shivanshu and Prashanth, Usvsn Sai and Raff, Edward and Skowron, Aviya and Sutawika, Lintang and Van Der Wal, Oskar},
  booktitle = {Proceedings of the 40th International Conference on Machine Learning},
  pages     = {2397--2430},
  year      = {2023},
  editor    = {Krause, Andreas and Brunskill, Emma and Cho, Kyunghyun and Engelhardt, Barbara and Sabato, Sivan and Scarlett, Jonathan},
  volume    = {202},
  series    = {Proceedings of Machine Learning Research},
  month     = {23--29 Jul},
  publisher = {PMLR},
  url       = {https://proceedings.mlr.press/v202/biderman23a.html}
}

@article{kopcke2010evaluation,
  title={Evaluation of Entity Resolution Approaches on Real-World Match Problems},
  author={K{\"o}pcke, Hanna and Thor, Andreas and Rahm, Erhard},
  journal={Proceedings of the VLDB Endowment},
  volume={3},
  number={1--2},
  pages={484--493},
  year={2010},
  doi={10.14778/1920841.1920904}
}

@inproceedings{clark2020transformers,
  title     = {Transformers as Soft Reasoners over Language},
  author    = {Clark, Peter and Tafjord, Oyvind and Richardson, Kyle},
  booktitle = {Proceedings of the Twenty-Ninth International Joint Conference on Artificial Intelligence},
  pages     = {3882--3890},
  year      = {2020},
  doi       = {10.24963/ijcai.2020/537},
  url       = {https://arxiv.org/abs/2002.05867}
}

@inproceedings{petroni-etal-2019-language,
  title     = {Language Models as Knowledge Bases?},
  author    = {Petroni, Fabio and Rockt{\"a}schel, Tim and Riedel, Sebastian
               and Lewis, Patrick and Bakhtin, Anton and Wu, Yuxiang
               and Miller, Alexander},
  booktitle = {Proceedings of the 2019 Conference on Empirical Methods in
               Natural Language Processing and the 9th International Joint
               Conference on Natural Language Processing},
  pages     = {2463--2473},
  year      = {2019},
  publisher = {Association for Computational Linguistics},
  doi       = {10.18653/v1/D19-1250},
  url       = {https://aclanthology.org/D19-1250/}
}

@misc{vicuna2023,
    title = {Vicuna: An Open-Source Chatbot Impressing GPT-4 with 90\%* ChatGPT Quality},
    url = {https://lmsys.org/blog/2023-03-30-vicuna/},
    author = {Chiang, Wei-Lin and Li, Zhuohan and Lin, Zi and Sheng, Ying and Wu, Zhanghao and Zhang, Hao and Zheng, Lianmin and Zhuang, Siyuan and Zhuang, Yonghao and Gonzalez, Joseph E. and Stoica, Ion and Xing, Eric P.},
    month = {March},
    year = {2023}
}

@misc{almazrouei2023falcon,
  title         = {The Falcon Series of Open Language Models},
  author        = {Almazrouei, Ebtesam and Alobeidli, Hamza and Alshamsi, Abdulaziz and Cappelli, Alessandro and Cojocaru, Ruxandra and Debbah, M{\'e}rouane and Goffinet, {\'E}tienne and Hesslow, Daniel and Launay, Julien and Malartic, Quentin and Mazzotta, Daniele and Noune, Badreddine and Pannier, Baptiste and Penedo, Guilherme},
  year          = {2023},
  eprint        = {2311.16867},
  archivePrefix = {arXiv},
  primaryClass  = {cs.CL},
  doi           = {10.48550/arXiv.2311.16867},
  url           = {https://arxiv.org/abs/2311.16867}
}

@inproceedings{clark-etal-2019-boolq,
  title     = "{B}ool{Q}: Exploring the Surprising Difficulty of Natural Yes/No Questions",
  author    = "Clark, Christopher and Lee, Kenton and Chang, Ming-Wei and
               Kwiatkowski, Tom and Collins, Michael and Toutanova, Kristina",
  booktitle = "Proceedings of the 2019 Conference of the North American Chapter
               of the Association for Computational Linguistics: Human Language
               Technologies, Volume 1 (Long and Short Papers)",
  pages     = "2924--2936",
  year      = "2019",
  publisher = "Association for Computational Linguistics",
  url       = "https://aclanthology.org/N19-1300/"
}

@article{borkan2019nuanced,
  title   = {Nuanced Metrics for Measuring Unintended Bias with Real Data for Text Classification},
  author  = {Borkan, Daniel and Dixon, Lucas and Sorensen, Jeffrey and Thain, Nithum and Vasserman, Lucy},
  journal = {arXiv preprint arXiv:1903.04561},
  year    = {2019},
  url     = {https://arxiv.org/abs/1903.04561}
}

@inproceedings{mei2021capturing,
  title     = {Capturing Semantics for Imputation with Pre-trained Language Models},
  author    = {Mei, Yinan and Song, Shaoxu and Fang, Chenguang and Yang, Haifeng and Fang, Jingyun and Long, Jiang},
  booktitle = {2021 IEEE 37th International Conference on Data Engineering (ICDE)},
  pages     = {61--72},
  year      = {2021},
  publisher = {IEEE},
  doi       = {10.1109/ICDE51399.2021.00013},
  url       = {https://doi.org/10.1109/ICDE51399.2021.00013}
}

@misc{cobbe2021training,
  title         = {Training Verifiers to Solve Math Word Problems},
  author        = {Cobbe, Karl and Kosaraju, Vineet and Bavarian, Mohammad and
                   Chen, Mark and Jun, Heewoo and Kaiser, Lukasz and Plappert,
                   Matthias and Tworek, Jerry and Hilton, Jacob and Nakano,
                   Reiichiro and Hesse, Christopher and Schulman, John},
  year          = {2021},
  eprint        = {2110.14168},
  archivePrefix = {arXiv},
  primaryClass  = {cs.LG},
  url           = {https://arxiv.org/abs/2110.14168}
}

@inproceedings{maas-etal-2011-learning,
  title     = {Learning Word Vectors for Sentiment Analysis},
  author    = {Maas, Andrew L. and Daly, Raymond E. and Pham, Peter T. and
               Huang, Dan and Ng, Andrew Y. and Potts, Christopher},
  booktitle = {Proceedings of the 49th Annual Meeting of the Association for
               Computational Linguistics: Human Language Technologies},
  pages     = {142--150},
  year      = {2011},
  publisher = {Association for Computational Linguistics},
  url       = {https://aclanthology.org/P11-1015/}
}

@misc{zhong2021arlsat,
  title         = {{AR-LSAT}: Investigating Analytical Reasoning of Text},
  author        = {Zhong, Wanjun and Wang, Siyuan and Tang, Duyu and Xu, Zenan
                   and Guo, Daya and Wang, Jiahai and Yin, Jian and Zhou, Ming
                   and Duan, Nan},
  year          = {2021},
  eprint        = {2104.06598},
  archivePrefix = {arXiv},
  primaryClass  = {cs.CL},
  url           = {https://arxiv.org/abs/2104.06598}
}

@article{kocisky-etal-2018-narrativeqa,
  title   = {The {N}arrative{QA} Reading Comprehension Challenge},
  author  = {Ko{\v{c}}isk{\'y}, Tom{\'a}{\v{s}} and Schwarz, Jonathan and
             Blunsom, Phil and Dyer, Chris and Hermann, Karl Moritz and Melis,
             G{\'a}bor and Grefenstette, Edward},
  journal = {Transactions of the Association for Computational Linguistics},
  volume  = {6},
  pages   = {317--328},
  year    = {2018},
  doi     = {10.1162/tacl_a_00023},
  url     = {https://aclanthology.org/Q18-1023/}
}

@inproceedings{choi-etal-2018-quac,
  title     = "{Q}u{AC}: Question Answering in Context",
  author    = "Choi, Eunsol and He, He and Iyyer, Mohit and Yatskar, Mark and
               Yih, Wen-tau and Choi, Yejin and Liang, Percy and Zettlemoyer, Luke",
  booktitle = "Proceedings of the 2018 Conference on Empirical Methods in Natural Language Processing",
  pages     = "2174--2184",
  year      = "2018",
  publisher = "Association for Computational Linguistics",
  doi       = "10.18653/v1/D18-1241",
  url       = "https://aclanthology.org/D18-1241/"
}

@inproceedings{wu2021lime,
  title     = {{LIME}: Learning Inductive Bias for Primitives of Mathematical Reasoning},
  author    = {Wu, Yuhuai and Rabe, Markus N. and Li, Wenda and Ba, Jimmy and
               Grosse, Roger B. and Szegedy, Christian},
  booktitle = {Proceedings of the 38th International Conference on Machine Learning},
  series    = {Proceedings of Machine Learning Research},
  volume    = {139},
  pages     = {11251--11262},
  year      = {2021},
  publisher = {PMLR},
  url       = {https://proceedings.mlr.press/v139/wu21c.html}
}

@inproceedings{lin-etal-2022-truthfulqa,
  title     = "{T}ruthful{QA}: Measuring How Models Mimic Human Falsehoods",
  author    = "Lin, Stephanie and Hilton, Jacob and Evans, Owain",
  booktitle = "Proceedings of the 60th Annual Meeting of the Association for
               Computational Linguistics (Volume 1: Long Papers)",
  pages     = "3214--3252",
  year      = "2022",
  publisher = "Association for Computational Linguistics",
  doi       = "10.18653/v1/2022.acl-long.229",
  url       = "https://aclanthology.org/2022.acl-long.229/"
}

@inproceedings{rein2024gpqa,
      title={{GPQA}: A Graduate-Level Google-Proof Q\&A Benchmark},
      author={David Rein and Betty Li Hou and Asa Cooper Stickland and Jackson Petty and Richard Yuanzhe Pang and Julien Dirani and Julian Michael and Samuel R. Bowman},
      booktitle={First Conference on Language Modeling},
      year={2024},
      url={https://openreview.net/forum?id=Ti67584b98}
}

@inproceedings{murphy2025ai,
  title={AI-Assisted Wordle Demo: Combining LLMs and Rule-Based Solvers for Enhanced Gameplay},
  author={Murphy, Colin and Liu, Chang},
  booktitle={2025 IEEE Conference on Games (CoG)},
  pages={1--2},
  year={2025},
  organization={IEEE}
}

@article{Tan2024JudgeBenchAB,
  title={JudgeBench: A Benchmark for Evaluating LLM-based Judges},
  author={Sijun Tan and Siyuan Zhuang and Kyle Montgomery and William Y. Tang and Alejandro Cuadron and Chenguang Wang and Raluca A. Popa and Ion Stoica},
  journal={ArXiv},
  year={2024},
  volume={abs/2410.12784},
  url={https://api.semanticscholar.org/CorpusID:273374769}
}

@misc{merrill2026terminalbenchbenchmarkingagentshard,
      title={Terminal-Bench: Benchmarking Agents on Hard, Realistic Tasks in Command Line Interfaces}, 
      author={Mike A. Merrill and Alexander G. Shaw and Nicholas Carlini and Boxuan Li and Harsh Raj and Ivan Bercovich and Lin Shi and Jeong Yeon Shin and Thomas Walshe and E. Kelly Buchanan and Junhong Shen and Guanghao Ye and Haowei Lin and Jason Poulos and Maoyu Wang and Marianna Nezhurina and Jenia Jitsev and Di Lu and Orfeas Menis Mastromichalakis and Zhiwei Xu and Zizhao Chen and Yue Liu and Robert Zhang and Leon Liangyu Chen and Anurag Kashyap and Jan-Lucas Uslu and Jeffrey Li and Jianbo Wu and Minghao Yan and Song Bian and Vedang Sharma and Ke Sun and Steven Dillmann and Akshay Anand and Andrew Lanpouthakoun and Bardia Koopah and Changran Hu and Etash Guha and Gabriel H. S. Dreiman and Jiacheng Zhu and Karl Krauth and Li Zhong and Niklas Muennighoff and Robert Amanfu and Shangyin Tan and Shreyas Pimpalgaonkar and Tushar Aggarwal and Xiangning Lin and Xin Lan and Xuandong Zhao and Yiqing Liang and Yuanli Wang and Zilong Wang and Changzhi Zhou and David Heineman and Hange Liu and Harsh Trivedi and John Yang and Junhong Lin and Manish Shetty and Michael Yang and Nabil Omi and Negin Raoof and Shanda Li and Terry Yue Zhuo and Wuwei Lin and Yiwei Dai and Yuxin Wang and Wenhao Chai and Shang Zhou and Dariush Wahdany and Ziyu She and Jiaming Hu and Zhikang Dong and Yuxuan Zhu and Sasha Cui and Ahson Saiyed and Arinbjörn Kolbeinsson and Jesse Hu and Christopher Michael Rytting and Ryan Marten and Yixin Wang and Alex Dimakis and Andy Konwinski and Ludwig Schmidt},
      year={2026},
      eprint={2601.11868},
      archivePrefix={arXiv},
      primaryClass={cs.SE},
      url={https://arxiv.org/abs/2601.11868}, 
}

@article{meng2025psychology,
  title={A Psychology-based Unified Dynamic Framework for Curriculum Learning},
  author={Meng, Guangyu and Zeng, Qinkai and Lalor, John P and Yu, Hong},
  journal={Computational Linguistics},
  pages={1--49},
  year={2025},
  publisher={MIT Press 255 Main Street, 9th Floor, Cambridge, Massachusetts 02142, USA~…}
}

\FloatBarrier
\vspace{40pt}
\appendix

{\Large{\bfseries Appendix}}
% \vspace{8pt}

%
\begin{tcolorbox}[
  enhanced,
  breakable,
  colback   = white,
  colframe  = tocrulecolor,
  boxrule   = 0.8pt,
  arc       = 3pt,
  title     = {\textbf{Table of Contents}},
  fonttitle = \color{white},
  coltitle  = white,
  attach boxed title to top left = {yshift=-2mm, xshift=4mm},
  boxed title style = {colback=toctitlecolor, arc=2pt, boxrule=0pt},
  left=6pt, right=6pt, top=6pt, bottom=6pt,
]

\apptocline{app:stats}{A\quad Summary Statistics Every Eval Ever Datastore}{\pageref{app:stats}}
\apptocline{app:stats}{A.1\quad Overview of inference platform distribution}{\pageref{app:stat-tab1}}
\apptocline{app:stats}{A.2\quad Overview of the top 25 models in \eee{}}{\pageref{app:stat-tab2}}
\apptocline{app:stats}{A.3\quad Overview of evaluation runs by source organization}{\pageref{app:stat-tab3}}
\apptocline{app:stats}{A.4\quad Overview of evaluation activity across the top 25 benchmarks}{\pageref{app:stat-tab4}}

\apptocline{sec:app-schema}{B\quad Full Every Eval Ever Schema Field Reference}{\pageref{sec:app-schema}}

\apptocline{sec:app-converters}{C\quad Converter Implementation Details}{\pageref{sec:app-converters}}
\apptocsubline{sec:app-conv-helm}{C.1\quad HELM Converter}{\pageref{sec:app-conv-helm}}
\apptocsubline{sec:app-conv-lmeval}{C.2\quad lm-eval-harness Converter}{\pageref{sec:app-conv-lmeval}}
\apptocsubline{sec:app-conv-inspect}{C.3\quad Inspect AI Converter}{\pageref{sec:app-conv-inspect}}
\apptocsubline{sec:app-comm}{C.4\quad Community Converters}{\pageref{sec:app-comm}}

\apptocline{app:cost}{D\quad Conservative Estimation of Costs}{\pageref{app:cost}}

\apptocline{app:governance}{E\quad Governance Card}{\pageref{app:governance}}

\apptocline{app:casestudies}{F\quad Case Studies Reproducibility and Implementation Details}{\pageref{app:casestudies}}
\apptocline{app:casestudies1}{F.1\quad Case Study 1: Cost-Accuracy Tradeoffs in Agentic Evals}{\pageref{app:casestudies1}}
\apptocline{app:casestudies2}{F.2\quad Case Study 2: Version-Dependent Perplexity Metrics}{\pageref{app:casestudies2}}
\apptocline{app:case3}{F.3\quad Case Study 3: Evaluation Reproducibility Gaps}{\pageref{app:case3}}
\apptocline{app:casestudies4}{F.4\quad Case Study 4: Evaluation Meta-Analysis using Item Response Theory}{\pageref{app:casestudies4}}
\end{tcolorbox}
\newpage
\section{Summary Statistics Every Eval Ever Datastore}
\label{app:stats}

We present a high level summary and breakdown of key fields in \eee{} in
% We count how much information was gathered, such as unique number of models, in Table~\ref{tab:schema_statistics}. We also count how much information we see on each category (e.g., benchmark or model) 
 Tables \ref{tab:inference_platform_consolidated}, 
% \ref{tab:aggregate_evaluation_record}, \ref{tab:instance_level_evaluation_record}, \ref{tab:evaluation_result_record}
\ref{tab:model_evaluation_shaded}, 
\ref{tab:org_evaluation_results}, \ref{tab:benchmark_evaluation_activity}.

% \begin{table}[b]
% \caption{Summary statistics for key \eee{} fields. This table provides an overview of the scale and diversity of the data, including the total number of result rows, unique models, and benchmarks represented.}
% \label{tab:schema_statistics}
% \centering
% \small
% \setlength{\tabcolsep}{10pt}
% \renewcommand{\arraystretch}{1.2}
% \begin{tabular}{llr}
% \toprule
% \textbf{Statistic} & \textbf{Value} \\
% \midrule
% Evaluations                 & 229572 \\
% % Schema-Level Data           & 38020  \\
% Unique Models               & 22235  \\
% Unique Benchmarks           & 2273   \\
% % Instance-Level Data         & 516    \\
% Eval Data Pull Requests     & 135    \\
% Code Pull Requests          & 127    \\
% Unique Evaluation Harnesses & 31     \\
% Unique Source Organizations & 28     \\
% \bottomrule
% % \vspace{0.1mm}
% \end{tabular}
% \end{table}

\subsection{Overview of inference platform distribution}
\label{app:stat-tab1}
\begin{table}[h!]
\caption{Inference platform distribution by evaluation runs and model diversity. Over 98\% of runs fall under \textit{Unreported} or \textit{Unknown} categories. Among identified providers, Ollama and OpenAI lead in both volume and unique model representation.}
\vspace{20pt}
\label{tab:inference_platform_consolidated}
\centering
\small
\setlength{\tabcolsep}{6pt}
\renewcommand{\arraystretch}{1.2}
\begin{tabular}{l rr rr}
\toprule
\textbf{Inference Platform} & \multicolumn{2}{l}{\textbf{Eval. Runs}} & \multicolumn{2}{l}{\textbf{Models}} \\
\midrule
\rowcolor{gray!10} Unreported & 184{,}929 & \textcolor{cyan}{\rule{12.0pt}{6pt}} (80.56\%) & 17{,}101 & \textcolor{cyan}{\rule{12.0pt}{6pt}} (75.71\%) \\
Unknown & 42{,}260 & \textcolor{cyan}{\rule{2.7pt}{6pt}} (18.41\%) & 5{,}282 & \textcolor{cyan}{\rule{3.7pt}{6pt}} (23.38\%) \\
\midrule
\rowcolor{gray!10} Ollama & 849 & \textcolor{cyan}{\rule{0.5pt}{6pt}} (0.37\%) & 55 & \textcolor{cyan}{\rule{0.5pt}{6pt}} (0.24\%) \\
OpenAI & 411 & \textcolor{cyan}{\rule{0.5pt}{6pt}} (0.18\%) & 32 & \textcolor{cyan}{\rule{0.5pt}{6pt}} (0.14\%) \\
\rowcolor{gray!10} Google & 312 & \textcolor{cyan}{\rule{0.5pt}{6pt}} (0.14\%) & 28 & \textcolor{cyan}{\rule{0.5pt}{6pt}} (0.12\%) \\
Together & 191 & \textcolor{cyan}{\rule{0.5pt}{6pt}} (0.08\%) & 13 & \textcolor{cyan}{\rule{0.5pt}{6pt}} (0.06\%) \\
\rowcolor{gray!10} Anthropic & 152 & \textcolor{cyan}{\rule{0.5pt}{6pt}} (0.07\%) & 16 & \textcolor{cyan}{\rule{0.5pt}{6pt}} (0.07\%) \\
Mistral & 100 & \textcolor{cyan}{\rule{0.5pt}{6pt}} (0.04\%) & 10 & \textcolor{cyan}{\rule{0.5pt}{6pt}} (0.04\%) \\
\rowcolor{gray!10} DeepSeek & 56 & \textcolor{cyan}{\rule{0.5pt}{6pt}} (0.02\%) & 4 & \textcolor{cyan}{\rule{0.5pt}{6pt}} (0.02\%) \\
Cohere & 48 & \textcolor{cyan}{\rule{0.5pt}{6pt}} (0.02\%) & 5 & \textcolor{cyan}{\rule{0.5pt}{6pt}} (0.02\%) \\
\rowcolor{gray!10} xAI & 31 & \textcolor{cyan}{\rule{0.5pt}{6pt}} (0.01\%) & 3 & \textcolor{cyan}{\rule{0.5pt}{6pt}} (0.01\%) \\
OpenRouter & 30 & \textcolor{cyan}{\rule{0.5pt}{6pt}} (0.01\%) & 10 & \textcolor{cyan}{\rule{0.5pt}{6pt}} (0.04\%) \\
\rowcolor{gray!10} AWS & 30 & \textcolor{cyan}{\rule{0.5pt}{6pt}} (0.01\%) & 3 & \textcolor{cyan}{\rule{0.5pt}{6pt}} (0.01\%) \\
Gemini & 27 & \textcolor{cyan}{\rule{0.5pt}{6pt}} (0.01\%) & 3 & \textcolor{cyan}{\rule{0.5pt}{6pt}} (0.01\%) \\
\rowcolor{gray!10} Aliyun & 22 & \textcolor{cyan}{\rule{0.5pt}{6pt}} (0.01\%) & 5 & \textcolor{cyan}{\rule{0.5pt}{6pt}} (0.02\%) \\
Perplexity & 21 & \textcolor{cyan}{\rule{0.5pt}{6pt}} (0.01\%) & 2 & \textcolor{cyan}{\rule{0.5pt}{6pt}} (0.01\%) \\
\rowcolor{gray!10} Local & 20 & \textcolor{cyan}{\rule{0.5pt}{6pt}} (0.01\%) & 1 & \textcolor{cyan}{\rule{0.5pt}{6pt}} (0.00\%) \\
Ark & 15 & \textcolor{cyan}{\rule{0.5pt}{6pt}} (0.01\%) & 5 & \textcolor{cyan}{\rule{0.5pt}{6pt}} (0.02\%) \\
\rowcolor{gray!10} Moonshot & 13 & \textcolor{cyan}{\rule{0.5pt}{6pt}} (0.01\%) & 2 & \textcolor{cyan}{\rule{0.5pt}{6pt}} (0.01\%) \\
MiniMax & 12 & \textcolor{cyan}{\rule{0.5pt}{6pt}} (0.01\%) & 1 & \textcolor{cyan}{\rule{0.5pt}{6pt}} (0.00\%) \\
\rowcolor{gray!10} StepFun & 11 & \textcolor{cyan}{\rule{0.5pt}{6pt}} (0.00\%) & 1 & \textcolor{cyan}{\rule{0.5pt}{6pt}} (0.00\%) \\
Qwen & 10 & \textcolor{cyan}{\rule{0.5pt}{6pt}} (0.00\%) & 2 & \textcolor{cyan}{\rule{0.5pt}{6pt}} (0.01\%) \\
\rowcolor{gray!10} Tencent & 10 & \textcolor{cyan}{\rule{0.5pt}{6pt}} (0.00\%) & 1 & \textcolor{cyan}{\rule{0.5pt}{6pt}} (0.00\%) \\
Zhipu & 9 & \textcolor{cyan}{\rule{0.5pt}{6pt}} (0.00\%) & 1 & \textcolor{cyan}{\rule{0.5pt}{6pt}} (0.00\%) \\
\rowcolor{gray!10} Kuaishou & 3 & \textcolor{cyan}{\rule{0.5pt}{6pt}} (0.00\%) & 1 & \textcolor{cyan}{\rule{0.5pt}{6pt}} (0.00\%) \\
\bottomrule
% \vspace{0.1mm}
\end{tabular}
\end{table}

\newpage
\subsection{Overview of the top 25 models in \eee{}}
\label{app:stat-tab2}
\begin{table}[h!]
\caption{This shows the breakdown of the top 25 models in \eee{} and the total number of evaluations across all runs. The data highlights a strong concentration of evaluation runs for the GPT-4 family, while also showing the emergence of frontier models like DeepSeek-R1 and Gemini-3 previews in current evaluation cycles.}
\vspace{20pt}
\label{tab:model_evaluation_shaded}
\centering
\small
\setlength{\tabcolsep}{10pt}
\renewcommand{\arraystretch}{1.2}
\begin{tabular}{l r @{\hspace{10pt}} l}
\toprule
\textbf{Models} & \multicolumn{2}{l}{\textbf{Eval. Runs}} \\
\midrule
\rowcolor{gray!10} GPT-4o & 4{,}443 & \textcolor{magenta}{\rule{12.0pt}{6pt}} (1.94\%) \\
GPT-4 & 1{,}211 & \textcolor{magenta}{\rule{3.3pt}{6pt}} (0.53\%) \\
\rowcolor{gray!10} Gemini-1.5-Pro & 1{,}007 & \textcolor{magenta}{\rule{2.7pt}{6pt}} (0.44\%) \\
GPT-4o-mini & 914 & \textcolor{magenta}{\rule{2.5pt}{6pt}} (0.40\%) \\
\rowcolor{gray!10} DeepSeek-R1 & 788 & \textcolor{magenta}{\rule{2.1pt}{6pt}} (0.34\%) \\
GPT-4.1 & 709 & \textcolor{magenta}{\rule{1.9pt}{6pt}} (0.31\%) \\
\rowcolor{gray!10} Claude-3.5-Sonnet & 610 & \textcolor{magenta}{\rule{1.6pt}{6pt}} (0.27\%) \\
Claude 3.5 Sonnet & 604 & \textcolor{magenta}{\rule{1.6pt}{6pt}} (0.26\%) \\
\rowcolor{gray!10} DeepSeek-V3 & 588 & \textcolor{magenta}{\rule{1.6pt}{6pt}} (0.26\%) \\
GPT-3.5 & 575 & \textcolor{magenta}{\rule{1.6pt}{6pt}} (0.25\%) \\
\rowcolor{gray!10} Gemini-2.5-Pro & 568 & \textcolor{magenta}{\rule{1.5pt}{6pt}} (0.25\%) \\
GPT-4V & 559 & \textcolor{magenta}{\rule{1.5pt}{6pt}} (0.24\%) \\
\rowcolor{gray!10} Human & 545 & \textcolor{magenta}{\rule{1.5pt}{6pt}} (0.24\%) \\
GPT-3.5-Turbo & 475 & \textcolor{magenta}{\rule{1.3pt}{6pt}} (0.21\%) \\
\rowcolor{gray!10} o4-mini & 473 & \textcolor{magenta}{\rule{1.3pt}{6pt}} (0.21\%) \\
GPT-5 & 452 & \textcolor{magenta}{\rule{1.2pt}{6pt}} (0.20\%) \\
\rowcolor{gray!10} Gemini 2.5 Pro & 431 & \textcolor{magenta}{\rule{1.2pt}{6pt}} (0.19\%) \\
Gemini-2.0-Flash & 422 & \textcolor{magenta}{\rule{1.1pt}{6pt}} (0.18\%) \\
\rowcolor{gray!10} Qwen2.5-VL-7B & 417 & \textcolor{magenta}{\rule{1.1pt}{6pt}} (0.18\%) \\
o3 & 398 & \textcolor{magenta}{\rule{1.1pt}{6pt}} (0.17\%) \\
\rowcolor{gray!10} Qwen2.5-VL-72B & 366 & \textcolor{magenta}{\rule{1.0pt}{6pt}} (0.16\%) \\
o1-mini & 354 & \textcolor{magenta}{\rule{1.0pt}{6pt}} (0.15\%) \\
\rowcolor{gray!10} Qwen2.5-72B-Instruct & 349 & \textcolor{magenta}{\rule{0.9pt}{6pt}} (0.15\%) \\
Claude-3.7-Sonnet & 342 & \textcolor{magenta}{\rule{0.9pt}{6pt}} (0.15\%) \\
\rowcolor{gray!10} InternVL2.5-8B & 339 & \textcolor{magenta}{\rule{0.9pt}{6pt}} (0.15\%) \\
\bottomrule
% \vspace{0.1mm}
\end{tabular}
\end{table}

\newpage
\subsection{Overview of evaluation runs by source organization}
\label{app:stat-tab3}
\begin{table}[h!]
\caption{The table shows individual evaluation runs by source organization. The dataset is characterized by a significant volume of records from alphaXiv, exceeding 160,000 entries. Notably, this schema includes a university consortium comprising Princeton University, New York University, University of Washington, University of California San Diego, and Canyon Crest Academy, which together contribute to the diverse academic representation.*
\vspace{20pt}
\label{tab:org_evaluation_results}
}
\centering
\small
\setlength{\tabcolsep}{10pt}
\renewcommand{\arraystretch}{1.2}
\begin{tabular}{l r l}
\toprule
\textbf{Organizations} & \multicolumn{2}{l}{\textbf{Eval. Runs}} \\
\midrule
\rowcolor{gray!10} alphaXiv & 162{,}616 & \textcolor{orange!80!black}{\rule{12.0pt}{6pt}} (71.23\%) \\
Hugging Face & 27{,}444 & \textcolor{orange!80!black}{\rule{2.0pt}{6pt}} (12.02\%) \\
\rowcolor{gray!10} Valsai & 8{,}144 & \textcolor{orange!80!black}{\rule{0.6pt}{6pt}} (3.57\%) \\
Artificial Analysis & 8{,}104 & \textcolor{orange!80!black}{\rule{0.6pt}{6pt}} (3.55\%) \\
\rowcolor{gray!10} crfm & 4{,}712 & \textcolor{orange!80!black}{\rule{0.5pt}{6pt}} (2.06\%) \\
UC Berkeley Gorilla & 3{,}350 & \textcolor{orange!80!black}{\rule{0.5pt}{6pt}} (1.47\%) \\
\rowcolor{gray!10} LLM Stats & 3{,}002 & \textcolor{orange!80!black}{\rule{0.5pt}{6pt}} (1.32\%) \\
Allen Institute for AI & 2{,}404 & \textcolor{orange!80!black}{\rule{0.5pt}{6pt}} (1.05\%) \\
\rowcolor{gray!10} TIGERLab & 2{,}220 & \textcolor{orange!80!black}{\rule{0.5pt}{6pt}} (0.97\%) \\
LiveBench & 1{,}761 & \textcolor{orange!80!black}{\rule{0.5pt}{6pt}} (0.77\%) \\
\rowcolor{gray!10} HumanCentered Eval & 1{,}218 & \textcolor{orange!80!black}{\rule{0.5pt}{6pt}} (0.53\%) \\
ARC Prize & 1{,}020 & \textcolor{orange!80!black}{\rule{0.5pt}{6pt}} (0.45\%) \\
\rowcolor{gray!10} Wordle Arena Project & 1{,}002 & \textcolor{orange!80!black}{\rule{0.5pt}{6pt}} (0.44\%) \\
kaggle & 912 & \textcolor{orange!80!black}{\rule{0.5pt}{6pt}} (0.40\%) \\
\rowcolor{gray!10} Princeton SAgE Team & 342 & \textcolor{orange!80!black}{\rule{0.5pt}{6pt}} (0.15\%) \\
ByteDanceSeed & 261 & \textcolor{orange!80!black}{\rule{0.5pt}{6pt}} (0.11\%) \\
\rowcolor{gray!10} Chang Liu's Lab & 238 & \textcolor{orange!80!black}{\rule{0.5pt}{6pt}} (0.10\%) \\
SWEbench & 127 & \textcolor{orange!80!black}{\rule{0.5pt}{6pt}} (0.06\%) \\
\rowcolor{gray!10} Mercor & 125 & \textcolor{orange!80!black}{\rule{0.5pt}{6pt}} (0.05\%) \\
Ai2 & 114 & \textcolor{orange!80!black}{\rule{0.5pt}{6pt}} (0.05\%) \\
\rowcolor{gray!10} CapArena & 110 & \textcolor{orange!80!black}{\rule{0.5pt}{6pt}} (0.05\%) \\
Google DeepMind & 108 & \textcolor{orange!80!black}{\rule{0.5pt}{6pt}} (0.05\%) \\
\rowcolor{gray!10} Exgentic & 90 & \textcolor{orange!80!black}{\rule{0.5pt}{6pt}} (0.04\%) \\
Princeton, NYU, UW, UCSD, CCA* & 87 & \textcolor{orange!80!black}{\rule{0.5pt}{6pt}} (0.04\%) \\
\rowcolor{gray!10} CocoaBench & 40 & \textcolor{orange!80!black}{\rule{0.5pt}{6pt}} (0.02\%) \\
AmazonScience & 16 & \textcolor{orange!80!black}{\rule{0.5pt}{6pt}} (0.01\%) \\
\rowcolor{gray!10} La Leaderboard & 5 & \textcolor{orange!80!black}{\rule{0.5pt}{6pt}} (0.00\%) \\
\bottomrule
% \vspace{0.1mm}
\end{tabular}
\end{table}

\newpage
\subsection{Overview of evaluation activity across the top 25 benchmarks}
\label{app:stat-tab4}
\begin{table}[h!]
\caption{Distribution of evaluation activity across the top 25 most popular benchmarks. The data shows a high density of testing within the Artificial Analysis LLM API framework, followed by foundational reasoning and knowledge benchmarks such as GPQA, IFEval, and MMLU-PRO, reflecting their role as industry standards for model performance assessment.}
\vspace{20pt}
\label{tab:benchmark_evaluation_activity}
\centering
\small
\setlength{\tabcolsep}{10pt}
\renewcommand{\arraystretch}{1.2}
\begin{tabular}{l r l}
\toprule
\textbf{Benchmark} & \multicolumn{2}{l}{\textbf{Eval. Runs}} \\
\midrule
\rowcolor{gray!10} Artificial Analysis LLM API & 8{,}104 & \textcolor{violet}{\rule{12.0pt}{6pt}} (3.53\%) \\
GPQA & 4{,}662 & \textcolor{violet}{\rule{6.9pt}{6pt}} (2.03\%) \\
\rowcolor{gray!10} IFEval & 4{,}662 & \textcolor{violet}{\rule{6.9pt}{6pt}} (2.03\%) \\
BBH & 4{,}585 & \textcolor{violet}{\rule{6.8pt}{6pt}} (2.00\%) \\
\rowcolor{gray!10} MATH Level 5 & 4{,}574 & \textcolor{violet}{\rule{6.8pt}{6pt}} (1.99\%) \\
MMLU-PRO & 4{,}574 & \textcolor{violet}{\rule{6.8pt}{6pt}} (1.99\%) \\
\rowcolor{gray!10} MUSR & 4{,}574 & \textcolor{violet}{\rule{6.8pt}{6pt}} (1.99\%) \\
BFCL leaderboard CSV & 3{,}350 & \textcolor{violet}{\rule{5.0pt}{6pt}} (1.46\%) \\
\rowcolor{gray!10} helm\_mmlu & 2{,}844 & \textcolor{violet}{\rule{4.2pt}{6pt}} (1.24\%) \\
LiveBench & 2{,}286 & \textcolor{violet}{\rule{3.4pt}{6pt}} (1.00\%) \\
\rowcolor{gray!10} MMLU-Pro leaderboard & 2{,}220 & \textcolor{violet}{\rule{3.3pt}{6pt}} (0.97\%) \\
RewardBench 2 & 1{,}379 & \textcolor{violet}{\rule{2.0pt}{6pt}} (0.60\%) \\
\rowcolor{gray!10} RewardBench & 1{,}025 & \textcolor{violet}{\rule{1.5pt}{6pt}} (0.45\%) \\
ARC Prize evaluations leaderboard & 1{,}020 & \textcolor{violet}{\rule{1.5pt}{6pt}} (0.44\%) \\
\rowcolor{gray!10} Japanese Financial Benchmark & 943 & \textcolor{violet}{\rule{1.4pt}{6pt}} (0.41\%) \\
global-mmlu-lite & 912 & \textcolor{violet}{\rule{1.4pt}{6pt}} (0.40\%) \\
\rowcolor{gray!10} OCRBench v2 & 874 & \textcolor{violet}{\rule{1.3pt}{6pt}} (0.38\%) \\
FinanceMATH & 817 & \textcolor{violet}{\rule{1.2pt}{6pt}} (0.36\%) \\
\rowcolor{gray!10} MM-InstructEval & 783 & \textcolor{violet}{\rule{1.2pt}{6pt}} (0.34\%) \\
CharXiv & 780 & \textcolor{violet}{\rule{1.2pt}{6pt}} (0.34\%) \\
\rowcolor{gray!10} EffiBench-X & 716 & \textcolor{violet}{\rule{1.1pt}{6pt}} (0.31\%) \\
SuperGPQA & 614 & \textcolor{violet}{\rule{0.9pt}{6pt}} (0.27\%) \\
\rowcolor{gray!10} SEED-Bench-2 & 607 & \textcolor{violet}{\rule{0.9pt}{6pt}} (0.26\%) \\
fibble\_arena\_daily & 559 & \textcolor{violet}{\rule{0.8pt}{6pt}} (0.24\%) \\
\rowcolor{gray!10} REST & 528 & \textcolor{violet}{\rule{0.8pt}{6pt}} (0.23\%) \\
\bottomrule
% \vspace{0.1mm}
\end{tabular}
\end{table}

\clearpage

\section{Full Schema Field Reference}
\label{sec:app-schema}
This appendix describes the top-level interface of the unified evaluation schema used in Every Eval Ever. The schema is splitted into two linked records: an aggregate evaluation record for run-level metadata and summary metrics, and a companion instance-level record for per-sample outcomes. In the current release, the canonical interfaces are \texttt{eval.schema.json} (version \texttt{0.2.2}) and \texttt{instance\_level\_eval.schema.json} (version \texttt{instance\_level\_eval\_0.2.2}). Both schemas define closed top-level interfaces: unspecified top-level fields are not permitted.

\subsection{Aggregate Evaluation Records}
\label{subsec:aggregate_evaluation_records}
The aggregate record represents a single evaluation run for one model and stores the provenance, model context, evaluation framework, and one or more reported metric results. It is defined by \texttt{eval.schema.json} version \texttt{0.2.2}. Its top-level fields are summarized in Table~\ref{tab:aggregate_evaluation_record}.

Each element of \texttt{evaluation\_results} is an \texttt{evaluation\_result} object. Its top-level structure is summarized in Table~\ref{tab:evaluation_result_record}.

Taken together, the fields in Tables~\ref{tab:aggregate_evaluation_record} and~\ref{tab:evaluation_result_record} define the aggregate representation of an evaluation run. The separation between \texttt{evaluation\_timestamp} and \texttt{retrieved\_timestamp} distinguishes when the evaluation was executed from when the standardized record was created, while the \texttt{evaluation\_results} array allows multiple benchmark outcomes or metrics to be attached to the same run-level record.

\subsection{Instance-Level Evaluation Records}
\label{subsec:instance_level_evaluation_records}

The instance-level record represents a single benchmark sample associated with an aggregate evaluation run. It is defined by \texttt{instance\_level\_eval.schema.json} version \texttt{instance\_level\_eval\_0.2.2} and is typically stored as a companion JSONL file. Its top-level fields are summarized in Table~\ref{tab:instance_level_evaluation_record}.

The instance-level schema is intentionally aligned with the aggregate schema but preserves sample-level detail needed for auditing and re-analysis. The \texttt{evaluation\_id} field links each row back to the aggregate JSON, while \texttt{evaluation\_result\_id} provides the preferred deterministic link to one specific aggregate metric result. The conditional use of \texttt{output} versus \texttt{messages} makes the schema applicable to standard single-turn tasks as well as conversational and tool-using evaluations.

\section{Converter Implementation Details}
\label{sec:app-converters}
This section describes how evaluation logs from HELM, lm-eval-harness, and Inspect AI are mapped into the unified \eeeshort{} schema. All three converters produce an aggregate \texttt{EvaluationLog}. When instance-level data is available, they also emit a JSONL file referenced by \texttt{detailed\_evaluation\_results}. In addition to these core framework converters, the repository includes community-contributed converters for public leaderboards and benchmark-specific result sources, summarized in a final subsection.
\subsection{Inspect AI Converter}
\label{sec:app-conv-inspect}

\paragraph{Input format.}
The Inspect converter accepts evaluation logs with extension \texttt{.eval} or \texttt{.json}. Both formats represent the same logical object and are read through the Inspect log API\footnote{\url{https://inspect.aisi.org.uk/reference/inspectAI.log.html}}. The top-level object is structurally rich: \texttt{eval} stores task, dataset, model path, package versions, generation settings, and task arguments; \texttt{plan} stores solver steps and plan configuration; \texttt{results} stores scorer-level aggregate metrics; \texttt{stats} stores run timestamps and summary counters; \texttt{samples} stores per-sample traces; and \texttt{reductions} stores scorer-level reduced sample values used for score resolution. This structure motivates field-wise extraction rather than direct key renaming.

\begin{table*}[h]
\caption{Top-level fields of the aggregate evaluation record.}
\label{tab:aggregate_evaluation_record}
\centering
\scriptsize
\setlength{\tabcolsep}{3pt}
\renewcommand{\arraystretch}{1.12}
\begin{tabular}{p{0.95in} p{0.60in} p{0.45in} p{1.35in} p{1.45in}}
\toprule
\textbf{Field} & \textbf{Type} & \textbf{Required} & \textbf{Description} & \textbf{Notes} \\
\midrule
\texttt{schema\_version} & \texttt{string} & Yes & Version of the schema used for the aggregate evaluation record. & For this appendix, the current released value is \texttt{0.2.2}. \\
\texttt{evaluation\_id} & \texttt{string} & Yes & Unique identifier for a specific evaluation run. & This field identifies the run as a whole rather than an individual metric result. The schema documentation recommends the form \texttt{eval\_name/}\allowbreak\texttt{model\_id/}\allowbreak\texttt{retrieved\_timestamp}. \\
\texttt{evaluation\_timestamp} & \texttt{string} & No & Timestamp indicating when the evaluation was run. & This is run time, not record-creation time. \\
\texttt{retrieved\_timestamp} & \texttt{string} & Yes & Timestamp indicating when the aggregate record was created. & This is record-creation time in Unix epoch format and is required. \\
\texttt{source\_metadata} & \texttt{object} & Yes & Metadata describing where the evaluation evidence comes from. & This nested object captures provenance such as whether the result comes from documentation or a direct evaluation run, the providing organization, and the evaluator-model relationship. \\
\texttt{eval\_library} & \texttt{object} & Yes & Metadata describing the evaluation framework used to obtain the result. & This nested object records the framework name and version, with optional additional details. \\
\texttt{model\_info} & \texttt{object} & Yes & Canonical description of the evaluated model. & This nested object captures model identity and inference context, including model identifier, developer, and optional inference platform or local inference engine. \\
\texttt{evaluation\_results} & \texttt{array} & Yes & Collection of metric results reported for the evaluation run. & This is the main result-bearing field. Each element stores one metric outcome together with its dataset source, metric configuration, score details, and optional generation configuration. A single run may report multiple metric results. \\
\texttt{detailed\_evaluation\_results} & \texttt{object} & No & Reference to a companion file containing per-sample outcomes. & When present, this object points to the instance-level JSONL file and records linkage metadata such as format, path, checksum, hash algorithm, and row count. \\
\bottomrule
\end{tabular}
\end{table*}

\paragraph{Aggregate mapping to \eeeshort{}}
Core \eeeshort{} fields are assembled from several Inspect structures. The converter derives \texttt{evaluation\_timestamp} from \texttt{stats.started\_at} with fallback to \texttt{eval.created}, derives \texttt{eval\_library} from \texttt{eval.packages}, and derives \texttt{evaluation\_results} from \texttt{results.scores}. It initializes \texttt{model\_info} from \texttt{eval.model} and, when available, refines that identifier with sample-level output model metadata.

Inspect model paths vary across providers, so the converter applies provider-specific normalization to produce a canonical \texttt{model\_info.id} and to infer \texttt{inference\_platform} and, when possible, \texttt{inference\_engine}. For example, \texttt{openai/azure/gpt-4o-mini} may be refined to \texttt{openai/gpt-4o-mini-2024-07-18}, while \texttt{ollama/qwen2.5:0.5b} is normalized to \texttt{ollama/qwen2.5-0.5b}. The converter derives per-result source data mainly from \texttt{eval.dataset}. In particular, \texttt{eval.dataset.location} is treated as a Hugging Face repository only when it matches the canonical \texttt{namespace/name} form; otherwise the converter preserves the raw Inspect dataset fields in \texttt{additional\_details}.

Each scorer metric is converted into an \texttt{EvaluationResult}, except standalone \texttt{stderr}, which is treated as uncertainty metadata rather than a separate metric. Metric values populate \texttt{score\_details.score}, and uncertainty is populated from scorer-reported \texttt{stderr} together with optional \texttt{std} or \texttt{stddev} and sample counts. When scorer parameters expose \texttt{grader\_model} and \texttt{grader\_template}, the converter also populates \texttt{metric\_config.llm\_scoring} to preserve judge context. Finally, \texttt{generation\_config} combines standard generation parameters with Inspect-specific execution context, including prompt template, available tools, serialized plan, limits, sandbox configuration, retry settings, and the reasoning flag.

\paragraph{Instance-level mapping}
When sample logs are present, each sample is converted into one \texttt{InstanceLevelEvaluationLog}. Interaction type is inferred from the message structure: tool-role messages yield \texttt{agentic}, multiple assistant turns without tools yield \texttt{multi\_turn}, and the remaining cases are treated as \texttt{single\_turn}. Input text is serialized from user messages, with references and choices preserved from sample targets and choices. For single-turn cases, output text and reasoning traces are stored in \texttt{output}; for multi-turn and agentic cases, the normalized message sequence is stored in \texttt{messages}, including tool calls.

Score resolution prioritizes \texttt{reductions} matched by sample and scorer, then sample-level reduced values, and finally direct sample scores. If no score is available, the converter falls back to reference matching, and correctness follows the resolved score semantics. The instance-level record also preserves token usage, latency and generation time, sample hash, stop reason, epoch, answer-attribution metadata, and full error traces when available.

\subsection{HELM Converter}
\label{sec:app-conv-helm}
\paragraph{Input format.}
The HELM converter operates on a HELM run directory. It requires \texttt{run\_spec.json}, \texttt{scenario\_state.json}, \texttt{scenario.json}, and \texttt{per\_instance\_stats.json}, while the optional \texttt{stats.json} file supplies aggregate metric values when present. Each file contributes a distinct part of the run state. \texttt{run\_spec.json} provides adapter, metric, and scenario specification metadata; \texttt{scenario\_state.json} provides request-level records, including prompts, references, outputs, and request timestamps; \texttt{scenario.json} provides scenario naming metadata used during dataset identification; and \texttt{per\_instance\_stats.json} provides per-sample metric and token statistics for instance-level conversion. When available, \texttt{stats.json} adds aggregate statistics such as mean, sum, count, and standard deviation. Because HELM distributes these signals across separate artifacts, the converter must coordinate extraction across multiple files rather than perform direct key renaming.

\paragraph{Aggregate mapping to \eeeshort{}}
Core \eeeshort{} fields are assembled from multiple HELM artifacts. The converter derives \texttt{model\_info} from \texttt{run\_spec.json}, preferably through deployment registry entries referenced by \texttt{adapter\_spec.model\_deployment}; if deployment lookup is unavailable, it falls back to adapter-level model fields and best-effort platform inference. It derives \texttt{evaluation\_results} primarily from \texttt{stats.json}, with candidate metric names anchored in \texttt{run\_spec.metric\_specs}.

The converter derives per-result source data and timestamp fields from \texttt{scenario\_state.json} and \texttt{scenario.json}. Dataset name comes from \texttt{scenario.name} when available, with fallback parsing from \texttt{run\_spec.name}; sample counts and sample identifiers come from request-state instance ids; and scenario class names and arguments are preserved in \texttt{additional\_details}. For each matched aggregate statistic, the score is taken from \texttt{mean} with fallback to \texttt{sum/count}, while uncertainty records \texttt{stddev} and sample-count metadata. If \texttt{stats.json} is absent, aggregate metric output may be empty.

\texttt{generation\_config} is derived from request-level and adapter-level settings, including \texttt{temperature}, \texttt{top\_p}, \texttt{top\_k}, \texttt{max\_tokens}, stop sequences, penalties, completion count, and a reasoning flag inferred from HELM thinking traces. The converter computes \texttt{evaluation\_timestamp} from the earliest available request datetime, with fallback to retrieval time, and forms \texttt{evaluation\_id} from dataset, model, and timestamp after path-safe normalization of model identifier separators.

\paragraph{Instance-level mapping}
When request-state records are available, the converter emits one consolidated JSONL file and references it through \texttt{detailed\_evaluation\_results}. Each row combines \texttt{request\_states} with \texttt{per\_instance\_stats}: prompt text, references, and choices come from request-state records and output mapping metadata, while completions and optional reasoning traces come from model outputs. Score resolution prefers per-instance \texttt{exact\_match} statistics and otherwise falls back to reference matching between generated completions and tagged correct references. The converter also records token usage, generation latency, stable sample identifiers, sample hash, answer-attribution metadata, and single-turn interaction typing.

\subsection{lm-eval-harness Converter}
\label{sec:app-conv-lmeval}
\paragraph{Input format.}
The lm-eval converter consumes result files named \texttt{results\_*.json}. Optional instance-level conversion uses files named \texttt{samples\_<task>\_*.jsonl} and is enabled only when sample logging is available and conversion is executed with \texttt{--include\_samples}. The aggregate record combines several top-level maps: \texttt{config} provides global run and model metadata, \texttt{configs} provides task-level dataset and generation settings, \texttt{results} provides task-level metric values, \texttt{higher\_is\_better} provides metric directionality, \texttt{n-samples} provides sample-count metadata for uncertainty reporting, \texttt{date} maps to \texttt{evaluation\_timestamp}, and \texttt{lm\_eval\_version} maps to \texttt{eval\_library.version}. This layout motivates task-wise field extraction rather than direct key renaming.

\begin{table*}[h]
\caption{Top-level fields of a single \texttt{evaluation\_result} entry within \texttt{evaluation\_results}.}
\label{tab:evaluation_result_record}
\centering
\scriptsize
\setlength{\tabcolsep}{3pt}
\renewcommand{\arraystretch}{1.12}
\begin{tabular}{p{0.95in} p{0.60in} p{0.45in} p{1.35in} p{1.45in}}
\toprule
\textbf{Field} & \textbf{Type} & \textbf{Required} & \textbf{Description} & \textbf{Notes} \\
\midrule
\texttt{evaluation\_result\_id} & \texttt{string} & No & Stable identifier for a specific metric result within the evaluation run. & This is the preferred deterministic join key for linking instance-level rows to one aggregate metric result. \\
\texttt{evaluation\_name} & \texttt{string} & Yes & Name of the evaluation associated with the metric result. & This identifies the benchmark or task reported by the metric entry. \\
\texttt{source\_data} & \texttt{object} & Yes & Description of the dataset source used by the metric result. & This object is a tagged union with three variants: URL-based source data, Hugging Face dataset source data, and \texttt{other} for private or custom datasets. \\
\texttt{evaluation\_timestamp} & \texttt{string} & No & Timestamp indicating when this specific metric result was produced. & This field is useful when different results within the same run were generated at different times. \\
\texttt{metric\_config} & \texttt{object} & Yes & Metadata defining how the metric should be interpreted. & This object captures metric semantics such as whether lower values are better, the score type, and optional normalized metric identifiers or parameters. \\
\texttt{score\_details} & \texttt{object} & Yes & Reported score and associated quantitative details. & This object stores the metric value itself and may additionally include uncertainty estimates or supplementary score details. \\
\texttt{generation\_config} & \texttt{object} & No & Configuration describing how model outputs were generated for this metric result. & This object can record generation arguments and additional run details, including settings relevant to agentic evaluations. \\
\bottomrule
\end{tabular}
\end{table*}

\paragraph{Aggregate mapping to \eeeshort{}}
The \texttt{results} object may contain both metric-bearing tasks and structural placeholders, so the converter first excludes placeholders and tasks without numeric metrics and then emits one \texttt{EvaluationLog} per retained task. For each retained task, it derives \texttt{evaluation\_timestamp} from \texttt{date}, derives \texttt{eval\_library.version} from \texttt{lm\_eval\_version}, derives \texttt{model\_info} from \texttt{config}, and derives both per-result source data and \texttt{generation\_config} from task-specific entries in \texttt{configs}.

\texttt{model\_info} is constructed from \texttt{config.model} and parsed \texttt{config.model\_args}. Because \texttt{model\_args} is often a comma-delimited string, the converter parses it heuristically and prioritizes \texttt{pretrained} when present; inference platform and inference engine are then inferred from model-type mappings, with an optional command-line override for engine name and version. Per-result source data is derived from task configuration fields such as \texttt{dataset\_path} and split metadata. Paths that match Hugging Face repository form are mapped to \texttt{source\_type=hf\_dataset}, while other paths are mapped to \texttt{source\_type=other}.

Metric keys typically follow the \texttt{metric,filter} convention, such as \texttt{exact\_match,none}, and uncertainty keys follow the \texttt{metric\_stderr,filter} convention. The converter decomposes these keys, creates one \texttt{EvaluationResult} per numeric metric, and maps standard error to \texttt{score\_details.uncertainty.standard\_error}. Metric directionality is derived from \texttt{higher\_is\_better} and inverted into \texttt{lower\_is\_better}; score bounds are inferred from a known-metrics table when available and left unset otherwise. Finally, \texttt{generation\_config} is derived from \texttt{generation\_kwargs}, including \texttt{temperature}, \texttt{top\_p}, \texttt{top\_k}, and \texttt{max\_gen\_toks}, while the remaining generation attributes and \texttt{num\_fewshot} are preserved in \texttt{additional\_details}.

\paragraph{Instance-level mapping}
When sample JSONL files are available, each sample row is converted into one \texttt{InstanceLevelEvaluationLog}. Prompt and references are extracted from \texttt{arguments} and \texttt{target}, and for multiple-choice tasks the answer options are reconstructed from \texttt{gen\_args\_*}. For generation tasks, the converter uses the first response text; for multiple-choice tasks, it selects the option with the highest log probability from \texttt{filtered\_resps} or \texttt{resps}. Scores and correctness are derived from per-sample metric fields, with fallback to \texttt{score=0.0} and \texttt{is\_correct=false} when no numeric metric value is available. The converter also records a sample hash, lm-eval hashes \texttt{doc\_hash}, \texttt{prompt\_hash}, and \texttt{target\_hash}, the applied filter name, and the serialized per-sample metric payload.

\begin{table*}[h]
\caption{Top-level fields of the instance-level evaluation record.}
\label{tab:instance_level_evaluation_record}
\centering
\scriptsize
\setlength{\tabcolsep}{3pt}
\renewcommand{\arraystretch}{1.12}
\begin{tabular}{p{0.95in} p{0.60in} p{0.45in} p{1.35in} p{1.45in}}
\toprule
\textbf{Field} & \textbf{Type} & \textbf{Required} & \textbf{Description} & \textbf{Notes} \\
\midrule
\texttt{schema\_version} & \texttt{string} & Yes & Version of the schema used for the instance-level record. & For this appendix, the current released value is \texttt{instance\_level\_eval\_0.2.2}. \\
\texttt{evaluation\_id} & \texttt{string} & Yes & Foreign key linking the instance-level record to the aggregate evaluation JSON. & This value must match the aggregate record and anchors the per-sample row to a specific evaluation run. \\
\texttt{model\_id} & \texttt{string} & Yes & Identifier of the evaluated model. & This is the model identifier for the sample-level record and supports joins or filtering even when the aggregate file is not loaded. It should use Hugging Face-style formatting, i.e., \texttt{model\_developer/model\_name}. \\
\texttt{evaluation\_name} & \texttt{string} & Yes & Name of the evaluation associated with the sample. & This is primarily a display and filtering field when no deterministic metric-level join key is available. \\
\texttt{evaluation\_result\_id} & \texttt{string} & No & Preferred foreign key to a specific aggregate metric result. & This is the preferred one-to-one link from an instance-level row to a particular element of \texttt{evaluation\_results}. If one underlying sample contributes to multiple aggregate metrics, separate instance-level records should be emitted. \\
\texttt{sample\_id} & \texttt{string} & Yes & Identifier of the source benchmark sample. & This is typically inherited from the original dataset. \\
\texttt{sample\_hash} & \texttt{string | null} & No & Hash-based identifier for the sample content. & This supports cross-model matching when \texttt{sample\_id} is unstable or inconsistent across sources. Operationally, it is computed from the concatenation of \texttt{input.raw} and \texttt{input.reference}, yielding a content-based identifier for the sample. \\
\texttt{interaction\_type} & \texttt{string} & Yes & Interaction regime for the sample. & Allowed values are \texttt{single\_turn}, \texttt{multi\_turn}, and \texttt{agentic}. This field controls which output container is valid. \\
\texttt{input} & \texttt{object} & Yes & Input content presented to the model for the sample. & This nested object stores the raw prompt, reference answers, and optional formatted input or answer choices. \\
\texttt{output} & \texttt{object | null} & No & Model output for single-turn evaluations. & \texttt{single\_turn} requires \texttt{output} and forbids \texttt{messages}. \\
\texttt{messages} & \texttt{array | null} & No & Message transcript for multi-turn or agentic evaluations. & \texttt{multi\_turn} and \texttt{agentic} require \texttt{messages} and forbid \texttt{output}. The message list can include tool calls and tool outputs. \\
\texttt{answer\_attribution} & \texttt{array} & Yes & Record of how the scored answer was extracted from the model behavior. & This array identifies the source location, extraction method, extracted value, and whether the extracted answer is terminal. \\
\texttt{evaluation} & \texttt{object} & Yes & Instance-level scoring outcome for the sample. & This nested object stores the sample score and correctness, and may additionally report turn counts or tool-call counts. \\
\texttt{token\_usage} & \texttt{object | null} & No & Token accounting for the sample. & This nested object records input, output, and total token counts, with optional cache and reasoning-token fields. \\
\texttt{performance} & \texttt{object | null} & No & Latency and runtime measurements for the sample. & This nested object supports per-sample performance analysis, including latency and time-to-first-token metrics. \\
\texttt{error} & \texttt{string | null} & No & Error information associated with the sample. & This field can capture failures such as refusals, timeouts, or API errors. \\
\texttt{metadata} & \texttt{object | null} & No & Optional sample-level metadata. & This object can store benchmark-specific annotations such as subject, difficulty, or tags without changing the top-level interface. \\
\bottomrule
\end{tabular}
\end{table*}

\subsection{Community-Contributed Converters}
\label{sec:app-comm}

\paragraph{AlpacaEval.}
The AlpacaEval converter fetches the public AlpacaEval 1.0 and 2.0 leaderboard CSVs. It preserves pairwise preference metrics against the published baselines, including win rate, length-controlled win rate, discrete win rate, and average response length for each model.

\paragraph{ARC-AGI.}
The \texttt{arc\_agi} adapter reads the ARC Prize evaluations leaderboard JSON from \texttt{arcprize.org}. It records the published ARC score together with cost-per-task and total-cost fields while normalizing the often informal model aliases used on the leaderboard.

\paragraph{Artificial Analysis.}
The \texttt{artificial\_analysis} adapter ingests the Artificial Analysis LLM API, which combines benchmark scores with pricing and latency measurements for frontier models. It carries over composite indices such as the Artificial Analysis intelligence, coding, and math indices, benchmark scores such as MMLU-Pro, GPQA, HLE, LiveCodeBench, SciCode, AIME, and tau2, and token-pricing and latency summaries.

\paragraph{BFCL.}
The \texttt{bfcl} adapter reads the BFCL leaderboard CSV published by Berkeley Gorilla. It preserves the leaderboard's overall rank, overall accuracy, latency and cost fields, and the benchmark's finer-grained tool-calling slices, including non-live, live, multi-turn, and web-search accuracies.

\paragraph{CocoaBench.}
The \texttt{cocoabench} adapter reads CocoaBench's published per-system CSV of agent performance, time, and cost. It preserves overall benchmark accuracy together with average runtime per task, average cost per task, and total evaluation cost for each released agent-model system.

\paragraph{Exgentic.}
The \texttt{exgentic} adapter consumes Exgentic open-agent leaderboard aggregates, either from local \texttt{results.json} files or the Hugging Face dataset. These runs span agent benchmarks such as AppWorld, SWE-bench, BrowseComp+, and Tau2, and the adapter preserves benchmark score, session counts, and run-cost summaries for each agent-model submission.

\paragraph{Global MMLU Lite.}
The \texttt{global-mmlu-lite} adapter fetches the Global MMLU Lite leaderboard from the Kaggle Benchmarks API. It preserves the reported Global MMLU Lite score for each model together with any confidence-interval or standard-deviation information exposed by the leaderboard payload.

\paragraph{Open LLM Leaderboard v2.}
The \texttt{hfopenllm\_v2} adapter ingests the Hugging Face Open LLM Leaderboard v2 API. It preserves the benchmark panel used by that leaderboard, including IFEval, BBH, MATH Level 5, GPQA, MUSR, and MMLU-Pro, together with basic model metadata such as architecture, precision, and parameter count when available.

\paragraph{LLM Stats.}
The \texttt{llm\_stats} adapter consumes the LLM Stats API's combined model, benchmark, and score payloads. It is designed for a broad benchmark catalog rather than a single leaderboard, so it preserves benchmark-specific provenance URLs, relationship metadata, pricing and context-window model details, and the score entries attached to each model.

\paragraph{Multi-SWE-Bench.}
The \texttt{multi\_swe\_bench} adapter clones the Multi-SWE-Bench experiments repository and reads verified submissions under each language-specific leaderboard. It preserves resolved-instance rates and submission metadata for C, C++, Go, Java, JavaScript, Rust, and TypeScript tracks.

\paragraph{RewardBench.}
The \texttt{rewardbench} adapter fetches RewardBench v1 leaderboard CSV data and RewardBench v2 JSON results from Hugging Face. It preserves the v1 overall, chat, chat-hard, safety, reasoning, and prior-set scores, as well as the v2 factuality, precise instruction following, math, safety, focus, and tie-handling metrics.

\paragraph{SciArena.}
The \texttt{sciarena} adapter reads the SciArena leaderboard API maintained by Allen AI. It preserves the published rank, arena rating, and cost-per-100-calls metadata for each model, while keeping the source model aliases close to the leaderboard's own naming.

\paragraph{SWE-bench Verified.}
The \texttt{swe\_bench\_verified} adapter reads verified submission directories from the public SWE-bench experiments repository. It preserves the standard verified leaderboard signal, namely the fraction of the 500 benchmark instances resolved by each submission, along with submission metadata and agent tooling context.

\paragraph{SWE-PolyBench.}
The \texttt{swe\_polybench} adapter reads submission artifacts for SWE-PolyBench and SWE-PolyBench Verified from the public experiments repository. It preserves resolved-instance rates separately for each dataset variant and programming language, so one submission may yield distinct records for different language tracks.

\paragraph{Terminal-Bench 2.0.}
The \texttt{terminal\_bench\_2} adapter captures the published Terminal-Bench 2.0 leaderboard for agentic coding systems. It preserves the leaderboard's accuracy and standard-error values for each agent-model pair on the 87-task benchmark, together with the agent and model organization metadata shown on the leaderboard.

\FloatBarrier

\section{Conservative Estimation of Costs}\label{app:cost}
We explain here our assumptions on how we estimate the cost for running evaluations to reproduce all of our data. While we note that this is a vast underapproximation of the actual cost of reproduction all this work, we still see it as a sign for the importance of collecting such data.

\subsection{Dataset and Evaluation Scale}
The dataset comprises approximately 230{,}000 model--benchmark evaluation pairs, where each evaluation represents running a model on a single benchmark.

Each benchmark is assumed to contain 1{,}000 examples, with roughly 100 input tokens and 20 output tokens per example.

Under these assumptions, each evaluation uses about 100{,}000 input tokens and 20{,}000 output tokens, for a total of 120{,}000 tokens before additional overhead. 

\subsection{LLM-as-Judge Overhead}

Modern evaluation pipelines frequently incorporate an additional language model to automatically grade or compare outputs, commonly referred to as an ``LLM-as-judge.'' Based on production observations, this introduces an additional 60\% token overhead relative to the base evaluation.

This overhead is modeled as a multiplicative factor applied uniformly to both input and output tokens, such that the adjusted token count is given by $1.6$ times the base tokens. Consequently, each evaluation involves approximately 160{,}000 input tokens and 32{,}000 output tokens after accounting for this overhead.

\subsection{Total Token Volume}

Aggregating across all 230{,}000 evaluations, the total token volume is obtained by multiplying the per-evaluation total of 192{,}000 tokens by the number of evaluations. This results in approximately $4.416 \times 10^{10}$ tokens, corresponding to roughly 44 billion tokens processed in total.

\subsection{Cost Model}

The total inference cost is computed as the sum of input and output token costs. Specifically, the cost is given by the product of input tokens and their per-million-token price, plus the product of output tokens and their corresponding price. The pricing parameters are denoted by $C_{\text{in}}$ for input tokens and $C_{\text{out}}$ for output tokens.

We consider three levels of approximation corresponding to different pricing regimes.

\subsection{Low-Cost Estimate (No Judge)}

As a lower bound, we consider a highly cost-efficient model with pricing of \$0.10 per million input tokens and \$0.40 per million output tokens. This estimate excludes any judge overhead and therefore uses the base token counts.

Under these assumptions, the input cost per evaluation is computed as $100{,}000 \times \frac{0.10}{10^6}$, which equals \$0.01. The output cost per evaluation is $20{,}000 \times \frac{0.40}{10^6}$, which equals \$0.008. The total cost per evaluation is therefore \$0.018.

Across all 230{,}000 evaluations, the total cost is approximately $230{,}000 \times 0.018$, which yields about \$4{,}140. This corresponds to a total low-cost estimate of approximately \$4.1K.

\subsection{Mid-Cost Estimate (Sonnet with Judge)}

For a more realistic estimate, we consider a mid-tier model with pricing of \$3 per million input tokens and \$15 per million output tokens. This estimate incorporates the 60\% judge overhead.

With adjusted token counts, the input cost per evaluation is $160{,}000 \times \frac{3}{10^6}$, which equals \$0.48, and the output cost is $32{,}000 \times \frac{15}{10^6}$, which also equals \$0.48. The total cost per evaluation is therefore \$0.96.

Across all evaluations, the total cost is approximately $230{,}000 \times 0.96$, which yields about \$220{,}800. This corresponds to a total mid-cost estimate of approximately \$221K.

\subsection{High-Cost Estimate (Opus with Judge)}

Finally, we consider a higher-end model with pricing of \$5 per million input tokens and \$25 per million output tokens, again including the 60\% judge overhead.

Under these conditions, the input cost per evaluation is $160{,}000 \times \frac{5}{10^6}$, which equals \$0.80, and the output cost is $32{,}000 \times \frac{25}{10^6}$, which also equals \$0.80. The total cost per evaluation is therefore \$1.60.

Across all evaluations, the total cost is approximately $230{,}000 \times 1.60$, resulting in about \$368{,}000. This corresponds to a total high-cost estimate of approximately \$368K.

\subsection{Summary} Under the stated assumptions, the total cost of evaluating 230{,}000 model--benchmark pairs ranges from a lower bound of approximately \$4K, assuming no judge and highly optimized pricing, to approximately \$370K when using a high-end model with judge overhead. A mid-tier estimate of roughly \$220K is also given. While the lower bound is likely unrealistic, the others might be closer to actual pricing as most models are not of the smaller kinds and usually top or middle models are evaluated, with or without a judge.

\FloatBarrier
\section{Governance Card}\label{app:governance}
\eee{} is a community project. This appendix documents the governance mechanisms currently in place. We follow the spirit of the Croissant governance process \citep{croissant2024} and adapt it to the specifics of \eee{}. The governance model remains dynamic, and we expect it to evolve as the project progresses.

\subsection{Decision-Making and Roles}\label{app:governance-decision}

The project recognizes three key roles. \emph{Core maintainers} are responsible for repository upkeep, schema releases, converter maintenance, reviewing contributions, and final decisions on contested proposals. \emph{Contributors} submit data, converters, schema proposals, tooling, or documentation through pull requests and issues or discussions through GitHub or, on occasion, Slack. \emph{Community reviewers} are volunteer experts who participate in schema discussions and review proposals in their area of expertise. Roles are not exclusive: maintainers also contribute and become so through community acceptance and after several contributions. 

Routine decisions (record additions that pass validation, bug fixes, documentation updates, and additive non-breaking schema fields) are made by maintainers on a rolling basis. Substantive decisions (breaking schema changes, new interaction types, deprecations, deduplication policy changes) follow the proposal process below.

\subsection{Schema Change Proposal Process}\label{app:governance-proposal}

Substantive schema changes follow a lightweight three-stage process modeled on the iterative methodology used to produce the \textit{ vx.x.x }schema (Section~\ref{sec:schema}).

\begin{enumerate}
    \item \textbf{Proposal.} A contributor opens an issue in the repository describing the proposed change, or is raised during discussion between maintainers. The problem it solves and the implications are discussed, and alternatives are weighed.
    \item \textbf{Community review.} The proposal is open for discussion until disagreements are resolved. If necessary, maintainers solicit feedback from relevant community experts based on the area of the proposal.
    \item \textbf{Resolution.} Maintainers summarize the discussion and propose a resolution: accept, accept with modification, defer, or decline. Decisions are made by consensus among maintainers; when consensus cannot be reached, a documented majority decision is recorded, with dissenting positions preserved in the schema change-log (Section~\ref{sec:schema}).
\end{enumerate}

\subsection{Conflicting Submissions and Duplicate Records}\label{app:governance-conflicts}

Because the schema assigns a unique UUID to each evaluation run and defers deduplication to the analysis layer (Section~\ref{sec:design-process}), conflicting or near-duplicate records are expected and, by themselves, are not a governance problem. The validator flags likely duplicates (same model, same benchmark, same metric, same evaluator) at submission time but does not reject them. When users encounter conflicting records that cannot be reconciled from metadata alone, they are encouraged to open an issue; \textit{maintainers} may then request additional metadata from the \textit{contributors}, annotate records with a disputed flag in \emph{additional\_details}, or, in cases of clear error, mark records as superseded (see Section~\ref{app:governance-corrections} below). We do not arbitrate which of two methodologically valid evaluation runs is "correct."

\subsection{Corrections, Retractions, and Supersession}\label{app:governance-corrections}

While there have not yet been any disputes over data contributions, we present here a proposal for how to address them when they do arise. We will continue to adapt this process in response to emerging real-world needs.
Records are immutable once accepted: modifying a record in place would invalidate downstream analyses that reference it. Three mechanisms handle errors and updates:

\begin{enumerate}
    \item \textbf{Correction.} For minor fixes (e.g., typos in identifiers), a new record is added that supersedes the original. The original is retained and annotated with a superseded\_by field pointing to the corrected UUID. Similarly, a preceded\_by field points to the original UIUD.
    \item \textbf{Retraction.} For records that were submitted in error or are based on faulty source data, the record is annotated with a retracted flag and a brief reason. The record itself is not deleted, so prior analyses remain reproducible.
    \item \textbf{Schema migration.} When schema versions advance, records remain valid under the version they were submitted with. Migration utilities are provided where possible, but historical records are not overwritten.
\end{enumerate}

\subsection{Code of Conduct and Acknowledgment}\label{app:governance-code}

The project follows a standard contributor code of conduct. \textit{Contributors} are acknowledged in three ways: through git commit history, through the contributor list maintained in the repository, and, for substantive contributions to a release, through co-authorship on the associated release paper. The first such instance is the present submission, organized as a shared task \citep{batzner2026shared}; subsequent releases will follow the same pattern with criteria documented in the contributor guide.

\subsubsection{Copy of the Contributor Guide}
\label{sec:contribGuide}
New data can be contributed to the Hugging Face Datastore using the following process:

Leaderboard/evaluation data is split-up into files by individual model, and data for each model is stored using \texttt{eval.schema.json}. The repository is structured into folders as \texttt{data/{benchmark_name}/{developer_name}/{model_name}/}.

\textbf{TL;DR How to successfully submit}
\begin{enumerate}
    \item Data must conform to \texttt{eval.schema.json} (current version: \texttt{0.2.2})
    \item The validation pipeline will automatically verify the data submitted in the pull request, but can also be manually triggered by typing \texttt{/eee validate changed} in a comment on the HF PR.
    \item A core maintainer will review and merge your submission
\end{enumerate}

\textbf{PR Naming Convention}

Use these prefixes in your pull request titles:
\begin{itemize}
    \item \texttt{[Submission]}  - New evaluation data
    \item \texttt{[Issue \#N]} - Fix for a specific GitHub issue
    \item \texttt{[Feature]} - New functionality not tied to an issue
    \item \texttt{[Docs]} - Documentation changes
\end{itemize}

\textbf{UUID Naming Convention}

Each JSON file is named with a \textbf{UUID (Universally Unique Identifier)} in the format \texttt{{uuid}.json}. The UUID is automatically generated (using standard UUID v4) when creating a new evaluation result file. This ensures that:
\begin{itemize}
    \item \textbf{Multiple evaluations} of the same model can exist without conflicts (each gets a unique UUID)
    \item \textbf{Different timestamps} are stored as separate files with different UUIDs (not as separate folders)
    \item A model may have multiple result files, with each file representing different iterations or runs of the leaderboard/evaluation
    \item UUID's can be generated using Python's \texttt{uuid.uuid4()} function.
\end{itemize}

\textbf{Example}: The model \texttt{openai/gpt-4o-2024-11-20} might have multiple files like:
\begin{itemize}
    \item \texttt{e70acf51-30ef-4c20-b7cc-51704d114d70.json} (evaluation run \#1)
    \item \texttt{a1b2c3d4-5678-90ab-cdef-1234567890ab.json} (evaluation run \#2)
\end{itemize}

Note: Each file can contain multiple individual results related to one model.

\textbf{How to add new eval:}
\begin{enumerate}
    \item Add a new folder under \texttt{data/} on the Hugging Face datastore with a codename for your eval.
    \item For each model, use the Hugging Face (\texttt{developer_name/model_name}) naming convention to create a 2-tier folder structure.
    \item Add a JSON file with results for each model and name it \texttt{{uuid}.json}.
    \item [Optional] Include a \texttt{utils/} folder in your benchmark name folder with any scripts used to generate the data (e.g., \texttt{utils/global-mmlu-lite/adapter.py}).
    \item [Submit] Two ways to submit your evaluation data:
    \begin{itemize}
        \item \textbf{Option A: Drag \& drop via Hugging Face} — Go to the datastore $\rightarrow$ click ``Files and versions'' $\rightarrow$ ``Contribute'' $\rightarrow$
        ``Upload files'' $\rightarrow$ drag and drop your data $\rightarrow$ select ``Open as a pull request to the main branch''.
        \item \textbf{Option B: Clone \& PR} — Clone the repo, add your data under \texttt{data}, and open a pull request
    \end{itemize}
\end{enumerate}

\textbf{Schema Instructions}
\begin{enumerate}
    \item \textbf{\texttt{model\_info}}: Use Hugging Face formatting (\texttt{developer_name/model_name}). If a model does not come from Hugging Face, use the exact API reference. Check examples in \texttt{data/livecodebenchpro}. Notably, some do have a \textbf{date included in the model name}, but others \textbf{do not}. For example:
    \begin{itemize}
        \item OpenAI: \texttt{gpt-4o-2024-11-20}, \texttt{gpt-5-2025-08-07}, \texttt{o3-2025-04-16}
        \item Anthropic: \texttt{claude-3-7-sonnet-20250219}, \texttt{claude-3-sonnet-20240229}
        \item Google: \texttt{gemini-2.5-pro}, \texttt{gemini-2.5-flash}
        \item xAI (Grok): \texttt{grok-2-2024-08-13}, \texttt{grok-3-2025-01-15}
    \end{itemize}
    \item \textbf{\texttt{evaluation_id}}: Use \texttt{{benchmark_name/model_id/retrieved_timestamp}} format (e.g. \texttt{livecodebenchpro/qwen3-235b-a22b-thinking-2507/1760492095.8105888}).
    \item  \textbf{\texttt{inference_platform}} vs \textbf{\texttt{inference_engine}}: Where possible specify where the evaluation was run using one of these two fields.
    \begin{itemize}
        \item \texttt{inference_platform}: Use this field when the evaluation was run through a remote API (e.g., \texttt{openai}, \texttt{huggingface}, \texttt{openrouter}, \texttt{anthropic}, \texttt{xai}).
        \item \texttt{inference_engine}: Use this field when the evaluation was run locally. This is now an object with \texttt{name} and \texttt{version} (e.g. \texttt{{"name": "vllm", "version": "0.6.0"}}).
    \end{itemize}
    \item The \texttt{source_type} on \texttt{source_metadata} has two options: \texttt{documentation} and \texttt{evaluation_run}. Use \texttt{documentation} when results are scraped from a leaderboard or paper. Use \texttt{evaluation_run} when the evaluation was run locally (e.g. via an eval converter).
    \item \textbf{\texttt{source_data}} is specified per evaluation result (inside \texttt{evaluation_results}), with three variants:
    \begin{itemize}
        \item \texttt{source_type: "url"} - link to a web source (e.g. leaderboard API)
        \item \texttt{source_type: "hf_dataset"} — reference to a Hugging Face dataset (e.g. \texttt{{"hf_repo": "google/IFEval"}})
        \item \texttt{source_type: "other"} — for private or proprietary datasets
    \end{itemize}
    \item The schema is designed to accommodate both numeric and level-based (e.g. Low, Medium, High) metrics. For level-based metrics, the actual 'value' should be converted to an integer (e.g. Low = 1, Medium = 2, High = 3), and the \texttt{level_names} property should be used to specify the mapping of levels to integers.
    \item \textbf{Timestamps}: The schema has three timestamp fields — use them as follows:
    \begin{itemize}
        \item \texttt{retrieved_timestamp} (required) — when this record was created, in Unix epoch format (e.g. \texttt{1760492095.8105888})
        \item \texttt{evaluation_timestamp} (top-level, optional) — when the evaluation was run
        \item \texttt{evaluation_results[].evaluation_timestamp} (per-result, optional) — when a specific evaluation result was produced, if different results were run at different times
    \end{itemize}
    \item Additional details can be provided in several places in the schema. They are not required, but can be useful for detailed analysis.
    \begin{itemize}
        \item \texttt{model_info.additional_details}: Use this field to provide any additional information about the model itself (e.g. number of parameters)
        \item \texttt{evaluation_results.generation_config.generation_args}: Specify additional arguments used to generate outputs from the model
        \item \texttt{evaluation_results.generation_config.additional_details}: Use this field to provide any additional information about the evaluation process that is not captured elsewhere
    \end{itemize}
\end{enumerate}

\textbf{Instance-Level Data}

For evaluations that include per-sample results, the individual results should be stored in a companion \texttt{{uuid}_samples.jsonl} file in the same folder (one JSONL per JSON, sharing the same UUID). The aggregate JSON file refers to its JSONL via the \texttt{detailed_evaluation_results} field. The instance-level schema (\texttt{instance_level_eval.schema.json}) supports three interaction types:
\begin{itemize}
    \item \textbf{\texttt{single_turn}}: Standard QA, MCQ, classification — uses \texttt{output} object
    \item \textbf{\texttt{multi_turn}}: Conversational evaluations with multiple exchanges — uses \texttt{messages} array
    \item \textbf{\texttt{agentic}}: Tool-using evaluations with function calls and sandbox execution — uses \texttt{messages} array with \texttt{tool_calls}
\end{itemize}

Each instance captures: \texttt{input} (raw question + reference answer), \texttt{answer_attribution} (how the answer was extracted), \texttt{evaluation} (score, is_correct), and optional \texttt{token_usage} and \texttt{performance} metrics. Instance-level JSONL files are produced automatically by the eval converters.
\begin{comment}
Example \texttt{single_turn} instance:
\begin{lstlisting}[language=json]
{
  "schema_version": "instance_level_eval_0.2.2",
  "evaluation_id": "math_eval/meta-llama/Llama-2-7b-chat/1706000000",
  "model_id": "meta-llama/Llama-2-7b-chat",
  "evaluation_name": "math_eval",
  "sample_id": 4,
  "interaction_type": "single_turn",
  "input": { "raw": "If $2^10 = 4^x$, what is the value of x?", "reference": "5" },
  "output": { "raw": "Rewrite 4 as 2\^2, so 4\^x = 2\^(2x). Since 2\^10 = 2\^(2x), x = 5." },
  "answer_attribution": [{ "source": "output.raw", "extracted_value": "5" }],
  "evaluation": { "score": 1.0, "is_correct": true }
}
\end{lstlisting}

\textbf{Agentic Evaluations}

For agentic evaluations (e.g., SWE-Bench, GAIA), the aggregate schema captures configuration under \texttt{generation_config.generation_args}:

\begin{lstlisting}[language=json]
{
  "agentic_eval_config": {
    "available_tools": [
      {"name": "bash", "description": "Execute shell commands"},
      {"name": "edit_file", "description": "Edit files in the repository"}
    ]
  },
  "eval_limits": {"message_limit": 30, "token_limit": 100000},
  "sandbox": {"type": "docker", "config": "compose.yaml"}
}
\end{lstlisting}

At the instance level, agentic evaluations use \texttt{interaction_type: "agentic"} with full tool call traces recorded in the \texttt{messages} array. See the Inspect AI test fixture for a GAIA example with docker sandbox and tool usage.
\end{comment}

\subsection{Worked Examples}
\subsubsection{Example 1: Conflicting MMLU Records}\label{app:governance-example1}

To make the governance mechanisms concrete, consider the LLaMA 65B/MMLU example from Section \ref{sec:intro}, where the model scores 63.7 under HELM and 48.8 under lm-eval-harness \citep{fourrier2023openllmleaderboard}. Under \eeeshort{}, both results are valid records: each receives its own UUID, each carries eval\_library metadata identifying the harness, and each preserves the generation configuration and prompt template available at submission time. The validator does not flag these as duplicates because the eval\_library field differs. A downstream user comparing the two records sees the discrepancy in the metadata directly and can decide whether to treat them as comparable, rather than discovering the difference through a blog post months later. If a third contributor later submits a third MMLU record for LLaMA 65B without specifying the harness, the validator emits a warning, the record is accepted with the missing field recorded as absent (Section~\ref{sec:schema-gen}), and any downstream analysis that requires harness-level disambiguation can filter it out. No human governance intervention is needed for this case; the schema and validator handle it. Governance intervention is reserved for cases where metadata is contested rather than merely missing.

\subsubsection{Example 2: Disputed Agentic Record}\label{app:governance-example2}

A contributor submits records scraped from a public agentic-evaluation leaderboard. The records pass validation, but the agent's developers later contest them, claiming the leaderboard ran a deprecated harness version and that the current version yields lower scores; another contributor argues the original entry should remain as the authoritative public record at the time of reporting. The schema cannot resolve this because both parties agree on the metadata. Maintainers handle it by retaining the original record (records are immutable; Section~\ref{app:governance-corrections}), annotating it with a disputed flag under \emph{additional\_details}, pointing to the issue thread, inviting the developers to submit a new record under the current harness, and documenting the resolution in the changelog. \eeeshort{} does not arbitrate which run is ``correct''; it ensures both runs and their dispute are visible.

\begin{lstlisting}[style=shell, caption={Installing and running a converter.}]
pip install `every-eval-ever[all]'
every_eval_ever convert helm    --log_path path/to/helm/logs
every_eval_ever convert inspect --log_path path/to/run.eval
every_eval_ever convert lm_eval --log_path path/to/results.json \
                                --include_samples
\end{lstlisting}

\newpage
\begin{lstlisting}[style=shell, caption={CLI validation examples.}]
# Validate a single aggregate file
uv run python -m every_eval_ever validate path/to/uuid.json

# Validate instance-level data
uv run python -m every_eval_ever validate path/to/uuid_samples.jsonl

# Validate an entire benchmark directory
uv run python -m every_eval_ever validate data/mmlu/
\end{lstlisting}

\FloatBarrier
\section{Case Studies: Reproducibility and Implementation Details}
\label{app:casestudies}

\subsection{Case 1}

We report the aggregate records used for the agentic cost--accuracy analysis in Section~\ref{case1}. CocoaBench ~\citep{cocoabenchteam2026cocoabenchevaluatingunifieddigital} is used to illustrate how runtime and cost can change the interpretation of accuracy across scaffold--backbone combinations. CORE-Bench Hard results from HAL ~\citep{kapoor2025holisticagentleaderboardmissing} are used as a representative within-benchmark slice showing how both scaffold and backbone choices affect the cost--accuracy tradeoff. 

The corresponding records are available in the \eee{} datastore under the CocoaBench and HAL benchmark directories, illustrated respectively in Tables~\ref{tab:cocoabench} and~\ref{tab:hal_corebench}.

\label{app:casestudies1}
\begin{table}[h]
\centering
\caption{Aggregate CocoaBench results represented in \eee.}
\label{tab:cocoabench}
\resizebox{\columnwidth}{!}{%
\begin{tabular}{llcccc}
\toprule
Agent Scaffold & Model Backbone & Accuracy (\%) & Avg. time (s) & Avg. cost (\$) & Total cost (\$) \\
\midrule
Codex & GPT-5.4 & 45.1 & 377.8 & 0.7 & 111.4 \\
OpenClaw & GPT-5.4 & 45.1 & 502.1 & 1.0 & 166.3 \\
Cocoa Agent & GPT-5.4 & 36.6 & 596.8 & 2.3 & 342.4 \\
OpenClaw & Claude Sonnet 4.6 & 34.0 & 693.7 & 2.0 & 300.6 \\
Claude Code & Claude Sonnet 4.6 & 25.5 & 673.8 & 1.5 & 194.5 \\
Cocoa Agent & Gemini 3.1 pro & 30.7 & 715.4 & 1.2 & 186.0 \\
\bottomrule
% \vspace{0.1mm}
\end{tabular}
}
\end{table}

\begin{table}[h]
\centering
\caption{Representative HAL results on CORE-Bench Hard represented in \eee.}
\label{tab:hal_corebench}
\begin{tabular}{llcc}
\toprule
Agent Scaffold & Model Backbone & Accuracy (\%) & Total cost (\$) \\
\midrule
Claude Code & Claude Opus 4.1 & 42.2 & 331.8 \\
Claude Code & Claude Sonnet 4.5 & 62.2 & 68.3 \\
Claude Code & Claude Opus 4.5 & 77.8 & 87.2 \\
CORE-Agent & Claude Sonnet 3.7 & 35.6 & 73.0 \\
CORE-Agent & Claude Opus 4.1 & 51.1 & 412.4 \\
CORE-Agent & Claude Opus 4.5 & 42.2 & 169.0 \\
\bottomrule
\end{tabular}
\end{table}

\subsection{Case 2}
\label{app:casestudies2}
This appendix gives the implementation details for Case Study 2 (Section~\ref{case2}). Records for the perplexity comparison in Table~\ref{tab:ppl_comparison} were obtained from two sources: lm-eval-harness logs ingested via the automated lm\_eval converter (Section~\ref{sec:converters}, App.~\ref{sec:app-conv-lmeval}), and GPTQ-style evaluation scripts contributed as manual records. The lm\_eval converter preserves metric keys verbatim from harness output --- \texttt{word\_perplexity} and \texttt{byte\_perplexity} are stored as distinct \texttt{evaluation\_name} values in the \texttt{MetricConfig} block (Section~\ref{sec:schema-results}), ensuring records with different normalization conventions are never silently aggregated (Section~\ref{sec:design-process}). GPTQ-style records were contributed with \texttt{metric\_name} set to reflect token-level normalization. The corresponding records are available in the \eee{} datastore under the WikiText benchmark directory.

\subsection{Case 3}
\label{app:case3}
%\subsection{HELM Reproducibility Audit Details}
%\label{sec:app-helm-audit}

This appendix gives the implementation details for Case Study~3
(Section~\ref{sec:case3}). It expands the four reproducibility patterns
summarized in the main text---non-comparable examples, serving artifacts,
stochastic disagreement, and residual answer differences---by describing how
official and local records are aligned, how the heatmap values are computed, and
the evidence supporting each diagnosis.
Code used to produce these experiments is available on 
\href{https://anonymous.4open.science/r/eval_audit-76F2}{Anonymous Github}.
% TODO: swap between anon and non-anon
%\href{https://github.com/AIQ-Kitware/eval_audit/tree/dff775d}{Github}.

We compare HELM-released records, which we call the \emph{official} side, to
local reproductions for Pythia-6.9B, Vicuna-7B v1.3, and Falcon-7B on fourteen
single-turn benchmarks, i.e., evaluations where each instance consists of a single prompt-response pair rather
than multi-turn evaluations containing multiple prompts and responses within an instance or agentic tool-use tasks.
Both sides are converted to \eee{} and aligned by \texttt{sample\_hash}, which
the HELM converter computes as a SHA-256 hash of the rendered prompt
concatenated with the first correct reference.

Figure~\ref{fig:cs3-helm-heatmap} summarizes the 42 \mbox{(model, benchmark)}
comparisons.
For each cell with a successful join, the displayed value is a micro-average
over aligned \mbox{(instance, core metric)} score pairs: the number whose
official. A micro average is equivalent to the local score difference divided by
the total number of score pairs compared. The local score difference is at most \texttt{1e-9}.
Thirty-nine cells align successfully under this metric; the three failed joins are all
Entity-Matching.

The heatmap is a metric-level audit calculated after content-based alignment.
Its unit of comparison is a scored \mbox{(instance, metric)} pair, so agreement
implies that the two recorded scores are numerically equal up to the tolerance rather 
than that the text outputs are identical.
\eee{} provides the data needed for this comparison---sample hashes,
per-instance outputs, per-metric scores, and generation parameters.

\subsubsection{Entity-Matching}
\label{sec:app-helm-entity-matching}

Entity-Matching is a non-comparability failure. Official and local records each
contain 1,000 examples from the Abt--Buy dataset \citep{kopcke2010evaluation}, and HELM positional identifiers overlap
1,000/1,000, but the \texttt{sample\_hash} sets have zero overlap.
\texttt{EntityMatchingScenario} constructs a candidate table with
\texttt{pd.merge}, then applies fixed-seed sampling over that ordered table and
assigns ids positionally. A pandas row-order change makes the same input files,
seed, and HELM recipe select different product pairs.

End-to-end HELM reproduction confirms the split: the audit environment
(pandas \texttt{2.3.3}, numpy \texttt{2.2.6}) matches the local
\texttt{scenario\_state.json}, while the HELM v0.3.0-era environment
(pandas 2.0.3, numpy 1.23.5) matches the official one; cross-environment prompt-reference overlap, measured by \texttt{sample\_hash},
is 0/1000. Thus, an id-based join would align different prompts,
while the \eee{} content join correctly reports \texttt{join\_failed}.

\subsubsection{SyntheticReasoning-Natural}
\label{sec:app-helm-sr-natural}

SyntheticReasoning-Natural is a synthetic-reasoning benchmark in natural language based on
RuleTaker-style facts and rules~\citep{helm,clark2020transformers}.
In our results, the SyntheticReasoning-Natural $\times$ Pythia-6.9B is the largest discrepancy among cells that joined successfully. The \texttt{sample\_hash} join succeeds,
so the official and local records compare the same prompt-reference pairs. The
divergence is in the scored outputs: official records contain empty
\texttt{output.raw} completions and score 0.0 under the set-match metrics,
while local records contain non-empty completions and recover non-zero scores.
Lower-level official artifacts show that the backend returned generated text,
but each completion began with the newline stop sequence and was trimmed to the
empty string before scoring.

The HELM recipe itself requests deterministic decoding:
\texttt{temperature=0.0}, \texttt{num\_outputs=1}, \texttt{max\_tokens=20}, and
\texttt{stop\_sequences=[``\textbackslash n'']}. Lower-level official artifacts
show that the model produced outputs, but each began with the newline stop
sequence and was trimmed before scoring. The audit trail indicates that HELM
v0.3.0 routed the official Pythia call through Together.ai when
\texttt{model\_deployment} was unset, whereas the local run used Hugging Face
transformers. Thus, \eee{} exposes the symptom, while the historical backend
must be recovered from HELM runtime artifacts or source code.

\subsubsection{WikiFact}
\label{sec:app-helm-wikifact}

WikiFact is a factual-knowledge probing scenario, related to LAMA-style
cloze probes for relational knowledge \citep{helm,petroni-etal-2019-language}.
WikiFact has similar cell-level score agreement for all models: 0.922 for
Pythia, 0.920 for Vicuna, and 0.927 for Falcon. Its recipe uses
\texttt{temperature=1.0}, \texttt{num\_outputs=5}, and
\texttt{max\_tokens=8}, so independent executions can differ.

A simple independent Bernoulli model for binary score-pair outcomes explains
why agreement can stabilize below one. Because the heatmap micro-averages over
aligned \mbox{(instance, core metric)} score pairs, let \(i\) index a binary
WikiFact score pair. Let \(X_i\) and \(Y_i\) be the binary scores from two
independent executions of the same stochastic recipe on score pair \(i\), and
let \(p_i\) be the probability that one full \texttt{num\_outputs=5} execution
scores 1 on that prompt-metric pair. For a fixed score pair, agreement has two
disjoint cases:
\[
  \underbrace{\Pr[X_i = Y_i]}_{\substack{
    \text{agreement on}\\
    \text{prompt }i
  }}
  =
  \underbrace{p_i^2}_{\substack{
    X_i=1,\;Y_i=1\\
    \text{both score }1
  }}
  +
  \underbrace{(1-p_i)^2}_{\substack{
    X_i=0,\;Y_i=0\\
    \text{both score }0
  }}.
\]
Averaging uniformly over score-pairs, with
\(\bar p = \mathbb{E}_i[p_i]\), gives
\[
  \underbrace{\mathbb{E}_i[p_i^2 + (1-p_i)^2]}_{\substack{
    \text{heterogeneous}\\
    \text{expected agreement}
  }}
  =
  \underbrace{
    \bar p^2 + (1-\bar p)^2
  }_{\substack{
    \text{homogeneous Bernoulli}\\
    \text{agreement at average rate}
  }}
  +
  \underbrace{2\,\mathrm{Var}_i(p_i)}_{\substack{
    \text{extra agreement from}\\
    \text{prompt-dependent difficulty}
  }}.
\]
The homogeneous Bernoulli agreement is therefore a lower bound in the
symmetric \texttt{@5}-versus-\texttt{@5} idealization, since the variance
term is nonnegative. Intuitively, easy score pairs tend to agree by both
succeeding, while hard score pairs tend to agree by both failing. The actual
WikiFact artifacts are slightly asymmetric: the official HELM
\texttt{scenario\_state.json} stores one completion per request state even
though the adapter records \texttt{num\_outputs=5}, and the official
\texttt{exact\_match} and \texttt{exact\_match@5} scores are pointwise
identical. The local audit artifacts store five completions, so the observed
comparison is better modeled as a official-\texttt{@1} versus local-\texttt{@5}
stochastic score-pair comparison. The observed WikiFact values are consistent
with this stochastic recipe, with any residual plausibly due to estimator
uncertainty or backend drift. \eee{} makes this diagnosis checkable by
preserving decoding parameters, per-instance scores, and raw outputs.

\subsubsection{Residual disagreements and scope}
\label{sec:app-helm-residual}

Several other cells fall below perfect agreement without exhibiting
non-comparability, empty completions, or stochastic decoding. Spot checks show
aligned instances and mostly matching answers; the residual differences are
genuine answer changes concentrated in Pythia-6.9B. A plausible lead is
serving-stack sensitivity, but the evidence cannot separate model strength from
checkpoint, quantization, precision, tokenizer, or generation-kernel
differences.

\subsection{Case 4}
\label{app:casestudies4}

In Case Study 4, we conduct an Item Response Theory (IRT) meta-analysis of instance-level evaluation data collected and stored in \eee.
IRT models estimate latent parameters for dataset \textit{items} (i.e., instances/examples) and \textit{subjects} (i.e., AI models), and have been used in prior evaluation practices \citep{polo2025statistical,polo2024efficient} leaderboards \citep{kipnis2024metabench,shabtaylivexiv}, meta-evaluations \citep{rodriguez2021evaluation} and applications such as curriculum learning \citep{rodriguez2021evaluation,meng2025psychology,polo2024tinybenchmarksevaluatingllmsfewer}, often those uses were limited by the amount of available data~\citep{habba2025dove,habba2026growing}.  
The one-parameter logistic (1PL) IRT model estimates item difficulty and subject ability, while more complex IRT models include more item-level parameters such as discriminability and feasibility \citep{rodriguez2021evaluation}. 
IRT models the probability of subject $j$ labeling item $i$ correctly ($z_{ij}=1$) as a function of subject $j$'s latent ability and item $i$'s latent difficulty.
Parameters are learned via optimization from a dataset of graded (i.e., correct or incorrect) responses from subjects for a set of items:

\begin{align}
	\label{eq:irt}
	p(z_{ij} &= 1 \vert \theta_j, b_i) = \frac{1}{1 + e^{-(\theta_j - b_i)}}\\
	\log \mathcal{L} &= \sum_{j=1}^J \sum_{i=1}^I \log p(Z_{ij}=z_{ij} \vert \theta_j, b_i)
\end{align}

Instance-level data collection is an expensive prerequisite and thus often a bottleneck for IRT research in NLP.
For the case study, we selected three datasets currently available in \eee{} with instance-level evaluations.
GPQA Diamond \cite{rein2024gpqa} includes responses for 198 items from 69 subjects; 
Wordle Arena \cite{murphy2025ai} includes responses for 63 items from 46 subjects;
JudgeBench \cite{Tan2024JudgeBenchAB} includes responses for 350 items from 55 subjects.
We extract the \textit{is\_correct} value for each instance-level record in a dataset to construct the response matrix $Z$.
For example, $Z^{\text{JudgeBench}}$ has 55 rows (subjects) and 350 columns (items).  

We fit a 1PL model for each dataset using the \texttt{py-irt}  package version 0.7.1 \citep{lalor2023py}. 
\texttt{py-irt} implements IRT model fitting via variational inference and can scale to large evaluation datasets via GPU-scaled training.
Specifically, the joint posterior distribution $p(\Theta,B|Z)$ is approximated by a variational distribution $q(\Theta, B)$, and latent variables are learned by minimizing the KL-Divergence between $q(\Theta,B)$ and $p(\Theta,B \vert Z)$ \cite{lalor2023py}.

\clearpage
%\input{checklist.tex}
%removed for arxiv, to be added back!

\end{document}